\documentclass{article}

\usepackage{arxiv}

\usepackage[utf8]{inputenc}
\usepackage[T1]{fontenc}
\usepackage{hyperref}
\usepackage{url}
\usepackage{booktabs}
\usepackage{amsmath}
\usepackage{amsfonts}
\usepackage{amssymb}
\usepackage{microtype}
\usepackage{graphicx}
\usepackage{float}
\usepackage{tikz}
\usetikzlibrary{arrows.meta,positioning,calc,fit,backgrounds}
\usepackage{pgfplots}
\pgfplotsset{compat=1.17}
\usetikzlibrary{pgfplots.groupplots}
\usepackage[numbers,sort&compress]{natbib}
\usepackage{doi}
\usepackage{xcolor}
\usepackage{colortbl}
\usepackage{xspace}
\usepackage{makecell}
\usepackage{multirow}
\usepackage{array}
\usepackage{subcaption}
\usepackage{cleveref}

\newcommand{\sysname}{\textsc{SocioVerse2}\xspace}
\newcommand{\sysnameTitle}{\textsc{SocioVerse2}}

\newcommand{\svone}{\textsc{SocioVerse}~1.0\xspace}

\newcommand{\bfpe}{\ensuremath{B = f(P, E)}\xspace}

\newcommand{\Pop}{\ensuremath{P}\xspace}
\newcommand{\Env}{\ensuremath{E}\xspace}

\newcommand{\userpool}{Population MCP\xspace}
\newcommand{\eventtool}{Event MCP\xspace}
\newcommand{\svskill}[1]{\textsc{#1}}

\definecolor{linknavy}{RGB}{17,45,110}

\definecolor{figink}{RGB}{40,58,82}
\definecolor{figblue}{RGB}{228,237,246}
\definecolor{figgray}{RGB}{244,245,247}
\definecolor{figaccent}{RGB}{193,94,44}
\definecolor{figaccentfill}{RGB}{250,236,227}

\tikzset{
  figfont/.style={font=\sffamily\small, text=figink},
  figsub/.style={font=\sffamily\scriptsize, text=figink!75},
  figlabel/.style={font=\sffamily\footnotesize, text=figink!85, inner sep=2pt},
  fignode/.style={figfont, draw=figink, line width=0.6pt, rounded corners=2.5pt,
                  fill=figblue, inner sep=5pt, align=center},
  figplain/.style={figfont, draw=figink!55, line width=0.5pt, rounded corners=2.5pt,
                   fill=figgray, inner sep=5pt, align=center},
  figmcp/.style={figfont, draw=figink, line width=0.9pt, rounded corners=2.5pt,
                 fill=white, inner sep=5pt, align=center},
  figaccentnode/.style={figfont, draw=figaccent, line width=0.8pt, rounded corners=2.5pt,
                        fill=figaccentfill, inner sep=5pt, align=center},
  figarrow/.style={-{Stealth[length=2.4mm,width=1.9mm]}, line width=0.7pt, draw=figink},
  figarrowlight/.style={-{Stealth[length=2mm,width=1.6mm]}, line width=0.5pt, draw=figink!60},
  figarrowaccent/.style={-{Stealth[length=2.4mm,width=1.9mm]}, line width=0.8pt, draw=figaccent},
  figdashed/.style={dash pattern=on 2.2pt off 1.8pt},
  figframe/.style={draw=figink!45, line width=0.6pt, figdashed, rounded corners=4pt, inner sep=8pt},
}

\title{\sysnameTitle: A Longitudinal Dynamic Social Simulation Framework under a Human-AI Co-evolutionary Paradigm}

\date{}

\renewcommand{\headeright}{Technical Report}
\renewcommand{\undertitle}{Technical Report}
\renewcommand{\shorttitle}{\textsc{SocioVerse2}}

\usepackage{authblk}

\author[1,2]{Xinnong~Zhang\textsuperscript{*}}
\author[1,2]{Jiayu~Lin\textsuperscript{*}}
\author[1,4]{Jia~Wang\textsuperscript{*}}
\author[2]{Yixu~Huang}
\author[2]{Xinyi~Mou}
\author[2]{Yingqian~Wu}
\author[2]{Jingcong~Liang}
\author[5]{Shijun~Lei}
\author[6]{Jianing~Shi}
\author[2]{Guanying~Li}
\author[7]{Siyuan~Wang}
\author[8]{Hanjia~Lyu}
\author[9]{Zhenfei~Yin}
\author[2]{Yunlu~Yin}
\author[2]{Siming~Chen}
\author[3]{Yulan~He}
\author[10]{Jiebo~Luo}
\author[1,2]{Xuanjing~Huang}
\author[2]{Liyin~Jin}
\author[2]{Baohua~Zhou}
\author[3]{Hanqi~Yan\textsuperscript{\dag}}
\author[1,2]{Zhongyu~Wei\textsuperscript{\dag}}

\affil[{{}}]{\parbox{\textwidth}{\centering
\textsuperscript{1}Shanghai Innovation Institute\quad
\textsuperscript{2}Fudan University\quad
\textsuperscript{3}King's College London\linebreak
\textsuperscript{4}Tongji University\quad
\textsuperscript{5}Northwestern Polytechnical University\linebreak
\textsuperscript{6}The London School of Economics and Political Science\quad
\textsuperscript{7}The Chinese University of Hong Kong\linebreak
\textsuperscript{8}Singapore Management University\quad \textsuperscript{9}University of Oxford\quad
\textsuperscript{10}University of Rochester}}
\affil[{{}}]{\parbox{\textwidth}{\centering\rule{0pt}{1.6em}\texttt{xnzhang23@m.fudan.edu.cn},\ \texttt{zywei@fudan.edu.cn}}}

\begin{document}
\maketitle
\begingroup
\renewcommand{\thefootnote}{}%
\footnotetext{\noindent\textsuperscript{*}Equal contribution.}
\footnotetext{\textsuperscript{\dag}Corresponding authors.}
\footnotetext{This work was completed during the research period at Shanghai Innovation Institute.}%
\addtocounter{footnote}{-2}%
\endgroup

\vspace{-4.4em}
\begin{center}
  \hypersetup{pdfborder={0 0 0}}%
  \textbf{Project page}: \href{https://socioverse.fudan-disc.com/}{\color{linknavy}\nolinkurl{https://socioverse.fudan-disc.com/}}
\end{center}
\vspace{1.2em}

\begin{abstract}
Social simulation offers the social sciences an experimental instrument that the real world cannot supply, and generative agents have transformed it by acting as silicon samples that unite agent-based modeling with real behavioral data.
Existing platforms verify collective behavior, align simulated populations with real societies in cross-sections, and employ autonomous agents for the research process. However, two social science requirements remain without systematic support: intervention in the content of a simulation and the researcher's control over the process that produces it.
We present \sysname, which extends \svone into a human-AI co-evolutionary paradigm built from two loops and one infrastructure.
The \emph{longitudinal simulation loop} simulates the target population with evolving environments and forks counterfactual branches via interventions.
The \emph{controllable research loop} takes the study itself as an editable state and updates state versions via controllable editing.
The \emph{social science agentic infrastructure} carries both loops through composable skills with researcher checkpoints, a population service over five persona pools, and an environment service over 21 real-world signal sources with point-in-time guarantees.
We validate \sysname across three case families and seven case studies, from reproducing canonical agent-based models to modeling policy processes on real records and nowcasting macro-economic indices beyond the response model's knowledge cutoff.
With the human-AI co-evolutionary paradigm, these cases go beyond system demonstrations to become substantive studies that investigate frontier questions in their respective disciplines.
Code, data services, and a workbench are released as open-source resources.

\end{abstract}

\newpage
\tableofcontents
\newpage

\section{Introduction}
\label{sec:intro}
Social simulations offer the social sciences an experimental instrument that the real world cannot supply: traditional field experiments are costly, hard to reproduce, and ethically bounded, while a simulated society places populations, environments, and interventions under direct control~\citep{schelling1971dynamic,epstein1996growing}. Generative agents have further transformed social simulations. By combining rule-based agent-based modeling (ABM) with real-world behavioral data, they act as ``silicon samples'' capable of humanoid reasoning and natural-language expression~\citep{park2023generative,argyle2023out}.
Recently, a fast-growing family of studies has built social simulators and platforms to expand the toolkit of social science research, from hand-built sandboxes~\citep{gao2023s3,vezhnevets2023concordia} to agent-operated workflows~\citep{wang2025yulan,piao2026agentsociety2}, from reactive verification that reproduces known phenomena to proactive exploration that asks what would follow from conditions never observed, such as an untested policy~\citep{li2026whatif,lee2025injectforkcompare}.

However, these platforms produce collective behaviors and simulated populations that rely solely on computational, systematic design and the abilities of autonomous agents~\citep{mou2024unveiling,park2024generative1000,piao2025agentsociety,zhang2025socioverse,yang2024oasis,piao2026agentsociety2}.
Recent audits show that autonomous research agents can fabricate results and drift from their announced plans, producing unqualified papers below publication standards~\citep {zhang2026autoresearch,miyai2025jrscientist,zhu2025implementation}.
Accordingly, the core requirements that facilitate social science lie in experimental intervention over research content and human scientists' control over the research process.
Human involvement has been shown to be an effective means of mitigating base models’ limited transferability. Researcher feedback at each stage significantly improves agent-generated research~\citep{schmidgall2025agentlab,gottweis2025coscientist} and human decision gates raise the share of feasible hypotheses in empirical social science~\citep{zhu2026hler}.
For such collaborative requirements between human researchers and autonomous agents, current social science simulation platforms offer little systematic support (\Cref{sec:background}).

\begin{figure*}[t]
  \centering
  \resizebox{\textwidth}{!}{\input{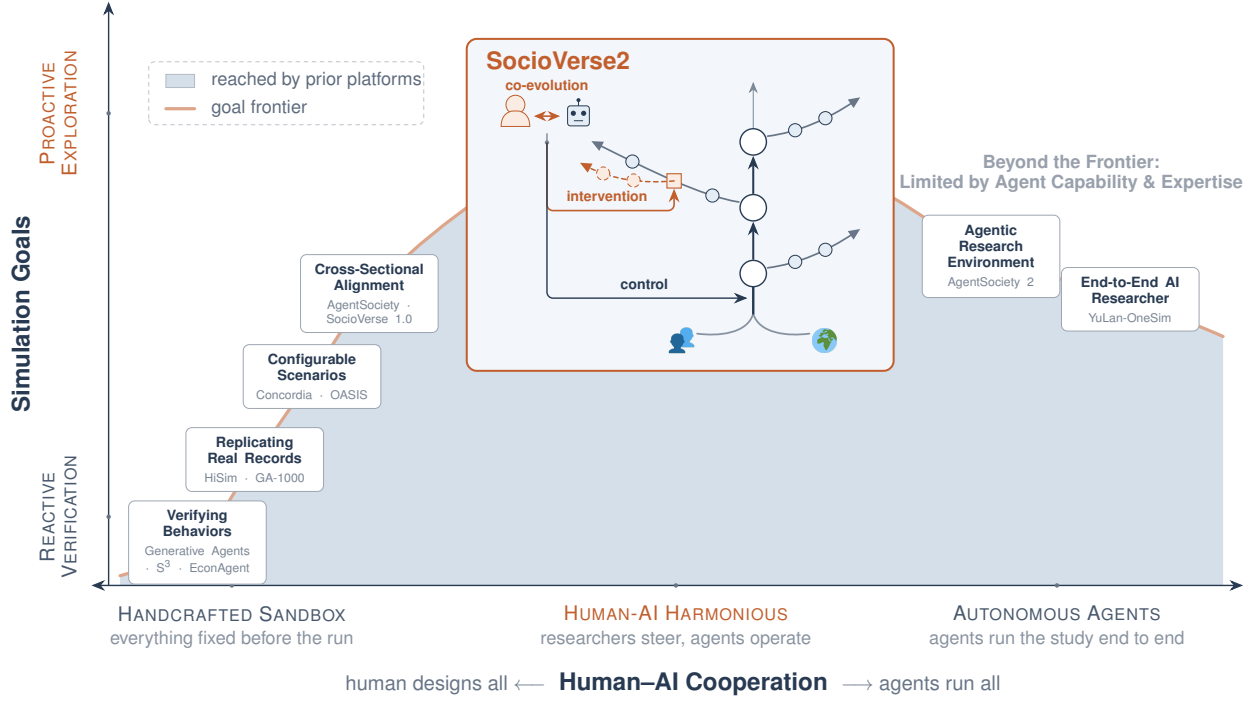}}
  \caption{
  {The development of existing LLM social simulation platforms along two axes: human-AI cooperation (x-axis) and simulation goals (y-axis). Our \sysname supports proactive exploration via integrating agentic infrastructure and counterfactual intervention capabilities based on the user query, also backed up by the human-in-the-loop control for better quality assurance.} 
  }
  \label{fig:landscape}
\end{figure*}

We formulate these considerations as a human–AI co-evolutionary paradigm along three dimensions.
\textbf{Longitudinal Intervention}: social phenomena emerge and evolve over time~\citep{abbott2001time,halaby2004panel}. During this dynamic process, designed intervention experiments are essential for isolating and studying different variables and supporting counterfactual hypotheses~\citep{holland1986statistics,rubin1974estimating}. Social simulation platforms should therefore create diverse control groups that cannot be simply reproduced from existing reality~\citep{campbell1969reforms,morgan2015counterfactuals}, as shown along the y-axis of \Cref{fig:landscape}.
\textbf{Controllable Loop}: as shown along the x-axis of \Cref{fig:landscape}, early rigid sandboxes fix every simulation setting, including information flow, agentic framework, interactive topology, etc., and thus limit the evolving ability of silicon samples~\citep{park2023generative,gao2023s3}. In contrast, current autonomous research platforms leave the simulation entirely in the coding agent’s hands and pursue an end-to-end simulator~\citep{wang2025yulan,lu2024aiscientist}. The importance of human-in-the-loop during the research process should be highlighted to create a controllable loop~\citep{horvitz1999mixed,amershi2019guidelines,shneiderman2020hcai}, especially in complex, iterative science experiments~\citep{messeri2024illusions,wang2023scientific}.
\textbf{Foundation Basis}: beyond the base model's ability, the quality of social simulations relies on the grounding data and agent infrastructure~\citep{bail2024generative}. Behaviors generated during the simulation should align with real-world populations and environments~\citep{granovetter1985economic,zhang2025socioverse,santurkar2023whose,dillion2023replace}. The system therefore must retrieve grounded information, assemble it into context, map it to actions, and update it iteratively through an adaptive agentic harness that expands the frontier of social science research.

In this paper, we present \sysname that aims to bridge the gap between current platforms and social science research requirements.
In \svone, we organized simulation around four alignment-centered engines: the social environment, the user engine, the scenario engine, and the behavior engine~\citep{zhang2025socioverse}.
\sysname carries this design forward and extends it into a human-AI co-evolutionary paradigm using \emph{Two loops} and \emph{One infrastructure}: the \textbf{longitudinal simulation loop}, the \textbf{controllable research loop}, and the \textbf{social science agentic infrastructure} (\Cref{sec:framework}).
The \sysname framework models social simulation as collective behaviors of silicon samples that depend on the joint effects of real-world heterogeneous populations and grounded social-environment signals, where a unified formulation is employed to demonstrate both the simulation loop and research loop~(\S\ref{sec:framework:overview}). We explain the implementation of the longitudinal feature in the simulation loop (\S\ref{sec:framework:behavior}), then introduce intervention operations that create branches to compare counterfactuals and variables (\S\ref{sec:framework:loop}). We introduce controllable editing in the research loop, enabling diverse versions across the research process (\S\ref{sec:framework:control}). Human social scientists can engage through both loops by designing counterfactual experiments, providing feedback, and making adjustments, thereby enabling co-evolution between humans and social science agents.
The loops are supported by a social science agentic infrastructure, including a complete lifecycle workflow and two standard model context protocols (MCP) for population alignment and environment grounding (\Cref{sec:infra}).

We validate \sysname across three case families and seven case studies with systematic evaluation methods (\S\ref{sec:evaluation:methods}).
\emph{Simulation mechanisms} focus on interpretation of social mechanisms and phenomena in different social science fields (\S\ref{sec:cases:mechanisms}).
\emph{Policy simulation} studies the potential results during the policymaking process (\S\ref{sec:cases:policy}).
\emph{Macro-index forecasting} uses real-world deterministic indices and signals as ground truth to evaluate the framework's quantitative performance (\S\ref{sec:cases:macro}).
Each family carries its own real-world grounding, and together they cover simulation goals from verifying known dynamics to estimating future potentials, from helping design policies to predicting indices ahead of time.
The two loops turn these cases from single runs into practical research. Every mechanism probe in the designs, from the intervention of the opinion dynamics to the ablations of the policy forecasting, is either a branch raised inside the simulation loop or a version raised by the researcher in the research loop, so that each reported gain is attributable to a controlled change of the population, the environment, or the behavior function. Researchers enter both loops in every case, declaring interventions, replacing components after inspecting the panel records, and deciding which version is reported, and it is this collaboration rather than one autonomous pass that carries a case to a result validated against ground truth.

In general, our contributions are:
\begin{itemize}
  \item \textbf{A Dynamic Longitudinal Simulation Framework.} We formalize the shift from cross-sectional prediction to longitudinal trajectory simulation.
  \item \textbf{A Human-AI Co-evolutionary Research Paradigm.} We combine autonomous agents with human social scientists in two collaborative and co-evolutionary loops via interventions and editable components.
  \item \textbf{A Social Science Agentic Infrastructure.} We release the \sysname runtime as a modular open-source package, together with standardized MCP services for population alignment and environment grounding, an agentic skill pipeline, and an online research workbench that make the entire study lifecycle auditable and reproducible.
\end{itemize}

\section{Background and Positioning}
\label{sec:background}

\begin{table}[t]
  \centering
  \caption{Representative LLM social simulation platforms, sorted by release date. \checkmark = supported, $\circ$ = partial, --- = not supported.}
  \label{tab:platform-comparison}
  \footnotesize
  \setlength{\tabcolsep}{5pt}
  \begin{tabular}{@{}lcclll@{}}
    \toprule
    \textbf{Platform} & \multicolumn{2}{c}{\textbf{Foundation Basis}} & \textbf{Human--AI} & \textbf{Simulation} & \textbf{Open} \\
    \cmidrule(lr){2-3}
    & \textbf{Real $P$} & \textbf{Real $E$} & \textbf{Division of Labor} & \textbf{Goal} & \textbf{Source} \\
    \midrule
    Generative Agents~\citep{park2023generative} & --- & --- & handcrafted & verify behaviors & \checkmark \\
    S$^3$~\citep{gao2023s3} & $\circ$ & $\circ$ & handcrafted & verify behaviors & --- \\
    Concordia~\citep{vezhnevets2023concordia} & --- & --- & configured & verify behaviors & \checkmark \\
    EconAgent~\citep{li2023econagent} & $\circ$ & --- & handcrafted & stylized facts & \checkmark \\
    HiSim~\citep{mou2024unveiling} & $\circ$ & \checkmark & handcrafted & real-data match & \checkmark \\
    GA-1000~\citep{park2024generative1000} & \checkmark & --- & handcrafted & real-data match & $\circ$ \\
    OASIS~\citep{yang2024oasis} & $\circ$ & $\circ$ & configured & verify behaviors & \checkmark \\
    AgentSociety~\citep{piao2025agentsociety} & $\circ$ & \checkmark & configured & real-data match & \checkmark \\
    \svone~\citep{zhang2025socioverse} & \checkmark & \checkmark & configured & real-data match & \checkmark \\
    YuLan-OneSim~\citep{wang2025yulan} & $\circ$ & $\circ$ & autonomous agents & verify behaviors & \checkmark \\
    AgentSociety 2~\citep{piao2026agentsociety2} & \checkmark & \checkmark & agentic, human-steered & reproduce prior studies & \checkmark \\
    \midrule
    \textbf{\sysname} & \checkmark & \checkmark & \textbf{human-AI harmonious} & \textbf{proactive exploration} & \checkmark \\
    \bottomrule
  \end{tabular}
\end{table}

Since large language models became the decision engine of social simulation agents, platforms have proliferated, and \Cref{fig:landscape} maps them on the two axes along which they differ.
\emph{Human-AI Cooperation} describes how a study is operated, from entirely human experts' handcrafted design to full agent automation. At one end, the hand-crafted sandbox fixes the information flow and interaction topology. The configurable framework relaxes this by exposing the population, the environment, and the scenario as parameters.
At the other end, autonomous agents take a question, carry out experiments independently, and finally return a report. Human researchers provide little feedback during the whole loop. We aim for harmonious collaboration in the middle, where agents operate the machinery interactively, guided by researchers' feedback.

\emph{Simulation Goals} describe what a study is for, from passive execution to active design. Reactive verification asks whether a simulation reproduces known behavior patterns or phenomena, and the results are expected. Proactive exploration asks what would happen from current conditions, which requires introducing unknown interventions in the simulation and modifying the simulation structure, rather than only observing it.

Within this landscape, the literature has advanced in three successive waves.
\begin{itemize}
  \item \textbf{Verifying behaviors in hand-built worlds.} Generative Agents~\citep{park2023generative} placed 25 LLM-driven characters in a hand-built village and demonstrated for the first time that LLM agents could spontaneously spread information, form relationships, and coordinate collective activities. Subsequent systems carried this evidence to social-network and macroeconomic settings~\citep{gao2023s3,li2023econagent}, and Concordia~\citep{vezhnevets2023concordia} and OASIS~\citep{yang2024oasis} recast the hand-built sandbox as a configurable framework, the latter scaling to one million agents. In this wave, the run itself remains closed: its trajectory is not an object of study, and nothing can be intervened on once it has started.
  \item \textbf{Alignment with real societies.} A second wave anchored simulation in real data. Within predefined scenarios, HiSim~\citep{mou2024unveiling} aligned simulated discourse with real social-media contexts and Generative Agent--1000~\citep{park2024generative1000} constructed agents from interviews with 1,000 real individuals. Among configurable platforms, AgentSociety~\citep{piao2025agentsociety} embedded 10k agents in realistic urban infrastructure and \svone~\citep{zhang2025socioverse} aligned simulated populations to census marginals over a large real-user pool. This wave improves the simulation fidelity, but the comparison it draws is cross-sectional. Real data also enters as a dataset assembled once per study, so changing the population or the environment means collecting it again.
  \item \textbf{Research automation by agentic workflows.} The most recent wave hands off the research process itself to agents. YuLan-OneSim~\citep{wang2025yulan} pairs code-free, agent-operated scenario construction that takes a topic and finishes the loop on its own, from environment construction through report review. AgentSociety~2~\citep{piao2026agentsociety2} couples AI social scientists and silicon participants in one runtime. The autonomous agents in this wave turn to actively execute to solve researchers' questions, while the research purpose is fixed once initialized. The lack of human steering restricts agents' global view of the research scope. The researcher is left with gated approval over what the agent exposes, not full control of the edit.
\end{itemize}

\sysname supplies the missing capability, offering dynamic longitudinal simulation operated through a human-AI coevolutionary paradigm.
\sysname complements end-to-end research automation via intervention and counterfactual experiments within the simulation and human-controllable loop outside the research process. On the other hand, \sysname strengthens the configured framework via lifecycle agentic infrastructure and data MCPs. Standing at the intersection of professional human social scientists' expertise and efficient agents' automation, we believe \sysname builds a promising and powerful paradigm for the next era of social simulation.

\begin{figure}[t]
  \centering
  \resizebox{\textwidth}{!}{\input{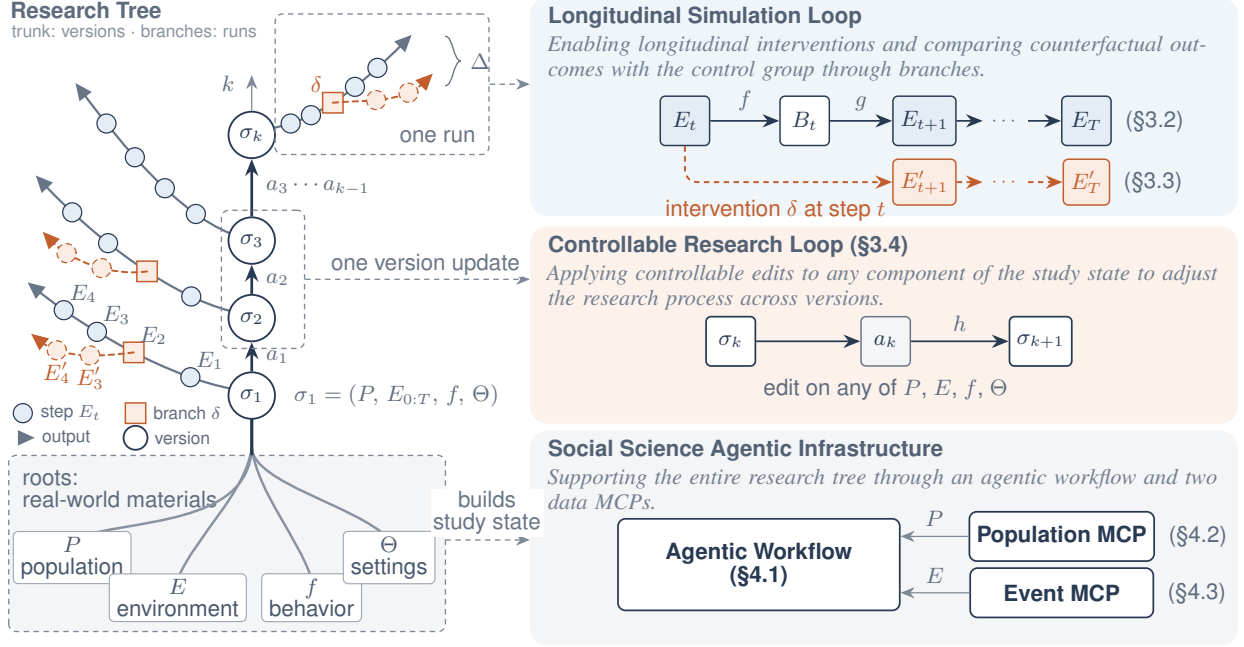}}
  \caption{An illustration of the \sysname{} framework, depicted as a research tree: overall pipeline shown on the left and detailed expansion on the right. From bottom to the top, the \textbf{agentic infrastructure} provides the research workflow and data to support the system, with initiated study state denoted as $\sigma_1$ consisting $P$, $E$, $f$, and $\Theta$; then the state will be continually updated to $\sigma_k$ for whole system evolvement, which is conducted through the component edits $\alpha_k$ generated in \textbf{controllable research loop}; in each updated version $\sigma_k$, we have the \textbf{longitudinal simultation loop}, in which we conduct counterctual thinking to derive multiple branches $E$ and $E'$ via the intervention $\delta$ at step $t^*$. Human researchers can engage in both loops to provide counterfactual design and system-refinement guidance.}
  \label{fig:architecture}
\end{figure}

\section{\sysname Human-AI Co-evolutionary Paradigm}
\label{sec:framework}

\Cref{sec:framework} and \Cref{sec:infra} present the \sysname framework as two loops and one agentic infrastructure, revealing three levels of granularity in social science research: from one step to one branch, and finally the whole tree of runs that a study focuses on, as shown in \Cref{fig:architecture}. We first formalize the necessary preliminaries in \S\ref{sec:framework:overview}, then we detail every component. Typically, given a research question, the researcher and the social science agent collaborate through \textbf{a longitudinal simulation loop} and \textbf{a controllable research loop}, supported by \textbf{social science agentic infrastructure}. The longitudinal simulation loop directly runs the target simulation experiment step by step through iterations between simulated behaviors and the environment (\S\ref{sec:framework:behavior}), and enables interventions and comparison of counterfactual outcomes with the control group through branches (\S\ref{sec:framework:loop}). The controllable research loop manages different versions throughout the research process by applying edits to any component of the study state (\S\ref{sec:framework:control}).

\subsection{Overview and Formalization}
\label{sec:framework:overview}

\paragraph{Theoretical Foundation.}
Lewin's field theory~\citep{lewin1936principles} expresses a foundational insight of social psychology: behavior arises jointly from the person and the environment.
We adopt the theory here for a practical formulation: it separates the population and environment and therefore provides a compact expression for both the passage of time and an introduced intervention.
\svone~\citep{zhang2025socioverse} realized the relation as alignment-centered components.
\sysname develops this line from single decisions into a social process: behavior computed at one step reshapes the environment that conditions the next, and the researcher's interventions enter the same environment and propagate through the same loop. Outside the loop, the social process itself is editable under the guidance of human researchers.

\paragraph{The \sysname Abstraction.}
\sysname condenses social simulation into
\begin{equation}
  \label{eq:bfpe}
  B = f(P,\, E),
\end{equation}
where $P$ is the population, $E$ the environment, $f$ the behavior engine, and $B$ the resulting behavior.
The abstraction is powerful precisely because its three terms are {independently replaceable}: $P$ may be drawn from a real-world user pool or synthesized from census marginals. $E$ may be grounded in live event streams or designed for a stylized experiment. $f$ may be a classical ABM rule, an LLM prompt chain, a learned reinforcement-learning policy, or a hybrid of these.
Any social simulation scenario, from Schelling segregation to macro-economic nowcasting, can be expressed as an instantiation of \Cref{eq:bfpe}. Consequently, an intervention can be viewed as an operation on $E$ with $P$ and $f$ held fixed, and a controlled adjustment can be viewed as modifying any components among \(B,f,P\) and $E$.

\paragraph{Longitudinal Simulation Loop.}
\sysname formulates the inner simulation and outer research process as two loops.
Within a simulation run, the environment carries the consequences of behavior forward: the population acts on the environment of step~$t$, and the behaviors it produces give the environment of step~$t+1$.
Then
\begin{equation}
  \label{eq:inner-loops}
  \underbrace{E_{t+1} = g\bigl(E_t,\; B_t\bigr)}_{\text{simulation loop, step } t}
\end{equation}
where $g$ is the update function of $E$. \S\ref{sec:framework:behavior} details the simulation loop in one step, from $E_t$ to $E_{t+1}$, and \S\ref{sec:framework:loop} further explains the intervention on the simulation loop, branching  $E'_{t+1}$ based on $E_t$.

\paragraph{Controllable Research Loop.}
Across runs, the research process itself moves in the same shape.
Let $\sigma = (P,\, E_{0:T},\, f,\, \Theta)$ denote the \emph{study state}, which implies the current version snapshot of the whole system during the research. $\Theta$ denotes the collection of the metrics, the resource manifest, and the seed.
Then
\begin{equation}
  \label{eq:outer-loops}
  \underbrace{\sigma_{k+1} = h\bigl(a_k,\;\sigma_k\bigr)}_{\text{research loop, version } k}
\end{equation}
where $a_k$ is an \emph{edit} on the study state and $h$ is the mapping function that applies it, producing the study state of the next version.
The edit can be proposed by either the agentic workflow or human researchers, and \S\ref{sec:framework:control} expands the research loop through versions, from $\sigma_k$ to $\sigma_{k+1}$. \sysname enables the human-AI co-evolution through these two loops under a unified iterative update form.

\paragraph{Social Science Agentic Infrastructure.}

In \sysname, the population $P$ is the set of agents (or silicon samples) whose behaviors the simulation tracks. Generally, a population is fixed with the initialization of the study to provide persistent subjects for the social simulation. The environment $E$ is the sum of information external to the agents that shapes their behavior. Generally, the environment is orchestrated along two orthogonal axes: modality and scope. Modality separates the \emph{physical} world from the \emph{information} world (i.e., geography and urban facilities v.s. news and policies), while scope separates \emph{macro} context from \emph{micro} context (i.e., broadcast v.s. direct message). Both $P$ and $E$ are assembled by a skill-based agentic research pipeline (\S\ref{sec:infra:pipeline}), where real-world data is acquired via standard MCPs across studies (\S\ref{sec:infra:pool} and \S\ref{sec:infra:signals}).

\subsection{The Longitudinal Simulation Loop}
\label{sec:framework:behavior}

This subsection answers how $E_t$ evolves into $E_{t+1}$. We expand the simulation loop of \Cref{eq:inner-loops} into detailed steps that the engine actually executes.
Let $P = \{p_1, \dots, p_N\}$ be the fixed population and $\{E_t\}_{t=0}^{T}$ the environment sequence.
At each step, the environment first evolves exogenously, then each agent observes the context, decides, and generates the action:
\begin{equation}
  \label{eq:loop}
  e_t^{i} = \omega(E_t,\, p_i), \qquad
  b_{i,t} = f\bigl(p_i,\; e_t^{i},\; m_{i,t}\bigr), \qquad
  B_t = \{b_{i,t}\}_{i=1}^{N}, \qquad
  E_{t+1} = g\bigl(E_t,\; B_t\bigr),
\end{equation}
where $\omega$ is the observation operator that projects the global environment onto agent~$i$. The term $m_{i,t} = \mu(b_{i,t-W:t-1})$ is the agent's bounded memory of its own recent actions, $B_t$ collects the step's individual behaviors, and $g$ updates them back into the environment.
The engine executes \Cref{eq:loop} as a fixed sequence of module calls, each with declared inputs and outputs.
\Cref{fig:mainloop} shows the data that moves along the five phases of one step with an example.

\subsubsection{Overall Step Sequence}
\label{sec:framework:behavior:steps}

At step 0, the population provider materializes $P$ from the population bundle under the declared seed, assigning each agent an identifier that is never reused, and the environment provider builds $E_0$. The initial states are recorded as the first panel rows, and no decision is computed. 
Every later step $t$ runs five phases in a fixed order.

\begin{figure}[t]
  \centering
  \resizebox{\textwidth}{!}{\input{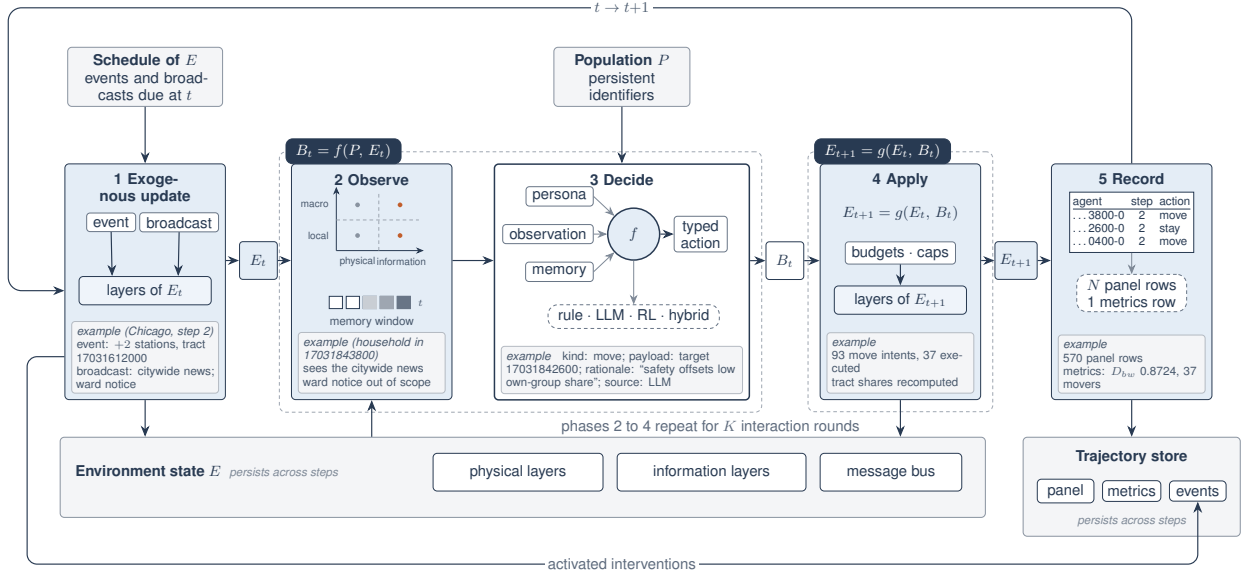}}
  \caption{An illustration of the mechanism within one step of the longitudinal simulation loop. (1) The environment's schedule update events and broadcast for the current step $E_t$; (2) each agent observes the environment along modality and scope axes; (3) the agent then decides a typed action through the behavior function given the population features, observation, and its memory; (4) given current collective behavior $B_t$, the environment is updated for the next step $E_{t+1}$; (5) all the information within a step is recorded as a trajectory during the research.}
  \label{fig:mainloop}
\end{figure}

We take the Chicago segregation case in \S\ref{sec:cases:chicago} as an example here, which studies households living in 781 census tracts in Chicago. Each household decides at every step whether to stay or move to another tract, and the population composition of the tracts changes as households move. Examples in \Cref{fig:mainloop} show details from $E_2$ to $E_3$ in this study.

\begin{enumerate}
  \item \textbf{Exogenous update.} The environment provider reads the scheduled events and broadcasts whose trigger step equals $t$, applies each event to its physical layer, activates each broadcast for its time-to-live, and returns the list of activated items. An intervention enters here as an additional event or broadcast and is processed by the same call as the study's regular schedule. 
  In the example, step~2 activates a scheduled event that adds two transit stations to tract 17031612000, together with a citywide news broadcast and a ward notice addressed to one district.
  \item \textbf{Observe.} For every agent, the environment provider assembles one observation along modality and scope axes. 
  In the example, households in tract 17031843800 receive the citywide news in their macro-information field, while the ward notice is addressed to another district, which belongs to other's local-information field and does not appear in their observations.
  \item \textbf{Decide.} The decision model receives the batch of observations together with each agent's memory window and returns one action per agent: an action type (move, stay, post, etc.), a payload with the type-specific content (a destination, a text, a number, etc.), a first-person rationale, and a source label stating whether the decision came from an LLM, a rule, a fallback, or a replay. In the example, the household's action is a ``move'', whose payload names the target tract 17031842600, and its rationale reads ``safety offsets low own-group share''.
  \item \textbf{Apply.} The environment provider collects the actions into the layer state, which realizes $E_{t+1} = g(E_t, B_t)$. All actions within a round are updated synchronously at the end of the round. {In the example, 93 households intend to move at $E_2$, 37 moves are executed, and the racial shares of the affected tracts are recomputed, which yields $E_{3}$.}
  \item \textbf{Record.} The engine appends one panel row per agent, one metrics row computed by the study's metric collector, and pushes each agent's final action into its memory window. {In the example, $E_2$ records 570 panel rows, including one row per household and one metrics row logging the Black--White dissimilarity index $D_{bw}=0.8724$ and movers.}
\end{enumerate}

\subsubsection{Implementations of a Step}
\label{sec:framework:behavior:components}
The three components below detail the implementations and configurations in the longitudinal simulation loop.

\paragraph{Persona Declarations.}
Beyond identity, each persona specifies three study-relevant properties: (1) a cohort label that groups demographically similar agents for efficient batched decision-making (e.g., an archetype representing an ethnic group), (2) an optional sampling weight when a small agent set represents a larger population, and (3) an initial state that seeds its situation at step~0 (a home tract, an initial opinion, a starting budget). 
A study also declares the social connection among its agents, through spatial adjacency, an explicit social network, or no interaction at all, and the propagation of influence along those connections, from fully independent decisions to contagion-style local spread to broadcast-then-discussion patterns typical of social media. 
{Messages between connected agents travel through a mediated bus and are delivered by neighbor lookup, so an agent's local-information field carries the posts of its network neighbors from earlier rounds of the current step and from the previous step.}

\paragraph{Memory Window.}
Each agent owns a bounded window of its own recent actions, eight steps by default and configurable per study. Because the window is filled from the actions that the panel records, it can be rebuilt exactly from the panel table, allowing a branched run to inherit its agents' histories.

\paragraph{Behavior Function.}
The decision model is the only component that computes $f$, and {because every implementation returns the typed action of phase 3,} the environment and the record store are independent of the mechanism that produced a decision.
Four behavior functions are implemented.
\emph{Rule-$f$} evaluates a classical update rule and serves as the parity baseline.
\emph{LLM-$f$} composes the persona, the four observation fields, and the memory window into a prompt, parses the structured reply into an action.
\emph{RL-$f$} maps observations to actions through a trained policy, one per class of strategic actor, and enters the workspace through an adapter.
\emph{Hybrid-$f$} assigns different implementations to different agent groups within one run, typically LLM-$f$ for a small core and rule-$f$ for the remainder.
The typed-action contract is also what makes the rule-versus-LLM comparison possible, since both implementations act in the same discrete action space. 

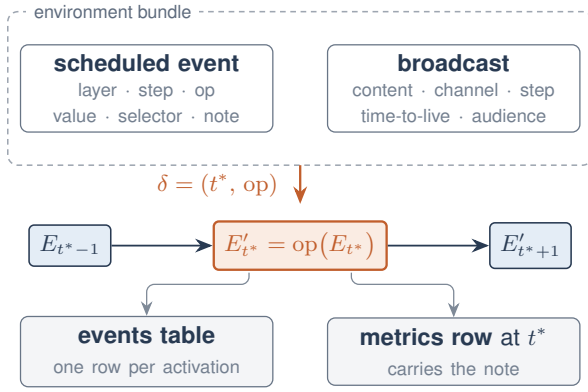
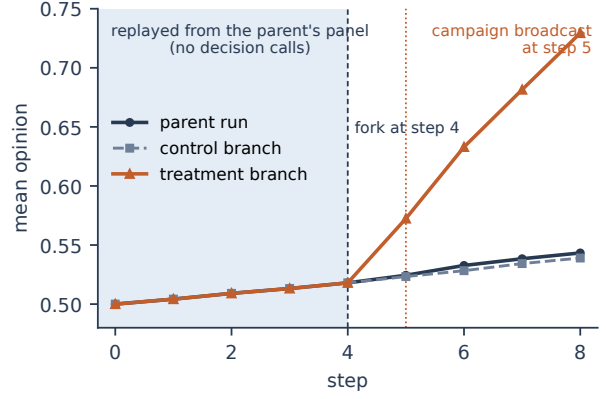
\begin{figure}[t]
  \centering
  \begin{subfigure}[b]{0.475\linewidth}
    \centering
    \resizebox{\linewidth}{!}{\begin{tikzpicture}[
  bundleframe/.style={draw=figink!45, line width=0.6pt, figdashed, rounded corners=4pt},
  rec/.style={figfont, font=\sffamily\footnotesize, draw=figink!70, line width=0.5pt,
              rounded corners=3pt, fill=white, text width=31mm, inner xsep=5pt,
              inner ysep=4pt, align=center},
  Enode/.style={figfont, font=\sffamily\footnotesize, draw=figink, line width=0.6pt,
                rounded corners=2pt, fill=figblue, inner sep=4pt, align=center},
  opnode/.style={figfont, font=\sffamily\footnotesize, draw=figaccent, line width=0.9pt,
                 rounded corners=2pt, fill=figaccentfill, text=figaccent,
                 inner sep=4pt, align=center},
  store/.style={figfont, font=\sffamily\footnotesize, draw=figink!70, line width=0.5pt,
                rounded corners=3pt, fill=figgray, text width=31mm, inner xsep=5pt,
                inner ysep=4pt, align=center},
  note/.style={figsub, font=\sffamily\scriptsize, text=figink!70, align=center},
]
  \draw[bundleframe] (0.30,4.05) rectangle (8.30,6.15);
  \node[figsub, font=\sffamily\scriptsize, anchor=west, fill=white, inner xsep=3pt]
    at (0.55,6.15) {environment bundle};
  \node[rec] (ev) at (2.20,5.10)
    {\textbf{scheduled event}\\[1.5pt]{\sffamily\scriptsize\color{figink!65}layer $\cdot$ step $\cdot$ op\\value $\cdot$ selector $\cdot$ note}};
  \node[rec] (bc) at (6.40,5.10)
    {\textbf{broadcast}\\[1.5pt]{\sffamily\scriptsize\color{figink!65}content $\cdot$ channel $\cdot$ step\\time-to-live $\cdot$ audience}};

  \draw[figarrowaccent, line width=0.9pt] (4.30,4.05) -- (4.30,3.52);
  \node[figfont, font=\sffamily\footnotesize, text=figaccent, anchor=east] at (4.12,3.78)
    {$\delta = (t^{*},\, \mathrm{op})$};

  \node[Enode] (Epre)  at (1.15,2.95) {$E_{t^{*}-1}$};
  \node[opnode] (Eop)  at (4.30,2.95) {$E'_{t^{*}} = \mathrm{op}\bigl(E_{t^{*}}\bigr)$};
  \node[Enode] (Epost) at (7.45,2.95) {$E'_{t^{*}+1}$};
  \draw[figarrow] (Epre) -- (Eop);
  \draw[figarrow] (Eop) -- (Epost);

  \node[store] (evt) at (2.20,1.50)
    {\textbf{events table}\\[1.5pt]{\sffamily\scriptsize\color{figink!65}one row per activation}};
  \node[store] (met) at (6.40,1.50)
    {\textbf{metrics row} at $t^{*}$\\[1.5pt]{\sffamily\scriptsize\color{figink!65}carries the note}};
  \draw[figarrowlight, rounded corners=3pt] (3.60,2.58) -- (3.60,2.42) -- (2.20,2.42) -- (evt.north);
  \draw[figarrowlight, rounded corners=3pt] (5.00,2.58) -- (5.00,2.42) -- (6.40,2.42) -- (met.north);

\end{tikzpicture}}
    \caption{A declared intervention entering a run}
    \label{fig:fork-demo:decl}
  \end{subfigure}\hfill
  \begin{subfigure}[b]{0.495\linewidth}
    \centering
    \includegraphics[width=\linewidth]{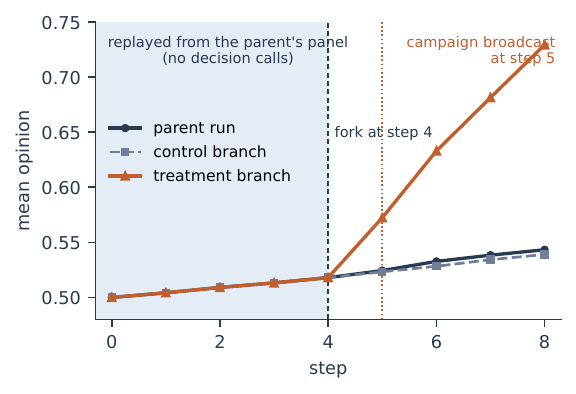}
    \caption{Replay-based branching in an opinion-diffusion study}
    \label{fig:fork-demo:traj}
  \end{subfigure}
  \caption{(a)~An intervention is a scheduled event or a broadcast in the environment bundle, activated in the exogenous phase of step~$t^{*}$. (b)~A branch is created when an intervention is activated. The example branch inherits its parent's first four steps by replay, without decision calls, and diverges afterward. The treatment branch adds a media campaign at step~5, while the control branch remains the same.}
  \label{fig:fork-demo}
\end{figure}

\subsection{Intervention and Counterfactual Branching in Simulation Loop}
\label{sec:framework:loop}

This subsection answers how $E_t$ branches into $E'_{t+1}$ given intervention $\delta$. We describe what an intervention does to the simulation loop and how a counterfactual experiment becomes computable.
A longitudinal study is rarely a single run: the researcher declares an intervention, compares the intervened run with the controlled run, and returns to the study with a revised question.
Two mechanisms carry that comparison, both operating on the artifacts of the previous subsection: an intervention is an entry in the environment bundle that the exogenous-update phase activates at its step, so it is declarative and auditable, and a \emph{branch} is a run that inherits a parent's recorded rows through replay.

{An intervention $\delta = (t^{*},\, \mathrm{op})$ names a fork step $t^{*}$ and an operation $\mathrm{op}$. The branch shares its parent's history through step $t^{*}{-}1$, the exogenous phase of step $t^{*}$ applies $\mathrm{op}$ to the environment, and the loop then carries the change forward, which is the simulation counterpart of an intervention on the environment with $P$ and $f$ held fixed. Primed symbols denote the intervened branch, as in \Cref{fig:architecture}:}
\begin{equation}
  \label{eq:intervene}
  E'_{t} = E_{t}\;\; (t < t^{*}), \qquad
  E'_{t} = \mathrm{op}\bigl(E_t\bigr)\;\;(t = t^{*}), \qquad
  E'_{t+1} = g\bigl(E'_{t},\, B'_{t}\bigr)\;\; (t > t^{*}).
\end{equation}
A \emph{branch} of a parent run at fork step $t^{*}$ keeps the parent's realized behavior up to $t^{*}$ and computes only what follows:
\begin{equation}
  \label{eq:branch}
  b'_{i,t} =
  \begin{cases}
    b_{i,t}, & t < t^{*} \quad \text{(replayed)},\\[2pt]
    f\bigl(p_i,\; \omega(E'_{t}, p_i),\; m'_{i,t}\bigr), & t \geq t^{*},
  \end{cases}
\end{equation}
{where $B'_t = \{b'_{i,t}\}_{i=1}^{N}$ and the memory window $m'_{i,t}$ are built from the branch's own actions after $t^{*}$.}
Because $P$, $f$, and the seed are the same objects in both branches and their histories are identical up to $t^{*}$, the difference between an intervened branch and a control branch is attributable to $\delta$ alone:
\begin{equation}
  \label{eq:contrast}
  \Delta_t(\delta) \;=\; Y\bigl(E'_{t}\bigr) - Y\bigl(E_{t}\bigr), \qquad t \geq t^{*},
\end{equation}
{where $Y$ is any quantity measured from the environment or the panel at step $t$, and $E_t$ is the control branch, which shares the same history through $\{0:t^*{-}1\}$ and receives no operation.}
\Cref{eq:contrast} enables the counterfactual design that reality never supplies or hypotheses during theory verifications.

\paragraph{Declared Interventions.}
An \emph{intervention} is an entry in the environment bundle, implemented in two types: a scheduled event or a broadcast (\Cref{fig:fork-demo:decl}).
A scheduled event names a physical layer, a step, a numeric operation, a value, and a selector for the affected units, together with a note. A {broadcast} carries an information content, a channel, a step, a time-to-live, and an audience that is either everyone or a selector.
An intervention is verified to ensure every event targets a declared layer, and every broadcast requires a declared information layer.
Both kinds activate in the exogenous phase of their step, before any agent observes, and the engine writes each activation to the events table and attaches the notes to that step's metrics row, so the change in a metric series is read against before the intervention.

\paragraph{Replay-based Branching.}
A branch is a new run that inherits a parent run's first $t^{*}{-}1$ steps and diverges afterward.
The engine produces it by replay: it materializes $P$ and $E_0$ from the same bundles and seed, then for steps $1$ to $t^{*}{-}1$ activates the parent's schedule, reads the parent's stored actions for that step from its panel table, applies them to the environment with the source label {replay}, pushes them into the agents' memory windows, and records the resulting rows. From step $t^{*}$, it calls the decision model as usual.
Because the environment transition of a study is a function of the recorded actions, the replayed rows reproduce the parent's rows exactly. The environment reaches the same $E_{t^{*}{-1}}$, and each agent resumes with the same memory window it had in the parent.
Two branches forked from the same parent therefore share an identical realized history, and everything that differs between them after step $t^{*}$ is attributable to the schedules declared for the steps after the fork.
\Cref{fig:fork-demo} shows the mechanism in an opinion-diffusion study: a parent run of eight steps is forked at step~4 into a control branch with no further intervention and a treatment branch whose bundle adds a media campaign at step~5. The two branches reproduce the parent's rows for steps 0 to 4 exactly, invoke the decision model only for steps 5 to 8, and diverge from step~5.
The same mechanism extends a finished run to a longer horizon at no cost for the inherited steps and resumes an interrupted run from its last completed step.

\subsection{The Controllable Research Loop}
\label{sec:framework:control}

This subsection answers how to control the study state version, from $a_k$ to $a_{k+1}$. We step outside the simulation loop, extending to the research loop in which the research itself is updated.
The \Cref{eq:outer-loops} carries two objects, an edit action $a_k$ on the study state and the map $h$ that applies it.
The research agent may raise an edit $a_k$ from the reasoning over last trajectory. The researcher may raise it directly as well on any artifact and at any stage of the study via either verbalized instructions, additional materials, or direct code modifications.
What the researcher holds is therefore not a veto over agent proposals but authorship of the edit itself, and the loop is \emph{controllable} in that stronger sense: the research process is editable at every point, not merely approvable at the points an agent chooses to expose.

Let an edit be a pair $a_k = (c,\, u)$, where $c \subseteq \{P, E, f, \Theta\}$ names the components of $\sigma$ it modifies and $u$ is the update applied to them, and let $\sigma[\,c \leftarrow u\,]$ denote $\sigma$ with those components replaced. Then three cases exhaust $h$:
\begin{equation}
  \label{eq:edit}
  \sigma_{k+1} = h\bigl(\sigma_k,\, (c, u)\bigr) =
  \begin{cases}
    \sigma_k, & c = \varnothing,\;\text{unchanged}\\[3pt]
    \sigma_k\bigl[\,E_{t^{*}:T} \leftarrow u\,\bigr], & c = \{E\},\;\text{branch}\\[3pt]
    \sigma_k\bigl[\,c \leftarrow u\,\bigr], & \text{otherwise},\;\text{version}
  \end{cases}
\end{equation}
which are respectively no change, a branch at fork step $t^{*}$, and a new version.
The first case keeps the loop reversible. As every version snapshots its entire workspace, $\sigma_k$ is always traceable, so declining an edit or undoing one already applied can be realized at every stage.

\begin{figure*}[t]
  \centering
  \resizebox{\textwidth}{!}{\begin{tikzpicture}[
  ztitle/.style={font=\sffamily\small\bfseries, text=figink!75},
  art/.style={figfont, font=\sffamily\footnotesize, draw=figink!55, line width=0.5pt,
              rounded corners=2.5pt, fill=white, text width=41mm, inner xsep=5pt,
              inner ysep=4.5pt, align=left},
  arte/.style={art, draw=figaccent, line width=0.8pt, fill=figaccentfill},
  ver/.style={figfont, font=\sffamily\footnotesize, draw=figink, line width=0.6pt,
              rounded corners=3pt, fill=white, text width=31mm, inner xsep=4pt,
              inner ysep=3.5pt, align=center},
  vere/.style={ver, draw=figaccent, fill=figaccentfill, text=figaccent},
  move/.style={figfont, font=\sffamily\scriptsize, draw=figink!55, line width=0.45pt,
               rounded corners=2.5pt, fill=white, text width=30mm, inner xsep=4pt,
               inner ysep=3.5pt, align=center},
  elab/.style={font=\sffamily\scriptsize, text=figink!70, inner sep=1.5pt, fill=white},
  elaba/.style={elab, text=figaccent},
]
\newcommand{\ckpt}[2]{%
  \filldraw[draw=figaccent, line width=0.6pt, fill=white] (#1,#2) circle (1.7mm);
  \draw[draw=figaccent, line width=0.7pt, line cap=round]
    (#1-0.075,#2+0.005) -- (#1-0.02,#2-0.055) -- (#1+0.085,#2+0.06);}

  \begin{scope}[on background layer]
    \fill[figgray, rounded corners=6pt] (0.35,2.55) rectangle (5.85,7.75);
  \end{scope}
  \node[ztitle, anchor=north west] at (0.60,7.55) {Study State $\sigma_k$};

  \node[art] (aP) at (3.10,6.65) {$P$\;\; population bundle};
  \node[art] (aE) at (3.10,5.55) {$E_{0:T}$\;\; environment bundle and schedule};
  \node[art] (af) at (3.10,4.45) {$f$\;\; model module};
  \node[art] (aT) at (3.10,3.35) {$\Theta$\;\; metrics, resources, seed};

  \node[ztitle, anchor=north west] at (6.20,7.55) {Study Tree};
  \node[figsub, anchor=north west] at (6.22,7.18)
    {\scriptsize each node is a study state $\sigma$};

  \node[ver]  (v0) at (7.75,5.20) {$\sigma_1$ baseline\\[-1pt]{\tiny 8 steps, no intervention}};
  \node[ver]  (v1) at (12.10,6.40) {$\sigma_2$ control branch\\[-1pt]{\tiny replayed to $K$}};
  \node[ver]  (v2) at (12.10,4.80) {$\sigma_3$ treatment branch\\[-1pt]{\tiny campaign $\delta$ at $K{+}1$}};
  \node[vere] (v3) at (12.10,3.25) {$\sigma_4$ pool replaced\\[-1pt]{\tiny same $E$, same $f$}};
  \node[vere] (v4) at (16.85,6.40) {$\sigma_5$ longer horizon\\[-1pt]{\tiny extends $\sigma_2$ to $2T$}};
  \node[vere] (v5) at (16.85,4.80) {$\sigma_6$ rule-$f$ ablation\\[-1pt]{\tiny same $P$, same $E$}};

  \draw[figarrow] (v0.east) -- (v1.west);
  \draw[figarrow] (v0.east) -- (v2.west);
  \draw[figarrowaccent, figdashed, rounded corners=4pt] (v0.south) -- (7.75,3.25) -- (v3.west);
  \draw[figarrowaccent, figdashed] (v1.east) -- (v4.west);
  \draw[figarrowaccent, figdashed] (v2.east) -- (v5.west);

  \ckpt{10.00}{5.92}\ckpt{10.00}{5.02}\ckpt{9.20}{3.25}
  \ckpt{15.00}{6.40}\ckpt{15.00}{4.80}

  \node[elab,  anchor=south] at (10.05,6.06) {\scriptsize branch};
  \node[elab,  anchor=north] at (10.05,4.86) {\scriptsize branch $+\,\delta$};
  \node[elaba, anchor=south] at (9.20,3.41) {\scriptsize new version};

  \draw[{Stealth[length=1.9mm,width=1.6mm]}-{Stealth[length=1.9mm,width=1.6mm]},
        line width=0.55pt, draw=figink!65, figdashed] (14.35,6.08) -- (14.35,5.14);
  \node[elab, rotate=90, anchor=south] at (14.21,5.60) {\scriptsize paired contrast};

  \draw[figarrow] (6.25,2.30) -- (6.85,2.30);
  \node[figsub, anchor=west] at (6.95,2.30) {\scriptsize branch: inherits the parent's history};
  \draw[figarrowaccent, figdashed] (11.35,2.30) -- (11.95,2.30);
  \node[figsub, anchor=west] at (12.05,2.30) {\scriptsize version: a fresh run from edited artifacts};
  \ckpt{17.30}{2.30}
  \node[figsub, anchor=west] at (17.55,2.30) {\scriptsize edit applied here};

  \begin{scope}[on background layer]
    \fill[figblue!45, rounded corners=6pt] (0.35,0.25) rectangle (19.05,1.60);
  \end{scope}
  \node[ztitle, anchor=west, align=left] at (0.60,0.92) {Applicable Edit $a_k$};
  \node[move] (m1) at (6.40,0.92) {replace the population\\[-1pt]{\tiny changes $P$}};
  \node[move] (m2) at (9.90,0.92) {vary the schedule\\[-1pt]{\tiny changes $E$}};
  \node[move] (m3) at (13.40,0.92) {swap the behavior function\\[-1pt]{\tiny changes $f$}};
  \node[move] (m4) at (16.90,0.92) {repin the data sources\\[-1pt]{\tiny changes $\Theta$}};

\end{tikzpicture}}
  \caption{The controllable research loop as a tree of versions. An edit $a_k$ may be raised by the researcher or by an agent and reaches any component of the study state, and every version snapshots its whole workspace, so the parent is always restorable.}
  \label{fig:studytree}
\end{figure*}
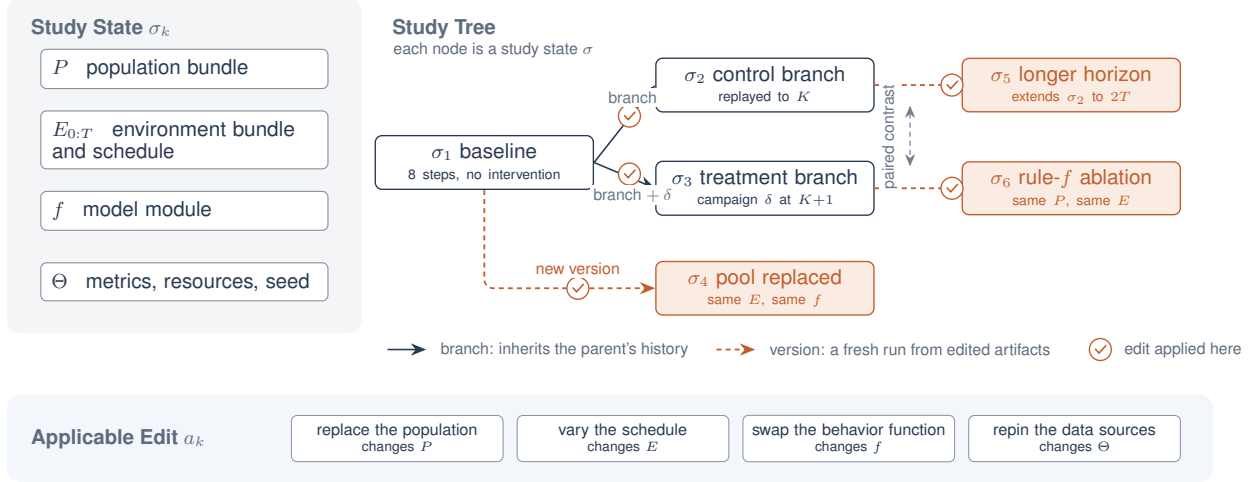

\paragraph{Versions.}
Every branch, extension, or re-run is a version of the same study (\Cref{fig:studytree}).
A version manifest in the study workspace lists each version with its identifier, parent version, note, creation time, and, for branches, the source version and fork step. Creating a version snapshots the complete workspace of the current version, including the code, bundles, trajectory, and reports, so that earlier results remain inspectable and re-runnable.
The \svskill{iterate} skill (\S\ref{sec:infra:pipeline}) creates versions on request and re-enters the workflow at the earliest stage a change affects. Notably, the branch differs from the version in the edit domain. A branch changes $E$ alone, holding $P$, $f$, and the seed fixed, and inheriting the parent's realized history, which belongs precisely to the inner longitudinal simulation loop. In contrast, a \emph{version} may change any artifact arbitrarily, so it does not inherit a history. Instead, it supports comparing one study design with another.

\paragraph{Applicable Controls.}
Because $P$, $E$, and $f$ are separate replaceable artifacts and every run is a version with a recorded lineage, a study can grow by adding versions that vary one axis while holding the others fixed:
\begin{itemize}
  \item \textbf{Population replacement.} The same environment and behavior function can be applied to a different demographic composition (Census-synthesized U.S.\ adults versus personas from an open pool), testing whether findings generalize across populations.
  \item \textbf{Environment variation.} The same population and behavior function can be exposed to fundamentally different environments (changing cities or interaction topologies for the same target groups), enabling controlled comparisons of environmental effects that can not be realized by event or broadcasting interventions.
  \item \textbf{Behavior-function ablation.} Applying rule-$f$, LLM-$f$, and hybrid-$f$ on the same $(P, E)$ pair can isolating the decision model.
  \item \textbf{Data-source substitution.} A different vintage of economic indicators or a different user pool through the resource manifest alone (\Cref{sec:infra}), with no change to model code.
\end{itemize}
These designs can be combined together and applied in one version to provide a comprehensive result.
All versions store the records under the same contract, so a per-step metric or a per-agent trajectory is compared across versions with one query.

\paragraph{Human-AI Co-evolutionary Paradigm.} Through the longitudinal simulation loop and controllable research loop, \sysname provides a paradigm in which human social scientists and autonomous agents collaborate and evolve with the research progression. This {human-AI co-evolutionary} paradigm sits in the middle of the landscape of \Cref{sec:background} and draws on both of its ends at once. Autonomous agents get professional and experienced insights from human experts, filling the gap in transformational ability. Human researchers benefit from the efficient agentic infrastructure and generative agent-based simulation platforms, realizing innovation in research tools.
We expect the human-AI {co-evolutionary} paradigm rather than either extreme to carry the next generation of social science simulation.

\section{Social Science Agentic Infrastructure}
\label{sec:infra}

\Cref{sec:framework} defines what a study \emph{is}. This section describes what it takes to run one, and it is the substance of our third contribution: the social science agentic infrastructure that carries both loops of \Cref{sec:framework}. It materializes a study state $\sigma$ from real data, executes the longitudinal simulation loop on it, and gives the controllable research loop the named points at which an edit $a_k$ is raised, applied by $h$, and recorded.
It is organized in three layers.
The \textbf{workflow layer} decomposes the research lifecycle into composable skills, each of which reads the artifacts of earlier stages, writes one validated artifact of its own, and pauses for a researcher decision, so that the controllable research loop of \S\ref{sec:framework:control} is realized at named points rather than in principle (\S\ref{sec:infra:pipeline}).
The \textbf{service layer} supplies a population service that anchors \Pop to real people and a signal service that grounds \Env in real-world data, both exposed as typed model context protocol (MCP) tools rather than as datasets copied into a study (\S\ref{sec:infra:pool} and \S\ref{sec:infra:signals}).
The \textbf{governance layer} records what each stage did, which data it used, and what the researcher approved, so that a finished study can be audited and re-run (\S\ref{sec:infra:governance}).
The runtime, the two services, and the case configurations are released as open-source packages, and an online workbench drives the same skills from a browser for researchers who do not work at a terminal.

Keeping the services outside the framework is what makes both loops practical. Whenever the study changes, the agentic research pipeline can query MCPs instantly to build new environment orchestration or population distribution accordingly.
The population and environment sources therefore remain independently replaceable, the same typed interface serves both the internal skill pipeline and external agent ecosystems, and a third party can implement a compatible MCP server without modifying the framework.
The infrastructure is also where the co-evolutionary paradigm meet. Agents operate the skills and query the services, and the researcher's edits enter through the same typed artifacts and checkpoints, so that a change raised by either side is applied, validated, and recorded in one place (\S\ref{sec:infra:governance}).

\subsection{Agentic Research Pipeline}
\label{sec:infra:pipeline}

\Cref{sec:framework} defines the simulation engine. This subsection describes the workflow layer that operates it, and it is where the controllable research loop of \S\ref{sec:framework:control} becomes concrete: each skill is a stage at which an edit $a_k$ can be raised, and each checkpoint is a point at which the edit is applied with the researcher's knowledge.
\sysname takes a complementary approach compared to previous studies: the research lifecycle is decomposed into reusable skills, each of which reads the artifacts of earlier stages, writes one validated artifact of its own, and pauses for review, so that every intermediate result is inspectable and every decision point is recorded.

\paragraph{Skill Contracts.}
\Cref{tab:skill-io} lists the seven skills with their inputs, their output artifacts, and the gate that closes each stage.
Every artifact is a typed document validated on write and on read, so a document that fails validation cannot cross a stage boundary, and all artifacts of a study live in one workspace directory, which is therefore the complete state of the study.
\svskill{report} denotes the reporting stage as a whole: analysis of the trajectory store, literature grounding of the findings, and manuscript drafting, which the implementation realizes as separate skills under one entry point.

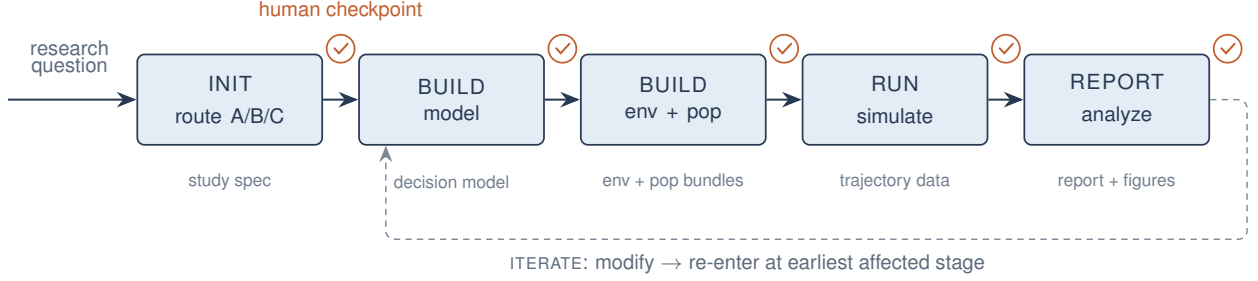
\begin{figure}[t]
  \centering
  \resizebox{\textwidth}{!}{\begin{tikzpicture}[skill/.style={fignode, text width=19mm, minimum height=11mm},
                    art/.style={figsub, font=\sffamily\tiny, text=figink!65}]
  \node[skill] (s1) at (0,0)    {\textsc{init}\\[1pt]{\scriptsize route A/B/C}};
  \node[skill] (s2) at (2.7,0)  {\textsc{build}\\[-1pt]{\scriptsize model}};
  \node[skill] (s3) at (5.4,0)  {\textsc{build}\\[-1pt]{\scriptsize env + pop}};
  \node[skill] (s4) at (8.1,0)  {\textsc{run}\\[1pt]{\scriptsize simulate}};
  \node[skill] (s5) at (10.8,0) {\textsc{report}\\[1pt]{\scriptsize analyze}};

  \node[art] at (0,-1.0)    {study spec};
  \node[art] at (2.7,-1.0)  {decision model};
  \node[art] at (5.4,-1.0)  {env + pop bundles};
  \node[art] at (8.1,-1.0)  {trajectory data};
  \node[art] at (10.8,-1.0) {report + figures};

  \draw[figarrow] (s1) -- (s2);
  \draw[figarrow] (s2) -- (s3);
  \draw[figarrow] (s3) -- (s4);
  \draw[figarrow] (s4) -- (s5);

  \newcommand{\ckpt}[2]{%
    \filldraw[draw=figaccent, line width=0.6pt, fill=white] (#1,#2) circle (1.7mm);
    \draw[draw=figaccent, line width=0.7pt, line cap=round]
      (#1-0.075,#2+0.005) -- (#1-0.02,#2-0.055) -- (#1+0.085,#2+0.06);}
  \foreach \x in {1.35,4.05,6.75,9.45,12.15}{\ckpt{\x}{0.62}}
  \node[figsub, text=figaccent] at (1.35,1.08) {\scriptsize human checkpoint};

  \draw[figarrow] (-2.7,0) -- (s1);
  \node[figsub, align=center] at (-1.95,0.5) {\scriptsize research\\[-2pt]\scriptsize question};

  \draw[figarrowlight, figdashed, rounded corners=3pt]
    (s5.east) -- ++(0.45,0) |- (1.9,-1.7) -- ([xshift=-0.8cm]s2.south);
  \node[figsub] at (6.3,-2.0)
    {\scriptsize \textsc{iterate}: modify $\to$ re-enter at earliest affected stage};
\end{tikzpicture}}
  \caption{The agentic research pipeline: composable skills each emit one validated artifact with a human checkpoint in between. \svskill{init} routes a query to reuse~(A), build-from-scratch~(B), or adapt-legacy~(C). \svskill{iterate} enables post-run refinement cycles.}
  \label{fig:pipeline}
\end{figure}

\begin{table}[t]
  \centering
  \caption{Skill contracts of the research pipeline: inputs, output artifacts, and the gate that closes each stage.}
  \label{tab:skill-io}
  \small
  \begin{tabular}{lp{4.6cm}p{4.6cm}l}
    \toprule
    Skill & Inputs & Outputs & Gate \\
    \midrule
    \svskill{init} & research query, study catalog, \eventtool & study definition, grounding record, resource manifest & intent confirmation \\
    \svskill{build-model} & study definition, reference studies & model module & assembly check \\
    \svskill{build-environment} & study definition, grounding record, \eventtool & environment bundle & schema validation \\
    \svskill{build-population} & study definition, environment bundle, \userpool & population bundle, roster & schema validation \\
    \svskill{run} & model module, environment and population bundles & simulation configuration, trajectory store & spend confirmation \\
    \svskill{report} & trajectory store, grounding record, literature service & report with figures, literature review, manuscript draft & researcher review \\
    \svskill{iterate} & change request, version manifest & new version, re-entry stage & version choice \\
    \bottomrule
  \end{tabular}
\end{table}

\paragraph{Checkpoints.}
Each skill pauses after writing its artifact and summarizes what it produced, and the researcher decides whether to advance by default. A researcher who prefers to run the whole pipeline can enable auto mode as well.
Four decisions are always put to the researcher: (1) confirmation of the intent brief before a study is created, (2) the choice between a new version, an in-place edit, and a branch before an existing study is changed, (3) the expected cost before a run that calls a language model, and (4) the use of any data service whose terms require per-study confirmation.
Every answer is logged with the study, so the audit trail records both what the pipeline did and what the researcher approved.

\paragraph{Three Onboarding Paths.}
When a user describes a research need, \svskill{init} consults the study catalog and routes the query along one of three paths:
\begin{itemize}
  \item \textbf{Path A (reuse/adjust).} The query closely matches an existing study.  The skill forks the matched study into a new study with its own version history, preserving the original as a read-only reference, and the user adjusts artifacts (population scale, intervention schedule) without writing code.
  \item \textbf{Path B (build from scratch).} No existing study matches.  \svskill{build-model} writes a model module that implements the engine's four component interfaces. The engine assembles it from the component registry.
  \item \textbf{Path C (adapt legacy).} The user has an existing simulator to integrate.  An engine-seam adapter wraps the legacy code with zero modification, projecting its internal state onto the \sysname artifacts.
\end{itemize}
Together the three paths form the framework's extensibility contract, covering parameter variation, new research topics, and incremental migration of existing simulation ecosystems.

\subsection{The Multi-Source Persona Pool and the \texorpdfstring{\userpool}{Population MCP}}
\label{sec:infra:pool}
On the population side, the \userpool is a multi-source aggregation service: it unifies five sub-pools, two internal real-user pools and three open persona pools, under one demographic label vocabulary. All the pools can be used aggregately to cover the research requirement.
The registered sources together span roughly one billion addressable persona records, of which 10{,}448{,}375 are locally indexed and queryable (\Cref{tab:pool-composition}).

\paragraph{Pool Composition.}
\Cref{tab:pool-composition} summarizes the five sub-pools.
The internal pools inherit the real-user asset introduced in \svone~\citep{zhang2025socioverse}: 973{,}928 profiles from X/Twitter and 9{,}158{,}404 from Xiaohongshu/RedNote, for a combined 10{,}132{,}332 records.  These profiles are tagged with demographic attributes through an LLM annotation pipeline with human review.
The preprocessed records are locally indexed for retrieval and sampling and serve as \emph{anchor pools} contributing real users with real behavioral text. The cross-pool field union reaches 40 queryable fields, of which roughly 20 belong to a canonical demographic vocabulary (age, gender, education, income, marital status, employment, religion, ideology, race/ethnicity, region, urbanicity, occupation, seniority, party, children, political engagement, and others).

\begin{table}[t]
  \centering
  \caption{Sub-pools registered in the \userpool, grouped by origin. ``Indexed'' counts locally indexed, queryable persona records. ``Fields'' counts structured demographic fields available for routing.}
  \label{tab:pool-composition}
  \small
  \setlength{\tabcolsep}{5pt}
  \begin{tabular}{llrr>{\raggedright\arraybackslash}p{5.6cm}}
    \toprule
    Category & Source & Indexed & Fields & Role \\
    \midrule
    \multirow{2}{*}{Internal real users}
      & \svone{} X/Twitter & 973{,}928 & 10 & anchor, real behavioral text \\
      & \svone{} RedNote & 9{,}158{,}404 & 8 & anchor, Chinese-language context \\
    \addlinespace
    \multirow{3}{*}{Open persona pools}
      & Nemotron-Personas USA & 59{,}994 & 7 & anchor / complement, census-aligned \\
      & MatrAIx Persona 1M & 56{,}583 & 20 & complement, high-dimensional \\
      & PersonaHub & 199{,}466 & --- & retrieval and text material \\
    \bottomrule
  \end{tabular}
\end{table}

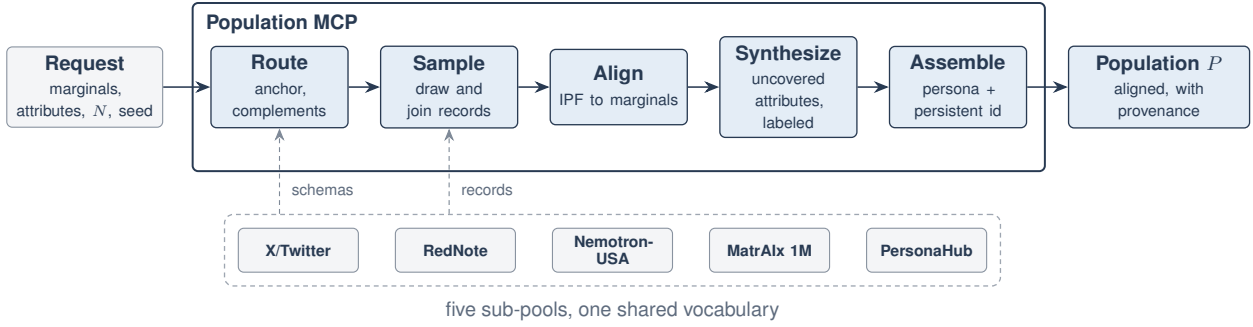
\begin{figure}[t]
  \centering
  \resizebox{\textwidth}{!}{\begin{tikzpicture}[
  stage/.style={fignode, text width=19mm, minimum height=12mm, inner sep=4pt},
  side/.style={figplain, text width=22mm, minimum height=12mm, inner sep=4pt},
  outnode/.style={fignode, text width=26mm, minimum height=12mm, inner sep=4pt},
  pool/.style={figplain, text width=17mm, minimum height=7mm, inner sep=3pt, font=\sffamily\scriptsize\bfseries},
  tag/.style={font=\sffamily\scriptsize, text=figink!75},
  note/.style={font=\sffamily\footnotesize, text=figink!75},
]
  \node[side] (req) at (0,0)
    {\textbf{Request}\\[1pt]\scriptsize marginals, attributes, $N$, seed};

  \node[stage] (route)  at (3.1,0)  {\textbf{Route}\\[1pt]\scriptsize anchor, complements};
  \node[stage] (sample) at (5.8,0)  {\textbf{Sample}\\[1pt]\scriptsize draw and join records};
  \node[stage] (align)  at (8.5,0)  {\textbf{Align}\\[1pt]\scriptsize IPF to marginals};
  \node[stage] (synth)  at (11.2,0) {\textbf{Synthesize}\\[1pt]\scriptsize uncovered attributes, labeled};
  \node[stage] (asm)    at (13.9,0) {\textbf{Assemble}\\[1pt]\scriptsize persona + persistent id};
  \begin{scope}[on background layer]
    \node[figmcp, fit=(route)(sample)(align)(synth)(asm), inner xsep=8pt, inner ysep=15pt] (mcp) {};
  \end{scope}
  \node[font=\sffamily\small\bfseries, text=figink, anchor=north west]
    at ([xshift=3pt,yshift=-2pt]mcp.north west) {Population MCP};

  \node[outnode] (out) at (17.1,0)
    {\textbf{Population $P$}\\[1pt]\scriptsize aligned, with provenance};

  \draw[figarrow] (req) -- (route);
  \draw[figarrow] (route) -- (sample);
  \draw[figarrow] (sample) -- (align);
  \draw[figarrow] (align) -- (synth);
  \draw[figarrow] (synth) -- (asm);
  \draw[figarrow] (asm) -- (out);

  \node[pool] (p1) at (3.4,-2.6)  {X/Twitter};
  \node[pool] (p2) at (5.9,-2.6)  {RedNote};
  \node[pool] (p3) at (8.4,-2.6)  {Nemotron-USA};
  \node[pool] (p4) at (10.9,-2.6) {MatrAIx 1M};
  \node[pool] (p5) at (13.4,-2.6) {PersonaHub};
  \begin{scope}[on background layer]
    \node[figframe, fit=(p1)(p2)(p3)(p4)(p5), inner sep=6pt] (reg) {};
  \end{scope}
  \node[note, anchor=north] at ([yshift=-3pt]reg.south) {five sub-pools, one shared vocabulary};
  \draw[figarrowlight, figdashed] (reg.north -| route.south) -- node[tag, right=2pt, pos=0.28] {schemas} (route.south);
  \draw[figarrowlight, figdashed] (reg.north -| sample.south) -- node[tag, right=2pt, pos=0.28] {records} (sample.south);
\end{tikzpicture}}
  \caption{One call to the \userpool. A request specifies target marginals, required attributes, size, and seed. The service routes it across the five registered pools, samples and joins records, aligns them to the marginals by IPF, synthesizes what no pool covers under a synthetic label, and assembles the study population $P$ with persistent identifiers and attribute provenance.}
  \label{fig:userpool}
\end{figure}

\paragraph{Open Pools.}
Three open pools extend coverage along different axes, and each enters the service with an explicitly documented capability boundary.
Nemotron-Personas USA~\citep{nvidia2025nemotron} contributes synthetic personas aligned to U.S.\ Census marginal distributions, each with a free-text persona description, and serves as a fully open counterpart to the internal pools for reproducible research.
MatrAIx Persona 1M~\citep{matraix2026persona} contributes the widest demographic schema, spanning 1,290 categorical dimensions upstream, from which we adopt the eighteen most analytically useful. Its authors state that the corpus is calibrated only on one-dimensional marginals and does not represent any real population, and a substantial share of its persona texts is generated from attributes, so records of this kind are labeled and never treated as real behavioral data.
PersonaHub~\citep{ge2024personahub} is the largest connected corpus, on the order of one billion synthetic personas, but is text-only: it carries no structured demographic fields, so it participates in retrieval and as textual material while being excluded, by design, from any distributional sampling.
New pools join through a declarative registration that records the source, its persona and demographic fields, and its attribution and usage terms. Adding an open dataset requires only this registration and a one-time normalization of its records into the shared vocabulary.

\paragraph{Multi-pool Routing and Demand-driven Synthesis.}
Studies with practical research problems often require persona attributes that no single pool fully covers.
For any requested set of attributes, the \userpool identifies an \emph{anchor pool} with the highest coverage, \emph{complement pools} that contribute missing dimensions through statistical matching on shared attributes, and any attributes that remain uncovered.
Uncovered attributes are filled by \emph{conditioned synthesis}: the missing values are generated conditional on each persona's existing demographic profile, preserving the attribute correlations observed in the data.
When a study requires joint-distribution fidelity, iterative proportional fitting reweights the pool's observed joint distribution to match the declared target marginals, preserving the attribute correlations present in the data.
If the target marginals call for population cells that no pool supports, the service either supplies clearly labeled synthetic respondents so that the delivered population still matches the requested margins, or, at the researcher's choice, excludes those cells and reports the gap.
In both modes, coverage limitations are reported explicitly.

\paragraph{Service Interface.}
The \userpool exposes ten typed tools organized into four functional groups.
\emph{Discovery} lists the registered pools, describes their schemas, and queries value distributions.
\emph{Routing} performs cross-pool field matching with coverage analysis.
\emph{Retrieval} covers profile search and individual lookup.
\emph{Simulation} runs end-to-end population sampling with optional IPF reweighting and parallel LLM questionnaire execution.

\paragraph{A Standard Flow.}
The standard flow from the user pool to a study's $P$ runs in six steps, as shown in \Cref{fig:userpool}.
(1)~Declare target marginal distributions, taken from census tables, survey microdata, or domain knowledge.
(2)~Route to the best-matching pool or pools via cross-pool field matching.
(3)~Sample from the anchor pool and complement missing dimensions from secondary pools via {statistical matching on shared keys}.
(4)~Apply IPF to align the joint distribution to target marginals.
(5)~Synthesize any remaining unresolved attributes conditioned on existing demographics, recording the source of every attribute for later attribution.
(6)~Assemble each persona with demographic tags, a natural-language profile, and domain-specific initial state.
The agentic research pipeline is compatible with both the \userpool and any user-uploaded population files.
A study can equally draw its population from a researcher-supplied data file or from a study-specific construction, and in every case the population is materialized once at initialization, deterministically under a declared seed, so that repeating a study reproduces exactly the same set of agents and the panel key of \S\ref{sec:framework:behavior} remains stable across versions.

\subsection{Real-World Signal Sources and the \texorpdfstring{\eventtool}{Event MCP}}
\label{sec:infra:signals}

On the environment side, the \eventtool serves 21+ structured signal sources with point-in-time guarantees.

\paragraph{Signal Source Catalog.}
The \eventtool aggregates 21 structured data sources organized into four categories (\Cref{tab:signal-sources}).
Beyond U.S.-centric sources, the catalog includes the World Bank API (GDP, CPI, unemployment, and trade indicators for seven major economies), enabling cross-national studies and macro-economic simulations that require non-U.S.\ context.
The catalog is extensible: each source is wrapped as an independent connector behind a common interface, so new data sources can be added without changes to the rest of the platform.

\begin{table}[t]
  \centering
  \caption{Signal sources available through the \eventtool, grouped by category.  Sources marked with $\star$ are enabled by default.}
  \label{tab:signal-sources}
  \small
  \begin{tabular}{llp{5.5cm}}
    \toprule
    Category & Source & Coverage \\
    \midrule
    \multirow{5}{*}{Macro / Market}
      & FRED$^\star$ & Federal Reserve economic indicators \\
      & BLS$^\star$ & Bureau of Labor Statistics \\
      & Treasury$^\star$ & U.S.\ Treasury rates \\
      & NY Fed SCE$^\star$ & Survey of Consumer Expectations \\
      & World Bank & GDP, CPI, unemployment for 7 economies \\
    \addlinespace
    \multirow{4}{*}{Financial}
      & Yahoo Finance$^\star$ & Equity / index prices \\
      & Credit Spreads$^\star$ & Corporate bond spreads \\
      & Housing Finance$^\star$ & Mortgage / housing indices \\
      & Alt.\ Fear \& Greed$^\star$ & Market sentiment composite \\
    \addlinespace
    \multirow{4}{*}{News / Trends}
      & NYT Archive & New York Times articles \\
      & Google Trends & Search interest time series \\
      & GDELT & Global event database \\
      & USTR$^\star$ & U.S.\ trade policy actions \\
    \addlinespace
    \multirow{4}{*}{Population / Survey}
      & Census ACS & American Community Survey \\
      & Household Pulse & Census Household Pulse Survey \\
      & CPS Basic / ASEC & Current Population Survey \\
      & SIPP & Survey of Income and Program Participation \\
    \bottomrule
  \end{tabular}
\end{table}

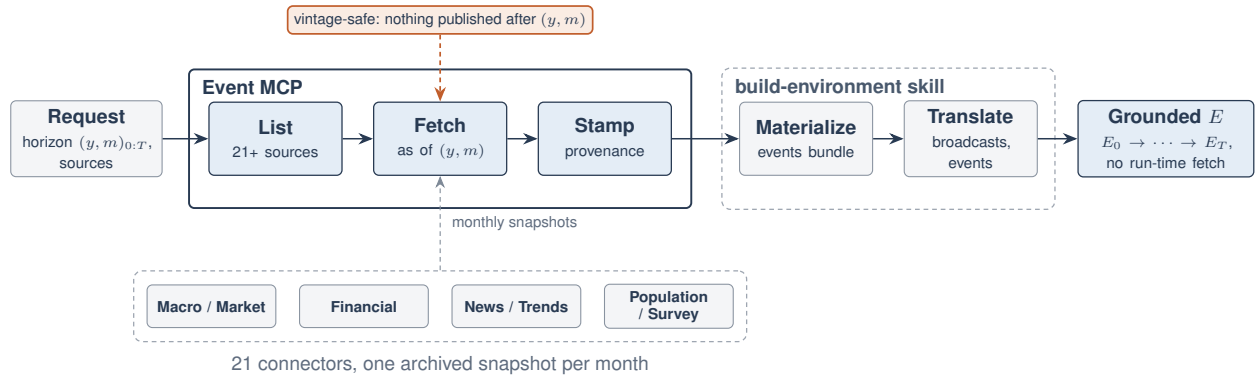
\begin{figure}[t]
  \centering
  \resizebox{\textwidth}{!}{\begin{tikzpicture}[
  stage/.style={fignode, text width=19mm, minimum height=12mm, inner sep=4pt},
  bstage/.style={figplain, text width=19mm, minimum height=12mm, inner sep=4pt},
  side/.style={figplain, text width=22mm, minimum height=12mm, inner sep=4pt},
  outnode/.style={fignode, text width=26mm, minimum height=12mm, inner sep=4pt},
  cat/.style={figplain, text width=19mm, minimum height=7mm, inner sep=3pt, font=\sffamily\scriptsize\bfseries},
  tag/.style={font=\sffamily\scriptsize, text=figink!75},
  note/.style={font=\sffamily\footnotesize, text=figink!75},
]
  \node[side] (req) at (0,0)
    {\textbf{Request}\\[1pt]\scriptsize horizon $(y,m)_{0:T}$, sources};

  \node[stage] (list)  at (3.1,0) {\textbf{List}\\[1pt]\scriptsize 21+ sources};
  \node[stage] (fetch) at (5.8,0) {\textbf{Fetch}\\[1pt]\scriptsize as of $(y,m)$};
  \node[stage] (stamp) at (8.5,0) {\textbf{Stamp}\\[1pt]\scriptsize provenance};
  \begin{scope}[on background layer]
    \node[figmcp, fit=(list)(fetch)(stamp), inner xsep=9pt, inner ysep=15pt] (mcp) {};
  \end{scope}
  \node[font=\sffamily\small\bfseries, text=figink, anchor=north west]
    at ([xshift=3pt,yshift=-2pt]mcp.north west) {Event MCP};

  \node[bstage] (mat)   at (11.8,0) {\textbf{Materialize}\\[1pt]\scriptsize events bundle};
  \node[bstage] (trans) at (14.5,0) {\textbf{Translate}\\[1pt]\scriptsize broadcasts, events};
  \begin{scope}[on background layer]
    \node[figframe, fit=(mat)(trans), inner xsep=8pt, inner ysep=15pt] (build) {};
  \end{scope}
  \node[font=\sffamily\small\bfseries, text=figink!85, anchor=north west]
    at ([xshift=3pt,yshift=-2pt]build.north west) {build-environment skill};

  \node[outnode] (out) at (17.7,0)
    {\textbf{Grounded $E$}\\[1pt]\scriptsize $E_0 \to \cdots \to E_T$, no run-time fetch};

  \draw[figarrow] (req) -- (list);
  \draw[figarrow] (list) -- (fetch);
  \draw[figarrow] (fetch) -- (stamp);
  \draw[figarrow] (stamp) -- (mat);
  \draw[figarrow] (mat) -- (trans);
  \draw[figarrow] (trans) -- (out);

  \node[figaccentnode, font=\sffamily\scriptsize, inner sep=3pt] (asof) at (5.8,1.95)
    {vintage-safe: nothing published after $(y,m)$};
  \draw[figarrowaccent, figdashed] (asof) -- (fetch);

  \node[cat] (c1) at (2.05,-2.75) {Macro / Market};
  \node[cat] (c2) at (4.55,-2.75) {Financial};
  \node[cat] (c3) at (7.05,-2.75) {News / Trends};
  \node[cat] (c4) at (9.55,-2.75) {Population / Survey};
  \begin{scope}[on background layer]
    \node[figframe, fit=(c1)(c2)(c3)(c4), inner sep=6pt] (catalog) {};
  \end{scope}
  \node[note, anchor=north] at ([yshift=-3pt]catalog.south) {21 connectors, one archived snapshot per month};
  \draw[figarrowlight, figdashed] (catalog.north -| fetch.south) -- node[tag, right=2pt] {monthly snapshots} (fetch.south);
\end{tikzpicture}}
  \caption{One call to the \eventtool. A request specifies the study horizon and sources. The service lists the catalog, fetches each source as of the requested year and month so that nothing published later is visible, and stamps provenance. The build-environment skill then materializes the result as an external-events bundle and translates it into broadcasts and scheduled events bound to steps, so the run itself fetches nothing.}
  \label{fig:eventtool}
\end{figure}

\paragraph{Point-in-time Access.}
For any forecasting or nowcasting study (e.g., \S\ref{sec:cases:consumersim}, \S\ref{sec:cases:pmi}, and \S\ref{sec:cases:marketsim}), credibility depends on the guarantee that the simulation at step~$t$ sees only data that was publicly available at the corresponding real-world date.
The \eventtool enforces this through \emph{vintage-safe} access: queries are parameterized by (year, month), and each collector returns only the data vintage that would have been available at that point.
This prevents look-ahead bias and makes all temporal evaluations methodologically sound, and it applies equally to an intervention added later in a study: a broadcast declared for a given month is grounded by the same as-of query, so the intervened branch is held to the same information discipline as the parent run.

\paragraph{Service Interface.}
The \eventtool answers three kinds of requests: listing the available sources and what they cover, retrieving data from multiple sources for a given year and month, and retrieving a single source in detail with optional sub-series selection.
Every response carries structured data together with its provenance, so that any environmental signal used in a study can be traced back to its origin.

\paragraph{A Standard Flow.}
The standard flow from the signal sources to a study's $E$ runs in six steps, as shown in \Cref{fig:eventtool}.
(1)~Declare the study horizon as a range of year-month pairs and select the sources the study needs.
(2)~List the catalog to confirm the coverage and fields of each selected source.
(3)~Fetch each source as of every month of the horizon, so that the payload of a month holds only what had been published by then.
(4)~Stamp every payload with its provenance, that is, the source, the series, and the vintage it came from.
(5)~Materialize the stamped payloads in the study workspace as an external-events bundle, so that the run reads from the workspace and never from the services, which is what makes a run self-contained and a replayed branch comparable with its parent.
(6)~Translate the bundle into the two channels of the environment, broadcasts for the information layer and scheduled events for the physical layer, each bound to the step of its month (\S\ref{sec:framework:behavior}).
Steps 2 to 4 take place inside the service and steps 5 and 6 inside the build-environment skill, so resolution happens at build time and not at run time.

\subsection{Governance, Auditability, and Reproducibility}
\label{sec:infra:governance}

\paragraph{Privacy and De-identification.}
The \userpool exposes only demographic tags and statistical profiles, never raw social-media content or personally identifiable information, and all access passes through the typed service interface.
All pools comply with the terms of service of their source platforms.
One open pool (MatrAIx) includes personas derived from public information about public figures. These records are used only in research settings, and ingestion removes fictional characters and minors.

\paragraph{Licensing and Attribution.}
Every registered pool and signal source carries a documented attribution and a usage tier.
Sources cleared for unrestricted use are available to production runs and published studies, whereas sources with restricted or undeclared terms are limited to research and demonstration use and require explicit per-study confirmation before any commercial run.
Generated study reports automatically include a data-basis section crediting every pool and signal source used, so licensing obligations propagate to the studies built on the platform.

\paragraph{Research Auditability.}
Every stage of the pipeline writes its artifact to disk, every intervention is logged in the events table, every run is versioned, and every study declares the data services, datasets, and tools it depends on.
Any conclusion in a report can therefore be traced through the trajectory store to the study definition, environment bundle, population bundle, and intervention schedule that produced it, and through the resource manifest to the pools and signal vintages behind them.

\paragraph{Versioning and Reproducibility.}
Pool snapshots are versioned. Signal sources are archived by vintage (year-month).
A study's resource manifest pins the pool and source versions it depends on, and the trajectory store records which external-events bundle was used for each run.
Combined with the deterministic-seed population construction and the archived environment artifacts, this ensures that any published result can be reproduced from the same data inputs.

\section{Evaluation on Social Science Simulation}
\label{sec:evaluation}

This section evaluates \sysname as an instrument for social science through seven case studies organized into three families, spanning from methodological benchmarking to real-world forecasting. 
\S\ref{sec:evaluation:methods} introduces the shared workflow and evaluation dimensions. \S\ref{sec:cases:mechanisms}, \S\ref{sec:cases:policy}, and \S\ref{sec:cases:macro} present each family in turn, progressing from reproducing known dynamics to predicting outcomes not yet observed at simulation time.
Each case is realized in the framework of \Cref{sec:framework} and operated through the infrastructure of \Cref{sec:infra}. Consequently, interventions and branches inside the longitudinal simulation loop, and reference conditions and ablations as versions raised in the controllable research loop, are both realized in seven cases.

\subsection{Evaluation Methods}
\label{sec:evaluation:methods}

\begin{figure}[t]
  \centering
  \resizebox{\textwidth}{!}{\begin{tikzpicture}[
  block/.style={draw=figink, line width=0.6pt, rounded corners=3pt, fill=figblue, inner sep=5pt, align=center, minimum width=38mm, minimum height=26mm},
  blockw/.style={block, fill=white},
  btitle/.style={font=\sffamily\small\bfseries, text=figink, inner sep=2pt},
  bline/.style={font=\sffamily\scriptsize, text=figink!85, align=center},
  chipP/.style={font=\sffamily\footnotesize, text=figink, draw=figink, line width=0.5pt, rounded corners=2pt, fill=white, inner xsep=5pt, inner ysep=2.5pt},
  evalbox/.style={font=\sffamily\footnotesize, text=figink, draw=figink, line width=0.6pt, rounded corners=3pt, fill=figgray, inner sep=4pt, align=left, text width=33mm, anchor=west},
  skill/.style={font=\sffamily\footnotesize\scshape, text=figink!85, draw=figink!55, line width=0.4pt, rounded corners=2pt, fill=white, inner xsep=6pt, inner ysep=3pt},
  glyph/.style={draw=figink, line width=0.5pt},
]
  \node[block] (rq) at (0,1.3) {};
  \node[btitle] at (0,1.65) {Research question};
  \node[bline] at (0,0.85) {target aggregate and trajectory\\[-1pt] directional hypothesis};

  \node[block] (fm) at (4.4,1.3) {};
  \node[btitle] at (4.4,1.75) {Framework Mapping};
  \node[chipP] at (3.6,1.0) {$P$}; \node[chipP] at (4.4,1.0) {$E$}; \node[chipP] at (5.2,1.0) {$f$};

  \node[block] (ed) at (8.8,1.3) {};
  \node[btitle] at (8.8,1.65) {Experimental design};
  \node[bline] at (8.8,0.85) {ground truth $\cdot$ references\\[-1pt] protocol $\cdot$ mechanism probe};

  \draw[figarrow] (rq) -- (fm);
  \draw[figarrow] (fm) -- (ed);

  \node[blockw] (run) at (13.2,1.3) {};
  \node[btitle] at (13.2,2.15) {Silicon samples};
  \begin{scope}[shift={(13.2,1.15)}]
    \foreach \i in {-1,0,1} {
      \fill[figblue, draw=figink, line width=0.4pt] (\i*0.45,0.42) circle (0.9mm);
      \fill[figblue, draw=figink, line width=0.4pt, rounded corners=1pt] (\i*0.45-0.12,0.1) rectangle (\i*0.45+0.12,0.32);
    }
    \draw[glyph] (-1.3,-0.55) -- (1.3,-0.55);
    \draw[glyph] (-1.3,-0.55) -- (-1.3,-0.05);
    \draw[draw=figink!85, line width=0.7pt] plot[smooth, tension=0.8] coordinates {(-1.25,-0.42) (-0.7,-0.2) (-0.2,-0.38) (0.4,-0.2) (1.25,-0.1)};
    \draw[draw=figaccent, line width=0.7pt, figdashed] plot[smooth, tension=0.8] coordinates {(0.1,-0.3) (0.6,-0.48) (1.25,-0.52)};
    \draw[draw=figaccent, line width=0.5pt, figdashed] (0.1,-0.55) -- (0.1,-0.05);
  \end{scope}
  \node[bline] at (13.2,0.35) {$E_t \rightarrow B_t \rightarrow E_{t+1}$};
  \draw[figarrow] (ed) -- (run);

  \node[evalbox] (q1) at (16.3,2.9)  {\textbf{Micro alignment}\\[1pt]{\scriptsize\color{figink!80}responses vs.\ real records}};
  \node[evalbox] (q2) at (16.3,1.83) {\textbf{Level fidelity}\\[1pt]{\scriptsize\color{figink!80}aggregates vs.\ ground truth}};
  \node[evalbox] (q3) at (16.3,0.77) {\textbf{Trajectory fidelity}\\[1pt]{\scriptsize\color{figink!80}course vs.\ observed course}};
  \node[evalbox] (q4) at (16.3,-0.3)  {\textbf{Mechanism}\\[1pt]{\scriptsize\color{figink!80}branches vs.\ the parent run}};
  \coordinate (fork) at (15.6,1.3);
  \draw[figarrow] (run.east) -- (fork);
  \foreach \q in {q1,q2,q3,q4} \draw[figarrow, rounded corners=4pt] (fork) |- (\q.west);
  \draw[draw=figaccent, line width=0.7pt, figdashed] (15.75,-0.3) -- (q4.west |- {(0,-0.3)});

  \coordinate (bandL) at (rq.west |- {(0,-1.55)});
  \coordinate (bandR) at (q1.east |- {(0,-1.55)});
  \node[draw=figink!45, line width=0.5pt, rounded corners=4pt, fill=figgray!60, fit=(bandL) (bandR), minimum height=11mm, inner sep=0pt] (band) {};
  \node[skill] (k1) at ($(bandL)!0.07!(bandR)$) {init};
  \node[skill] (k2) at ($(bandL)!0.20!(bandR)$) {build-model};
  \node[skill] (k3) at ($(bandL)!0.36!(bandR)$) {build-environment};
  \node[skill] (k4) at ($(bandL)!0.53!(bandR)$) {build-population};
  \node[skill] (k5) at ($(bandL)!0.66!(bandR)$) {run};
  \node[skill] (k6) at ($(bandL)!0.78!(bandR)$) {report};
  \node[skill] (k7) at ($(bandL)!0.91!(bandR)$) {iterate};
  \foreach \a/\b in {k1/k2,k2/k3,k3/k4,k4/k5,k5/k6,k6/k7} \draw[figarrowlight] (\a) -- (\b);

  \coordinate (rlL) at (rq.west |- {(0,3.05)});
  \coordinate (rlR) at (ed.east |- {(0,3.05)});
  \draw[draw=figaccent, line width=0.8pt] ([yshift=-4pt]rlL) -- (rlL) -- (rlR) -- ([yshift=-4pt]rlR);
  \node[font=\sffamily\footnotesize\bfseries, text=figaccent, anchor=south, align=center, inner sep=2pt] (rll) at ($(rlL)!0.5!(rlR)$)
    {Controllable research loop\\\mdseries edits on $P$, $E$, $f$, $\Theta$ raise versions};
  \coordinate (slL) at (run.west |- {(0,3.55)});
  \coordinate (slR) at (q1.east |- {(0,3.55)});
  \draw[draw=figink, line width=0.8pt] ([yshift=-4pt]slL) -- (slL) -- (slR) -- ([yshift=-4pt]slR);
  \node[font=\sffamily\footnotesize\bfseries, text=figink, anchor=south, align=center, inner sep=2pt] (sll) at ($(slL)!0.5!(slR)$)
    {Longitudinal simulation loop\\\mdseries one run, interventions fork branches};

  \begin{scope}[on background layer]
    \node[figframe, fit={(rq) (q1) (q4) (band) (rll) (sll)}, inner sep=10pt] (frame) {};
  \end{scope}
  \node[figlabel, anchor=north west, font=\sffamily\footnotesize\bfseries, fill=white, inner sep=3pt] at ([xshift=6pt, yshift=6pt]frame.north west) {agentic research pipeline};
\end{tikzpicture}}
  \caption{Realization of a case study in the framework and the flow of its results into evaluation. Each case is realized through three uniform steps: research question, framework mapping, and experimental design. Its silicon samples then act over the horizon, and the results are read from four angles. The agentic pipeline carries every step. The controllable research loop spans the three realization steps, where an edit to $P$, $E$, $f$, or $\Theta$ raises a new version, and the longitudinal simulation loop spans the run, where interventions fork branches.}
  \label{fig:evaluation}
\end{figure}
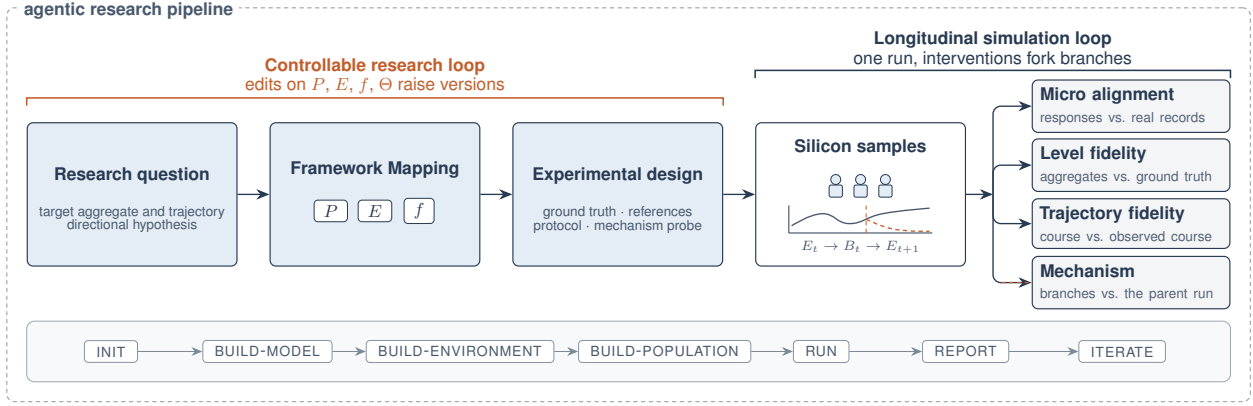

\paragraph{Case realization in the framework.}
Each case study follows the same three phases (\Cref{fig:evaluation}).

The \emph{research question} states what the simulated aggregate must match and what the trajectory is expected to add, with a directional hypothesis.
The \emph{framework mapping} states the population, the environment, and the behavior function separately, so that a case is an instantiation of \Cref{eq:bfpe} whose three terms can be inspected and replaced independently.
The \emph{experimental design} specifies, in fixed order, the ground truth and evaluation window, the reference conditions against which the simulation is scored, the longitudinal protocol, and, where applicable, a mechanism probe by ablation or intervention. The longitudinal protocol covers the time unit, step length in real time, horizon, the state carried across steps, and the number of repetitions.
These steps correspond to the artifacts that the agentic pipeline produces, so a case is executable from its case description: the population and environment bundles realize the mapping, and the intervention schedule and run configuration realize the design.
The simulated individuals then act over the horizon as \emph{silicon samples}, a fixed panel that observes the environment, decides through the behavior function, and changes the environment step by step, with interventions entering where the design declares them.
In terms of \Cref{eq:edit}, a reference condition is a version that replaces $f$ or $P$, an ablation is a version that removes one component, and an intervention branch is one that shares its parent's history, so every comparison in this section is a comparison between versions or between branches of one study, and the reported result of each case is the version its researchers accepted.

\paragraph{Result interpretation from four angles.}
The results of every case are read from four angles in a fixed order, each against the ground truth the design names.
\emph{Micro alignment} asks whether individual silicon samples respond as their real counterparts did, where individual records exist.
\emph{Level fidelity} asks whether endpoints or per-period aggregates agree with the ground truth.
\emph{Trajectory fidelity} asks whether the course over time agrees with the observed course, measured by quantities that change when only the order of steps changes.
\emph{Mechanism} asks what ablations, intervention branches, and per-agent records reveal about why the aggregate moves, by comparing branches of the same simulation. This dimension also identifies where the simulation's agreement with ground truth breaks down.
The ground truth differs by family: rule-based dynamics and observed opinion trajectories in the mechanism family, census patterns and historical procurement records in the policy family, and official statistics and registration data in the forecasting family.

\Cref{tab:case-protocol} places the seven designs side by side on the elements above. The reporting protocol for repetitions and uncertainty is consolidated in \Cref{app:cases:protocol}, and a two-sentence takeaway closes each case.

\begin{table}[t]
  \centering
  \caption{Experimental designs of the seven case studies on one template.}
  \label{tab:case-protocol}
  \footnotesize
  \renewcommand{\arraystretch}{1.2}
  \setlength{\tabcolsep}{5pt}
  \makebox[\textwidth][c]{\resizebox{1.08\textwidth}{!}{%
  \begin{tabular}{llllllll}
    \toprule
    Case & Units & Step & Steps & Ground truth & Reference & Probe & Runs \\
    \midrule
    \rowcolor{figgray}\multicolumn{8}{c}{\textit{Case Family I: Simulation Mechanisms}} \\
    Canonical ABM reproduction & 40--810 agents, 10 models & abstract & 30--200 & rule dynamics & rule control, 3 LLMs & context enrichment & 10 / 3 \\
    Hybrid opinion dynamics & 1,000 users & 12 h & 14 & attitude series & pure rule $\times$ 5 & 3 interventions & 3 \\
    \rowcolor{figgray}\multicolumn{8}{c}{\textit{Case Family II: Policy Simulation}} \\
    Chicago segregation & 19,235 households & abstract & 15 & 2010 Census & 5 rule models & 3 ablations & 5 / 10 \\
    Drug procurement & 325 markets & round & 50 & historical winners & enterprise-only, random & & 1 \\
    \rowcolor{figgray}\multicolumn{8}{c}{\textit{Case Family III: Macro-Index Forecasting}} \\
    Consumer-confidence nowcast & 250,000 households & month & 75 & official CCI & 12 baselines & 4 ablations, scale & 1 \\
    Purchasing-managers nowcast & 300 firms & month & 22 + 32 & official PMI & persistence, consensus & 7 ablations & 1 \\
    Car-market shares & 600 consumers & month & 41 & registrations & & & 1 \\
    \bottomrule
  \end{tabular}}}
\end{table}

\subsection{Case Family I: Simulation Mechanisms}
\label{sec:cases:mechanisms}

The two cases in this family examine the mechanics of LLM social simulation itself, establishing that the instrument reproduces known dynamics before it is applied to policy and forecasting questions, and they emphasize the reference conditions of the template: rule-based dynamics against LLM-driven decisions in the benchmark case (\S\ref{sec:cases:abm}), and pure rule-based populations against hybrid populations that couple LLM-driven core users with rule-based ordinary users in a Twitter-like environment (\S\ref{sec:cases:hisim}).
In terms of the two loops, the benchmark case varies $f$ across versions, a rule against three LLMs on the same $(P,E)$, and the opinion case declares interventions inside a run, so this case family exercises the controllable research loop and the longitudinal simulation loop in turn.

\subsubsection{Case Study 1: Reproducing Canonical ABMs with LLM Agents}
\label{sec:cases:abm}

\paragraph{Research Question.}
Classical agent-based models are the canonical demonstrations that collective patterns emerge from simple local rules: Schelling's residential model produces segregation from mild individual preferences~\citep{schelling1971dynamic}, and the Sugarscape economy grows wealth inequality from foraging agents on a resource landscape~\citep{epstein1996growing}.
Their dynamics are known, reproducible, and rule-based, which makes them the natural reference for asking whether LLM-driven decision functions preserve the mechanisms that social science has already validated.
The case asks two questions: whether, once the hand-coded rules are replaced by LLM-driven decisions over the same observations and actions, the same group-level outcomes arise, and whether the LLM-driven trajectory follows the same course toward those outcomes, and how the course changes when the decision context is enriched beyond the scalar statistics of the classical rule.
The hypothesis is directional: LLM-driven runs are expected to reach consistency scores close to the run-to-run agreement of the rule-based model itself, with the widest gaps on tasks whose reference is deterministic or whose actions are continuous, and enriched context is expected to change the tempo of convergence without changing its endpoint.

\paragraph{Framework Mapping.}
The benchmark organizes 10 canonical ABMs into four families, flow, market, organization, and diffusion, each expressed through the same three components. The full roster with source models, population sizes, and configuration parameters is catalogued in \Cref{tab:abm-suite} (\Cref{app:cases}).
\begin{itemize}
  \item \textbf{Population \Pop.} The agent population of each source model: residents on a grid, vehicles on a road, boids in a flock, traders in a market, or opinion holders in a network, with sizes from 40 (Boids) to 810 (Schelling).
  \item \textbf{Environment \Env.} The simulated setting of each model, a toroidal grid, a continuous field, a contact network, or a market with clearing rules, together with the task-specific observation each individual receives at every step.
  \item \textbf{Behavior function $f$.} Implemented twice per task, once as the classical rule and once as an LLM prompt chain, over identical observation and action types. Both implementations emit the same typed action (move or stay, cooperate or defect, spread or ignore, a heading vector), so both act in the same action space and are scored against the same reference.
\end{itemize}

\paragraph{Experimental Design.}
\begin{itemize}
  \item \textbf{Ground truth and window.} Rule dynamics: for every task, the ten-run mean of the rule-based model on each metric of the task's outcome vector (\Cref{tab:abm-metric-vectors}) is the reference value, and agreement is summarized by a \emph{consistency score} in $[0,1]$, one minus the root-mean-square relative deviation across the vector, computed per run and averaged per condition.
  \item \textbf{Reference conditions.} A \emph{control group} in which independent stochastic runs of the rule-based model are scored against the same reference, giving the agreement attainable under intrinsic randomness alone, and three LLM conditions on equal footing, GPT-4o, DeepSeek-V3, and Qwen3-235B, all queried at temperature 0.7 with structured output of at most 256 tokens.
  \item \textbf{Longitudinal protocol.} Ten tasks with 40 to 810 agents and 30 to 200 abstract steps each. The control group runs 10 independent seeds per task and each LLM condition 3 seeds per task.
  \item \textbf{Mechanism probe.} A context-enrichment comparison on Schelling ($20\times20$ grid, 400 agents, threshold $3/8$, 50 steps) with three configurations: the rule, LLM-$f$ over the rule's scalar neighborhood statistics, and LLM-$f$ over an enriched context in which the neighborhood is described in natural language and the persona carries decision factors such as moving cost, patience, and confidence in the status quo~\citep{zhang2025abm}.
\end{itemize}
{In terms of the controllable research loop, the control group and the three LLM conditions are versions of one study that vary $f$ on a fixed $(P,E)$, the behavior-function ablation among the applicable controls, and the context-enrichment probe is a further version that edits the observation each agent receives while the rule is held fixed.}

\begin{figure}[t]
  \centering
  \includegraphics[width=\linewidth]{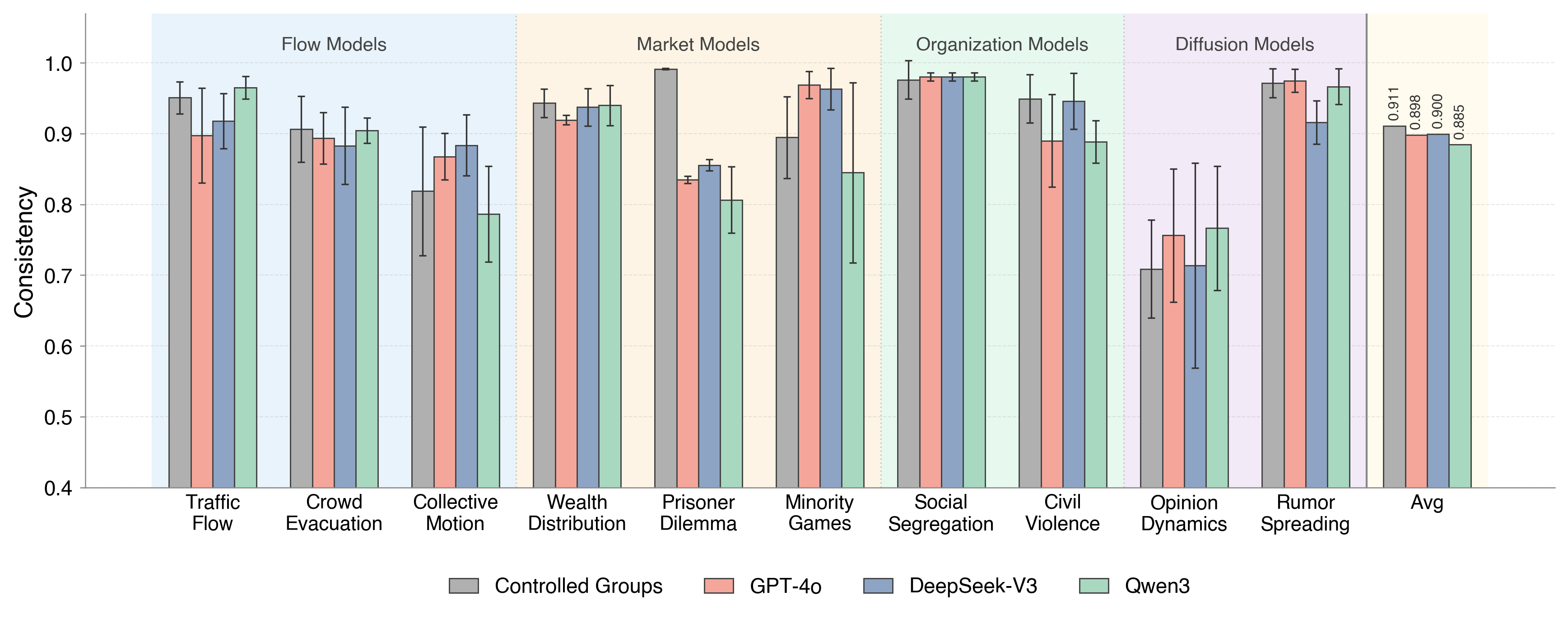}
  \caption{Consistency scores across the ABM benchmark suite (mean and standard deviation. Control group 10 runs, each LLM 3 runs). The control-group score of 0.911 reflects intrinsic stochastic variability. Adapted from~\citet{zhang2025abm}.}
  \label{fig:abm-consistency}
\end{figure}

\paragraph{Results.}
\emph{Level fidelity: LLM-driven decisions approach the ceiling set by stochastic variability alone.}
\Cref{fig:abm-consistency} reports the consistency scores across the suite.
The control group achieves a mean consistency of 0.911, the empirical upper bound for any behavior function evaluated under this protocol.
Against this ceiling, GPT-4o reaches 0.898, DeepSeek-V3 0.900, and Qwen3-235B 0.885 averaged over the ten tasks, so the gap between the best LLM condition and the rule-based model's own run-to-run agreement is about one hundredth.
The three LLM conditions lie within 0.015 of one another on the suite average, and their per-task profiles rise and fall together across the four families. GPT-4o has the smallest cross-task dispersion, and Qwen3-235B the largest, with its lowest scores on Opinion Dynamics and Collective Motion.
Task structure therefore accounts for more of the variation than model choice.

\emph{Trajectory fidelity: the agreement is mechanistic as well as numerical.}
Social Segregation is the most stable task in the suite, with the control group and all three LLM conditions within a few hundredths of one another near the top of the scale, and Rumor Spreading keeps two of the three LLM conditions close to its control group.
The qualitative comparison in \Cref{fig:abm-dynamics-compare} (\Cref{app:cases}) shows the same course under both implementations: Hegselmann--Krause opinions at confidence bound $\varepsilon=0.15$ converge to three clusters, and the SIR rumor on a Watts--Strogatz contact network reproduces the spreader peak followed by saturation of the informed population.
The mechanism probe locates where the LLM adds to the classical dynamics.
With the rule's scalar context, the LLM-driven Schelling population converges to full satisfaction by step 12, on the same course as the rule. With the enriched context it reaches the same fully segregated endpoint at step 18, because agents that weigh moving costs and patience relocate less readily at the first sign of dissatisfaction~\citep{zhang2025abm}.
Enriching the context thus changes the tempo of the trajectory and leaves its terminal pattern intact, which is the kind of difference that only a step-by-step record can register.

\emph{Boundary: the largest gaps arise where the reference is deterministic, the metric is unstable, or the action is continuous.}
Prisoner's Dilemma shows the widest separation between conditions: its round-robin tournament of fixed strategies makes the control group nearly self-identical, whereas all three LLM conditions fall well below it, since deterministic strategies leave no stochastic slack and any deviation in a single per-strategy score is counted in full.
Opinion Dynamics records the lowest scores of the suite for every condition, including the control group, with the widest error bars, because its metric vector counts final opinion clusters at three confidence bounds, a quantity that shifts discretely under small perturbations. All three LLM conditions match or exceed the control group on this task, and GPT-4o and DeepSeek-V3 also score above the control group on Minority Games, where LLM runs sit closer to the ten-run rule mean than individual stochastic rule runs do.
Collective Motion, whose actions are heading vectors, yields the lowest control-group score among flow models and the widest spread across LLM conditions, whereas Crowd Evacuation, the other continuous-action task, stays close to its control group. The difficulty is specific to flocking, whose metric vector (polarization, nearest-neighbor distance, spread) registers small heading deviations directly.
Together with Prisoner's Dilemma, these tasks delineate the regime in which exact numerical or geometric computation is required, where the typed-action contract that makes rule and LLM outputs comparable also exposes every deviation in full.

\paragraph{Takeaway.}
LLM-$f$ serves as a drop-in replacement for rule-$f$ across the ten models, reproducing the emergent outcomes to within the rule-based model's own run-to-run agreement and following the same course toward them, with enriched context altering the tempo of the dynamics and leaving their endpoint intact.
The typed-action contract that makes this comparison possible also marks its boundary, since tasks demanding exact numerical or continuous control expose every deviation, which is where the hybrid-$f$ pattern of \S\ref{sec:cases:hisim} applies.

\subsubsection{Case Study 2: Hybrid Opinion Dynamics on Social Media}
\label{sec:cases:hisim}

\paragraph{Research Question.}
Social media has become a central arena of social movements, and simulating how the public responds after a triggering event, from the leak of a court opinion to a viral hashtag, has become a matter of both scientific and practical concern~\citep{mou2024unveiling}.
Engagement on these platforms is highly skewed: a small set of active and influential opinion leaders shapes the discourse, while the bulk of participants respond with far simpler behavior.
Driving every user with an LLM is unaffordable at the population scales such events involve, and classical opinion-dynamics models cannot express the reasoning and discourse strategies of the influential minority.
This case asks whether a hybrid population, in which a few hundred LLM-driven core users are coupled with a large rule-based population of ordinary users, reproduces the observed collective opinion trajectory after a triggering event, and whether it does so at a cost that stays flat as the ordinary population grows.
The hypothesis is directional: for every rule governing the ordinary users, coupling in the LLM-driven core is expected to raise the correlation with the observed trajectory, with the largest correction where the rule alone misses the direction of change, and the run time is expected to depend on the number of core users only.

\paragraph{Framework Mapping.}
The case instantiates the framework's hybrid-$f$ pattern (\S\ref{sec:framework:behavior}) on a two-tier population situated in a Twitter-like environment, with the per-step group attitude distribution as the longitudinal trajectory.
\begin{itemize}
 \item \textbf{Population \Pop.} Each event is simulated with 1{,}000 users drawn from the real participants of that event: 300 core users, selected by influence and activity, are LLM-driven simulated individuals constructed from their real profiles, historical tweets (personal memory), and follower lists. The other 700 are rule-based simulated individuals whose initial attitudes are annotated from their real tweets.
 \item \textbf{Environment \Env.} A Twitter-like information environment provides each user with a timeline of the most recent visible tweets, follower-graph visibility, and a notification inbox, together with turn-indexed news injections that introduce real-world developments at designated steps. Each event window is simulated as 14 steps of 12 hours, so that a seven-day period is covered (in the \emph{Roe} scenario, the overturn is injected at step~0 and a report of protest arrests at step~11).
 \item \textbf{Behavior function $f$.} Core users run LLM-$f$: persona, timeline, and memory are composed into a prompt that yields a typed action (post, retweet, reply, like, or no action) with its content. Ordinary users run rule-$f$ under one of five classical opinion-dynamics models (bounded confidence~\citep{deffuant2000mixing}, Hegselmann--Krause, Lorenz, relative agreement, or social judgement), which update a scalar attitude. The two tiers are coupled through mirror agents: every core user has a counterpart inside the rule-based model whose attitude is overwritten at each step with the stance and intensity scored from that core user's actual message, so that the reasoning of opinion leaders propagates into the ordinary population. Influence in the reverse direction is not modeled, following the original design.
\end{itemize}

\paragraph{Experimental Design.}
\begin{itemize}
 \item \textbf{Ground truth and window.} The observed collective attitude trajectory of each movement, annotated from the real tweets of the 1{,}000 simulated users at every step. Three U.S.\ movements, MeToo (sexual harassment), \emph{Roe v.\ Wade} (abortion rights), and Black Lives Matter (racial justice), each observed in two event windows of about one week~\citep{mou2024unveiling}. The first window of each movement serves calibration and the second window validation. All results below are on the validation window.
 \item \textbf{Reference conditions.} Every hybrid configuration is paired with the pure rule-based model that shares its parameters, for each of the five opinion-dynamics rules, giving fifteen hybrid-versus-pure pairings across the three movements. Parameters are swept in pure rule-based form on the calibration window and transferred to the hybrid configuration, so the comparison holds the rule and its parameters fixed and varies only the presence of the LLM-driven core.
 \item \textbf{Longitudinal protocol.} 1{,}000 users per movement (300 core, 700 ordinary), 14 steps of 12 hours. Core users driven by LLM agents. Macro results averaged over three runs per scenario. Two levels of evaluation: \emph{micro} alignment tests whether individual core users reproduce the stance, content, and action type of their real counterparts in single-turn responses to the context they actually saw. \emph{macro} dynamics compares the simulated and observed group attitude trajectories over the 14 steps by Pearson correlation (Corr) and the deviation of mean attitude ($\Delta$Bias).
 \item \textbf{Mechanism probe.} The rule governing the ordinary users is treated as an experimental variable across the five models, and three platform interventions are applied on the MeToo scenario: feeding users opposing opinions, feeding neutral opinions, and opening public hashtag spaces for discussion~\citep{mou2024unveiling}.
\end{itemize}
The study is re-implemented natively on the \sysname engine as a from-scratch study, with the two tiers as implementations of one decision model and the mirror coupling as an environment update. In that re-implementation, the five ordinary-user rules are versions that vary $f$, and the three platform interventions are declared broadcasts, so each intervened run is a branch of the uninterrupted one.

\begin{figure}[!t]
 \centering
 \resizebox{\textwidth}{!}{\begin{tikzpicture}[
  every node/.style={font=\sffamily},
]
\pgfplotsset{
  hisimaxis/.style={
    width=6.1cm, height=4.4cm,
    xmin=-0.7, xmax=13.7, ymin=-0.05, ymax=1.05,
    xtick={0,4,8,12}, ytick={0,0.5,1},
    tick label style={font=\scriptsize},
    xlabel style={font=\footnotesize},
    axis line style={figink!60},
    tick style={figink!60},
    clip=false,
  },
}
\tikzset{
  evlabel/.style={font=\sffamily\tiny\itshape, text=figink, align=center, anchor=south},
  gtline/.style={draw=figink, densely dashed, line width=0.8pt},
  simline/.style={draw=figaccent, line width=0.9pt},
}
\begin{groupplot}[
  group style={group size=3 by 1, horizontal sep=0.55cm, ylabels at=edge left},
  hisimaxis,
]
  \nextgroupplot[xlabel={(a) MeToo}, ylabel={\footnotesize normalized sentiment},
                 legend style={font=\tiny, draw=none, fill=white, fill opacity=0.75,
                               text opacity=1, at={(0.97,0.97)}, anchor=north east,
                               inner sep=1.5pt, row sep=-1pt}]
    \fill[figgray] (axis cs:-0.5,-0.05) rectangle (axis cs:2.4,1.05);
    \node[evlabel] at (axis cs:1.0,1.09) {Moore case,\\[-1pt]Golden Globes};
    \addplot[simline, mark=*, mark size=0.9pt, mark options={solid, fill=figaccent}]
      coordinates {(0,0.387) (1,0.348) (2,1.000) (3,0.619) (4,0.619) (5,0.608)
                   (6,0.354) (7,0.287) (8,0.171) (9,0.000) (10,0.166) (11,0.641)
                   (12,0.624) (13,0.663)};
    \addplot[gtline, mark=*, mark size=0.9pt, mark options={solid, fill=figink}]
      coordinates {(0,0.279) (1,0.254) (2,0.625) (3,1.000) (4,0.854) (5,0.667)
                   (6,0.440) (7,0.309) (8,0.000) (9,0.121) (10,0.200) (11,0.217)
                   (12,0.615) (13,0.637)};
    \legend{simulation, ground truth}

  \nextgroupplot[xlabel={(b) Roe v.\ Wade}, yticklabels={}]
    \fill[figgray] (axis cs:-0.5,-0.05) rectangle (axis cs:0.6,1.05);
    \node[evlabel] at (axis cs:0.3,1.09) {Roe v.\,Wade\\[-1pt]overturned};
    \fill[figgray] (axis cs:9.7,-0.05) rectangle (axis cs:11.4,1.05);
    \node[evlabel] at (axis cs:10.5,1.09) {D.C.\ protest\\[-1pt]arrests};
    \addplot[simline, mark=*, mark size=0.9pt, mark options={solid, fill=figaccent}]
      coordinates {(0,0.000) (1,0.095) (2,0.121) (3,0.145) (4,0.256) (5,0.553)
                   (6,0.456) (7,0.628) (8,0.648) (9,0.568) (10,0.558) (11,0.771)
                   (12,1.000) (13,0.982)};
    \addplot[gtline, mark=*, mark size=0.9pt, mark options={solid, fill=figink}]
      coordinates {(0,0.000) (1,0.221) (2,0.208) (3,0.093) (4,0.328) (5,0.347)
                   (6,0.449) (7,0.372) (8,0.405) (9,0.217) (10,0.243) (11,0.325)
                   (12,0.561) (13,1.000)};

  \nextgroupplot[xlabel={(c) Black Lives Matter}, yticklabels={}]
    \fill[figgray] (axis cs:-0.5,-0.05) rectangle (axis cs:0.5,1.05);
    \node[evlabel] at (axis cs:0.4,1.09) {Floyd's\\[-1pt]death};
    \fill[figgray] (axis cs:1.6,-0.05) rectangle (axis cs:2.4,1.05);
    \node[evlabel] at (axis cs:2.0,1.24) {murder\\[-1pt]charges};
    \fill[figgray] (axis cs:5.6,-0.05) rectangle (axis cs:6.4,1.05);
    \node[evlabel] at (axis cs:6.0,1.09) {chokehold\\[-1pt]ban};
    \fill[figgray] (axis cs:7.6,-0.05) rectangle (axis cs:8.4,1.05);
    \node[evlabel] at (axis cs:8.4,1.24) {nationwide\\[-1pt]protests};
    \fill[figgray] (axis cs:10.6,-0.05) rectangle (axis cs:11.4,1.05);
    \node[evlabel] at (axis cs:11.2,1.09) {council\\[-1pt]support};
    \addplot[simline, mark=*, mark size=0.9pt, mark options={solid, fill=figaccent}]
      coordinates {(0,0.321) (1,0.000) (2,0.538) (3,0.462) (4,0.154) (5,0.090)
                   (6,1.000) (7,0.731) (8,0.731) (9,0.218) (10,0.667) (11,0.897)
                   (12,0.269) (13,0.218)};
    \addplot[gtline, mark=*, mark size=0.9pt, mark options={solid, fill=figink}]
      coordinates {(0,0.295) (1,0.332) (2,0.409) (3,0.151) (4,0.240) (5,0.505)
                   (6,0.286) (7,0.422) (8,0.425) (9,0.369) (10,0.622) (11,1.000)
                   (12,0.717) (13,0.000)};
\end{groupplot}
\end{tikzpicture}}
 \caption{Simulated and observed collective attitude trajectories over 14 steps for the three movements, min--max normalized. Shaded bands mark injected news events. Data from the representative runs of~\citet{mou2024unveiling}.}
 \label{fig:hisim-trajectories}
\end{figure}
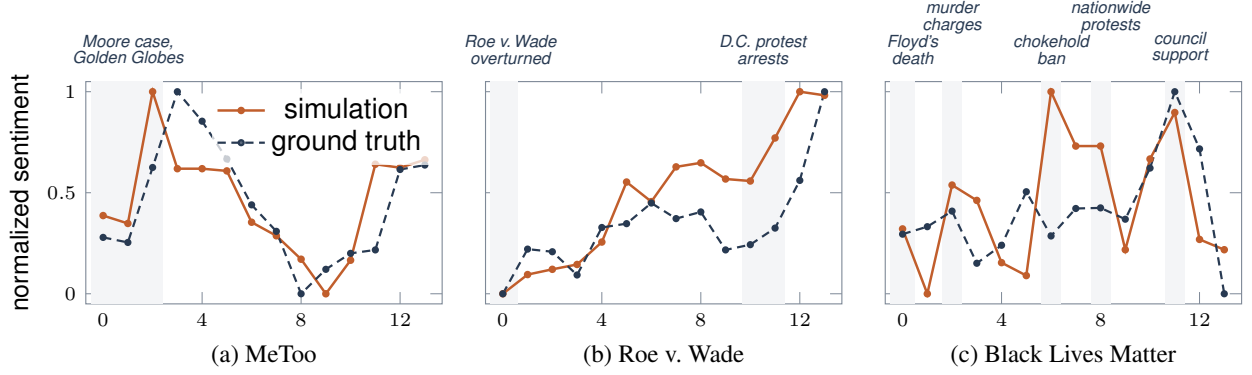

\paragraph{Results.}
\emph{Micro alignment: LLM-driven core users reproduce the positions of their real counterparts.}
At the individual level (\Cref{tab:hisim-micro}), the core users match the recorded responses of the people they simulate.
Stance accuracy lies between 0.90 and 0.97 across the three movements and content similarity exceeds 0.80, while the type of action a user takes (posting or retweeting) is predicted with accuracy between 0.67 and 0.78. Stance F1 is considerably lower than stance accuracy in all three movements.

\begin{table}[t]
 \centering
 \caption{Micro-level alignment of LLM core users against real user responses: stance (support, neutral, oppose), content type (five classes) and text similarity, and behavior type (post or retweet).}
 \label{tab:hisim-micro}
 \small
 \begin{tabular}{lcccccc}
 \toprule
 Scenario & \makecell{Stance\\Acc} & \makecell{Stance\\F1} & \makecell{Content\\Acc} & \makecell{Content\\Sim} & \makecell{Behavior\\Acc} & \makecell{Behavior\\F1} \\
 \midrule
 MeToo & 0.968 & 0.340 & 0.701 & 0.806 & 0.731 & 0.521 \\
 Roe v.\ Wade & 0.943 & 0.336 & 0.642 & 0.809 & 0.667 & 0.469 \\
 BLM & 0.899 & 0.374 & 0.735 & 0.841 & 0.780 & 0.576 \\
 \bottomrule
 \end{tabular}
\end{table}

\emph{Trajectory fidelity: coupling core users to the rule-based population improves the shape of the collective trajectory in every configuration.}
Across the fifteen model--scenario pairings reported in the original study, the hybrid variant attains higher correlation than its pure-ABM counterpart in all fifteen {(\Cref{tab:hisim-macro})}. The best hybrid configurations reach correlations of 0.724 (MeToo), 0.758 (\emph{Roe}), and 0.605 (BLM), against 0.483, 0.733, and 0.448 for the best pure-ABM model of each scenario.
The correction is most visible where the rule-based model alone misses the direction of change, as the two bounded-confidence rules do in the \emph{Roe} scenario.
Absolute-level bias does not improve uniformly. The Lorenz-based hybrid attains the lowest $\Delta$Bias in all three scenarios, whereas the relative-agreement hybrid trades a larger bias for the best correlation in two of them.

\begin{table}[t]
 \centering
 \caption{Macro-level trajectory evaluation on the validation window of each movement (mean of 3 runs). Corr: Pearson correlation with the observed attitude trajectory. $\Delta$Bias: absolute deviation of the mean attitude. Best pure-ABM is the rule-based model with the highest correlation per scenario. The hybrid rows are the configurations with the best correlation or the lowest bias.}
 \label{tab:hisim-macro}
 \small
 \begin{tabular}{lcccccc}
 \toprule
 & \multicolumn{2}{c}{MeToo} & \multicolumn{2}{c}{Roe v.\ Wade} & \multicolumn{2}{c}{BLM} \\
 \cmidrule(lr){2-3} \cmidrule(lr){4-5} \cmidrule(lr){6-7}
 Method & Corr & $\Delta$Bias & Corr & $\Delta$Bias & Corr & $\Delta$Bias \\
 \midrule
 Best pure-ABM & 0.483 & 0.0124 & 0.733 & 0.0352 & 0.448 & 0.0411 \\
 Hybrid w/ RA & \textbf{0.724} & 0.0117 & 0.427 & 0.0221 & \textbf{0.605} & 0.0376 \\
 Hybrid w/ Lorenz & 0.610 & \textbf{0.0035} & \textbf{0.758} & \textbf{0.0093} & 0.506 & \textbf{0.0023} \\
 \bottomrule
 \end{tabular}
\end{table}

\emph{The hybrid trajectories follow the direction and timing of the major attitude shifts in all three movements.} In \Cref{fig:hisim-trajectories}, the simulated series rises or falls with the observed one after the initial trigger, settles over the following steps, and moves again when a follow-up development is injected, which is the behavior the correlation summarizes.

\emph{Mechanism: the rule governing ordinary users is an experimental variable, and platform interventions register in the panel.}
No single rule fits every movement.
The Lorenz model gives the best bias and the best \emph{Roe} correlation, relative agreement gives the best correlation for MeToo and BLM, and the two bounded-confidence variants remain the weakest hybrids in the \emph{Roe} and BLM scenarios. The five interchangeable models therefore serve as an ablation of the ordinary-user mechanism, and the calibrate-in-rule, validate-in-hybrid protocol keeps this ablation affordable.
The three interventions on the MeToo scenario show that the same simulation supports platform-design questions: all three reduce the homogeneity between produced and consumed content, {and opening public discussion spaces does so at the lowest toxicity}~\citep{mou2024unveiling}.

\emph{Cost: fidelity is retained while the ordinary population grows, and the cost is pinned to the number of core users.}
In the scalability analysis of the original study, which fixes 300 core users and varies the number of ordinary users on the MeToo scenario, {the trajectory correlation} declines only slightly as the ordinary share increases, and runtime is dominated by LLM API latency, so that adding ordinary users imposes almost no additional cost until the population reaches the millions~\citep{mou2024unveiling}.

\paragraph{Takeaway.}
The hybrid-$f$ pattern reproduces observed opinion trajectories at a cost that grows with the number of LLM-driven opinion leaders and not with the population, because the mirror coupling carries the leaders' reasoning into a rule-based majority.
The case also shows why trajectories are the object of study: a snapshot records the attitude distribution in a cross-section, whereas the step-by-step record shows how that distribution moved after each injected development and each platform intervention.

\subsection{Case Family II: Policy Simulation}
\label{sec:cases:policy}

Policy simulation asks what follows from an intervention before it is administered.
The two cases in this family exercise the longitudinal machinery built for this question, populations grounded in real records and behavior that {evolves} with a changing environment, and they emphasize the mechanism probe of the template: ablations that isolate what drives the aggregate, and reference conditions that hold the environment fixed while the decision model changes.
The segregation case extends the Schelling model, validated as a synthetic benchmark in \S\ref{sec:cases:abm}, to real Chicago census data (781 tracts and calibrated household archetypes), shifting the ground truth from rules to the observed pattern of segregation (\S\ref{sec:cases:chicago}).
The procurement case models China's national drug procurement as a multi-agent Markov game among government, enterprises, and healthcare institutions, with historical procurement outcomes across multiple bidding rounds as the ground truth (\S\ref{sec:cases:procurement}).
{Both cases pose policy questions in the form of the longitudinal simulation loop: a policy is a declared intervention on $E$, and its effect is read as a contrast between branches that share a history, while the reference rules and ablations are versions that vary $f$ or remove one component.}

\subsubsection{Case Study 3: Chicago Segregation with Real Census Data}
\label{sec:cases:chicago}

\paragraph{Research Question.}
Residential segregation is among the most persistent structural features of American cities, and Chicago is its canonical case: in the 2010 Census the Black--White dissimilarity index across the city's 781 tracts stands at 0.835.
Schelling's model~\citep{schelling1971dynamic} explains how mild individual preferences generate such clustering, yet four properties of real urban segregation lie outside its representational scope: adjacency follows irregular physical geography, preferences are multi-group and asymmetric across racial groups, households weigh schools, safety, transit, affordability, and social ties alongside neighborhood composition, and the same aggregate index can arise from very different micro-level reasoning.
The case asks two questions: whether the Schelling dynamics validated as a synthetic sandbox in \S\ref{sec:cases:abm} extend to a real city with real census data, so that a demographically calibrated household panel released from a perturbed configuration returns to the observed pattern of segregation, and what the step-by-step record adds to the endpoint, namely whether the return follows a stable course across seeds, whether it stops at the empirical magnitude, and which elements of the decision context drive it.
The hypothesis is directional: LLM-driven households are expected to re-segregate toward the 2010 indices on every axis and to settle there, whereas households deprived of racial context in their prompt are expected to drift away from the pattern.

\paragraph{Framework Mapping.}
The case instantiates the framework on real urban geography: a census-calibrated household panel acts in the tract graph of Chicago, and residential decisions are produced by an LLM reasoning over each household's profile and neighborhood context.
\begin{itemize}
  \item \textbf{Population \Pop.} 241 household archetypes constructed from 2010 Census data along four dimensions (race, income bracket, family type, neighborhood type), compressing 2,654,858 residents into 19,235 weighted agents.
  Each archetype carries calibrated behavioral parameters (mobility rates, ideal own-group percentages, and satisfaction weights for seven factors: racial composition, housing cost, school quality, safety, transit, social anchor, and job proximity), grounded in nine published sociological studies.
  \item \textbf{Environment \Env.} The physical layer is a Queen-contiguity graph over the 781 census tracts, with per-tract attributes (racial composition, safety scores, school quality, transit access) updated endogenously as households move. The information layer carries policy broadcasts (citywide news) and local notices (tract-specific announcements) activated at designated steps.
  \item \textbf{Behavior function $f$.} An LLM-driven two-phase decision, \emph{satisfaction assessment} (``am I content here?'') followed by \emph{candidate evaluation} (``which neighborhood should I move to?''), with anti-overshoot controls (own-race ceiling filters, per-race move budgets with race-size scaling, per-tract inflow caps) that prevent runaway sorting.
  Each decision yields a typed action (stay or move to a named tract), a satisfaction score, and a natural-language rationale.
\end{itemize}
The model is integrated via Path~C (\S\ref{sec:infra:pipeline}): an engine-seam adapter of about 250 lines wraps the legacy segregation model (1,572 lines of Mesa-based code) with zero modification to the original source, implementing the environment provider, population provider, decision model, and metric collector on top of it, so that the legacy dynamics gain panel storage and intervention scheduling as they stand.

\paragraph{Experimental Design.}
\begin{itemize}
  \item \textbf{Ground truth and window.} The 2010 Census segregation indices and tract-level racial shares. Each run starts from the 2010 configuration with 15\% of households randomly displaced, which depresses the Black--White dissimilarity index from 0.835 to about 0.72 while leaving every other tract attribute unchanged; the census pattern is the attractor the dynamics are expected to recover over 15 steps.
  \item \textbf{Reference conditions.} Five rule-based references on the same tract graph, under the same per-step move budget, capacity limits, inflow caps, and execution machinery: random reassignment, classical Schelling with a single threshold $\tau{=}0.5$ under random and under best-improvement destination choice, a per-group threshold rule calibrated from the same archetype parameters the LLM receives, and a logit-style utility rule over the same satisfaction weights (\Cref{app:cases}), each run for ten perturbation seeds.
  \item \textbf{Longitudinal protocol.} 19,235 weighted agents over 781 tracts and 15 abstract steps, five perturbation seeds (42--46) with $t$-based 95\% confidence intervals on endpoint indices. Decisions use DeepSeek-V3 at temperature 0.7 with at most 512 output tokens; archetype-level batching issues one call per archetype--tract group, from which individual households draw their action, so that a full-city run issues approximately 275,000 calls (about 350 million input and 20 million output tokens) in place of one call per household per phase.
  \item \textbf{Mechanism probe.} Three ablations of the decision context on a contiguous 218-tract subset, each against a matched unablated control on the same subset: a race-blind prompt (five seeds), removal of the mobility constraints, and a shuffle of the non-racial tract attributes (three seeds each)~\citep{zhang2025abm}.
\end{itemize}
{The five references and the three ablations are versions in the sense of \Cref{eq:edit}: the references replace $f$, the race-blind prompt and the removal of the mobility constraints edit $f$, and the attribute shuffle edits $E_0$, each against a matched control that shares everything else.}

\paragraph{Results.}
\emph{Level fidelity: the LLM-driven population re-segregates toward the empirical 2010 pattern from a perturbed start.}
\Cref{tab:chicago-endpoints} reports the endpoint segregation indices averaged over the five seeds: $D_{BW}$ rises from 0.716 to 0.782 (target 0.835), $D_{HW}$ from 0.500 to 0.577 (target 0.608), and $D_{AW}$ from 0.423 to 0.435 (target 0.434), the last within 0.001 of the empirical value.
Measured against the perturbation gap, the trajectory closes roughly half of the Black--White gap and about seven tenths of the Hispanic--White gap within 15 steps, and all three dissimilarity endpoints fall within six percentage points of the census targets.
The geography is recovered as well as the magnitude: the dominant racial group of each tract at step 15 matches the 2010 Census across the South, West, and North Sides, with a tract-level coefficient of determination between simulated and empirical racial shares of 0.79.

\begin{table}[t]
  \begin{minipage}[c]{0.54\textwidth}
    \centering
    \caption{Endpoint segregation indices for the Chicago simulation (mean $\pm$ 95\% CI over 5 seeds). $D$ denotes dissimilarity, $P^*_B$ Black isolation, and $P^*_{BW}$ Black--White exposure.}
    \label{tab:chicago-endpoints}
    \footnotesize
    \begin{tabular}{lccc}
      \toprule
      Index & Perturbed ($t{=}0$) & LLM-ABM ($t{=}15$) & 2010 Census \\
      \midrule
      $D_{BW}$ & $0.716 \pm 0.007$ & $0.782 \pm 0.005$ & 0.835 \\
      $D_{HW}$ & $0.500 \pm 0.005$ & $0.577 \pm 0.004$ & 0.608 \\
      $D_{AW}$ & $0.423 \pm 0.006$ & $0.435 \pm 0.006$ & 0.434 \\
      $P^*_B$  & $0.658 \pm 0.004$ & $0.700 \pm 0.003$ & 0.797 \\
      $P^*_{BW}$ & $0.135 \pm 0.004$ & $0.104 \pm 0.002$ & 0.081 \\
      \bottomrule
    \end{tabular}
  \end{minipage}\hfill
  \begin{minipage}[c]{0.42\textwidth}
    \centering
    \captionof{figure}{Ablations on a 218-tract subset: net change of each index from step 0 to 15 (bars: means. Whiskers: 95\% CI). Adapted from~\citet{zhang2025abm}.}
    \label{fig:chicago-ablations}
    \includegraphics[width=\linewidth]{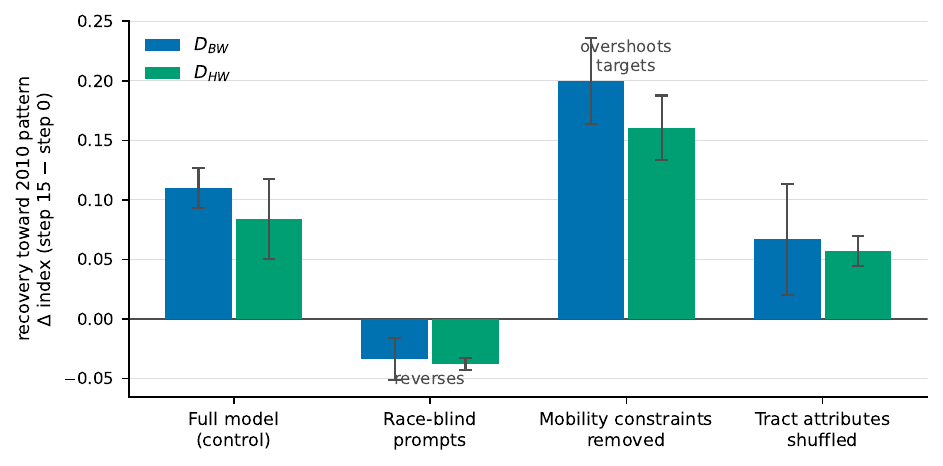}
  \end{minipage}
\end{table}

\begin{figure}[t]
  \centering
  \includegraphics[width=\textwidth]{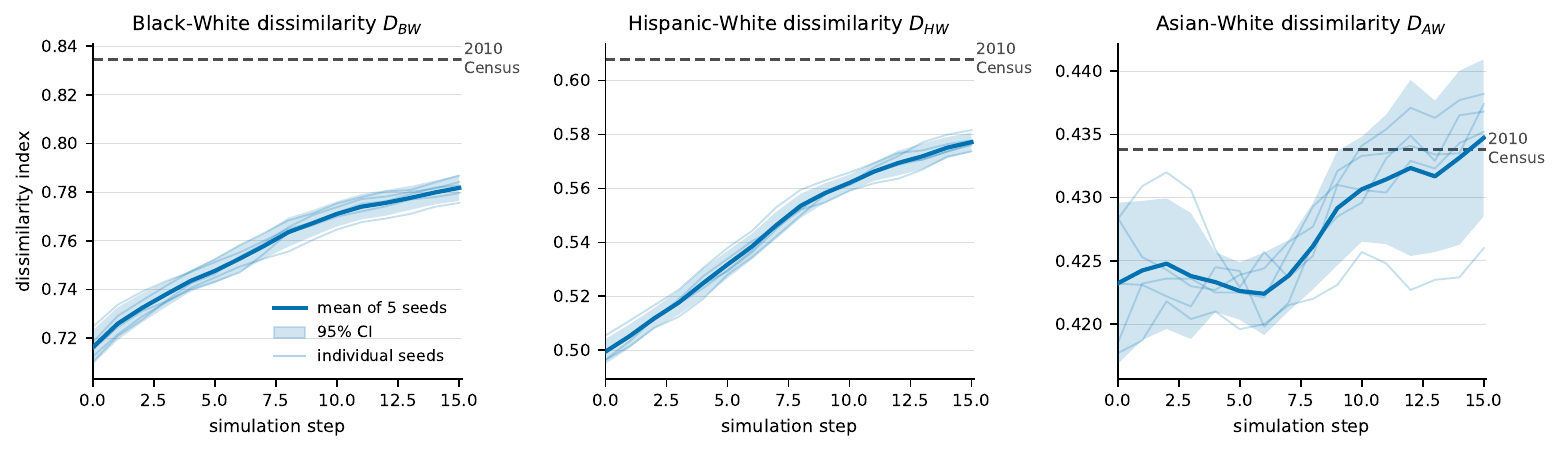}
  \caption{Trajectories of the three dissimilarity indices over 15 steps. Bold lines give the mean of 5 seeds with 95\% CI, thin lines individual seeds, and dashed lines the 2010 Census values.}
  \label{fig:chicago-trajectories}
\end{figure}

\emph{Trajectory fidelity: the recovery is monotone and stable across seeds on the two large-gap axes and self-limiting on the small-gap axis.}
\Cref{fig:chicago-trajectories} shows the Black--White and Hispanic--White indices rising steadily from the first step with confidence bands narrow enough that individual seeds are hard to distinguish, whereas the Asian--White index fluctuates within a wider band for the first eight steps and then converges on the census value without continued drift.
The population also settles: the number of movers per step falls from about 1,050 in the first step to about 250 by step 15, with under two percent of the represented population relocating in an average step, and between 60\% and 76\% of moves increase the mover's own-group share, a clear yet imperfect own-race preference produced endogenously by the behavior function.

\emph{Mechanism: the LLM behavior function is the only condition that recovers both large-gap axes while holding the small-gap axis at its empirical value, and the recovery depends on racial context and on the mobility constraints.}
On the same tract graph, move budget, and capacity limits, random reassignment collapses $D_{BW}$ to ${\sim}0.05$. Classical Schelling ($\tau{=}0.5$) holds $D_{BW}$ at 0.743 with random destinations and reaches 0.788 with best-improvement destinations. The per-group threshold rule overshoots $D_{BW}$ to $0.875$ and also overshoots $D_{AW}$, and the logit-style utility rule reaches $D_{BW}{=}0.839$ but overshoots Asian--White segregation in the same way, while every rule-based condition keeps relocating at the budget cap through step 15.
The joint endpoint of the LLM condition ($D_{BW}{=}0.78$, $D_{HW}{=}0.58$, $D_{AW}{=}0.44$) lies in a region of the multi-group dissimilarity space that a single similarity statistic cannot reach without group-specific hand tuning.
The ablations locate the source of this behavior (\Cref{fig:chicago-ablations}): a race-blind prompt reverses recovery ($\Delta D_{BW} = -0.033$ against $+0.110$ for the matched control, a swing of 0.14), removing the mobility constraints overshoots the census targets, and shuffling the non-racial tract attributes retains about 60\% of the recovery, so the racial context supplies the direction, the constraints supply the stopping point, and the remaining attributes modulate the pace.
The decision rationales expose the micro-mechanism behind the macro pattern: a White lower-middle-income family reports ``demographic familiarity (White residents below preferred threshold)'' alongside ``safety is very high (9.6/10), offsetting some demographic discomfort''. A Hispanic family in a comparable tract instead emphasizes ``strong social network proximity due to high Hispanic presence''. A Black middle-income family weighs ``moderate Black presence (19.8\%)'' that ``provides some demographic familiarity but is below their ideal'' against the higher Black share of surrounding tracts; and Asian archetypes prioritize school quality.
These documented preference asymmetries arise without any hand-coded preference rule, an interpretability through generation that threshold models cannot offer.

\emph{Boundary: the recovery is bounded by the mobility budget and the horizon.}
Half of the Black--White gap remains open after 15 steps while the movers per step have already fallen to about a quarter of their initial number, so the remaining distance to the census value reflects the calibrated mobility rates, the direction being set. The per-group and logit rules that do reach the Black--White target do so by overshooting the Asian--White axis.

\paragraph{Takeaway.}
Classical Schelling dynamics, reproduced on a synthetic grid in \S\ref{sec:cases:abm}, extend to real census geography with calibrated archetypes: the LLM-driven panel returns to the empirical magnitude and geography of segregation on every axis, and the step-by-step record shows a stable, self-limiting course whose direction is set by racial context and whose stopping point is set by the mobility constraints.
Brought in through a zero-edit adapter, the same study declares transit and housing interventions as scheduled events and broadcasts, which is the form in which the policy questions of this family are posed to the engine.

\subsubsection{Case Study 4: Drug Procurement as a Multi-Agent Markov Game}
\label{sec:cases:procurement}

\paragraph{Research Question.}
China's National Volume-Based Drug Procurement program, launched in 2018, has covered more than 490 drugs and thousands of participating firms and is credited with roughly 500 billion CNY in pharmaceutical savings to date~\citep{wang2026procuregymmultiagentmarkovgame}.
The program operates as a centralized competitive procurement mechanism in which the government sets procurement rules and price ceilings and consolidates aggregate demand, pharmaceutical enterprises formulate bids under production-cost, market-share, and capacity constraints and incomplete information, and medical institutions report anticipated demand and subsequently procure from the selected suppliers.
Because every stakeholder adapts its strategy to the rules in force, the effect of a rule change on prices, supplier selection, and clinical supply is difficult to estimate from historical regressions alone, and existing simulations treat enterprises as the only strategic actors while holding government and institutional behavior fixed~\citep{wang2026procuregymmultiagentmarkovgame}.
This case asks whether a simulation that assigns learned strategic behavior to all three stakeholder classes reproduces the outcomes of seven historical procurement rounds, and whether modeling the three actors jointly yields procurement outcomes and winner sets closer to the historical record than enterprise-only models.
Each stakeholder is given a distinct objective: the government maximizes social welfare. Enterprises balance the greater probability of selection associated with lower bids against the corresponding reduction in per-unit margins, and medical institutions maximize a composite utility that captures both the fulfillment of clinical demand and the financial benefits arising from the medical-insurance fund savings-retention mechanism.
The hypothesis is directional: joint learning is expected to raise the outcome of every stakeholder class relative to enterprise-only models while keeping the selected supplier sets at least as close to the historical record.

\paragraph{Framework Mapping.}
The case instantiates each historical drug--round market as an independent longitudinal simulation in which three classes of institutional actors interact through one complete procurement round per step, and it exercises the reinforcement-learning end of the framework's behavior-function spectrum.
\begin{itemize}
  \item \textbf{Population \Pop.} Three classes of institutional actors participate in each drug-specific task: one government actor, one medical-institution actor, and a set of heterogeneous pharmaceutical enterprise actors whose costs, capacities, and market positions are drawn from the historical record of that drug and round.
  \item \textbf{Environment \Env.} Each drug--round market is instantiated from historical procurement records, combining the applicable procurement rules, drug characteristics, bidder composition, and demand conditions with an evolving state. At every step, the environment performs market clearing, quantity and price settlement, role-specific reward computation, and the state transition that conditions the next round.
  \item \textbf{Behavior function $f$.} Role-specific reinforcement-learning policies map local observations to actions: Proximal Policy Optimization (PPO) for the government and the medical institution and Multi-Agent Proximal Policy Optimization (MAPPO) for enterprises, trained under centralized training and decentralized execution. The resulting joint behavior comprises the government's policy parameters, the enterprises' strategic bids, and the institution's demand reports and procurement execution.
\end{itemize}

\paragraph{Experimental Design.}
\begin{itemize}
  \item \textbf{Ground truth and window.} The historical record of seven rounds of China's National Volume-Based Drug Procurement program (Rounds~2--5 and~7--9), covering 325 drugs and 2,267 enterprises. Each observed drug--round market is instantiated as an independent simulation and policy-optimization task, and the historically selected enterprise set of each market is the reference for winner-set fidelity.
  \item \textbf{Reference conditions.} Three enterprise-only ProcureGym baselines in which pharmaceutical enterprises are the sole strategic class and adopt RL-, LLM-, or rule-based bidding controllers~\citep{wang2026procuregymmultiagentmarkovgame}, and a random-uniform baseline averaged over five independent runs per task.
  \item \textbf{Longitudinal protocol.} One government actor, one medical-institution actor, and the historical set of enterprise actors per task. One complete procurement round per step and 50 steps per episode. The synchronous environment lets all actors act on their current observations, after which market clearing, quantity and price settlement, role-specific rewards, and the state transition follow. Each task is trained for 10,000 episodes with seed 42. The selected checkpoint is evaluated over 1,000 deterministic episodes for the aggregate outcomes, and a fixed-seed replay of the same checkpoint produces the canonical joint strategy with its step-level decision trajectory. Checkpoint selection, network architecture, and optimization hyperparameters are documented in \Cref{app:cases}.
\end{itemize}

\begin{figure}[!t]
\centering
\includegraphics[width=\linewidth]{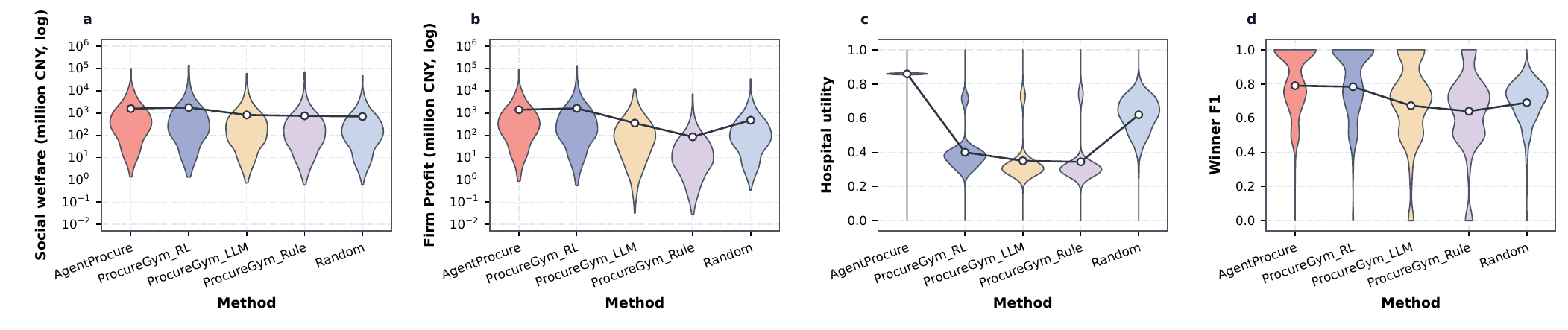}
\caption{Performance across 325 drugs and seven procurement rounds: the three-actor simulation against the enterprise-only ProcureGym baselines (RL, LLM, Rule) and Random on (a) government social welfare, (b) enterprise profit, (c) medical-institution utility, and (d) F1 agreement with the historical winner sets.}
\label{fig:procurement-overall}
\end{figure}

\paragraph{Results.}
\emph{Level fidelity: joint learning across the three stakeholder classes raises the median outcome of every stakeholder at once.}
Across 325 drug-level tasks, median government social welfare, enterprise profit, and medical-institution utility exceed those of both the enterprise-only ProcureGym RL baseline and the random baseline, as shown in~\Cref{tab:procurement-outcomes}.
The welfare and profit distributions span several orders of magnitude across drugs, and those of the two reinforcement-learning methods overlap substantially. The advantage of the three-actor simulation appears as an upward shift of the distribution center in both panels, so the gain from adding government and institutional decisions accrues alongside enterprise profit, as shown in~\Cref{fig:procurement-overall}(a)--(b).

\begin{table}[t]
  \centering
  \caption{Median stakeholder outcomes across 325 drug-level tasks.
  Social welfare and profit are in million CNY.
  Medical-institution utility combines fulfilment of clinical needs with financial benefits from retaining a share of health insurance fund savings and is normalized to a maximum of 1.
  Bold denotes the highest median in each column.
  Gains over ProcureGym RL are relative for welfare and profit and absolute for utility.}
  \label{tab:procurement-outcomes}

  \small
  \setlength{\tabcolsep}{5pt}
  \renewcommand{\arraystretch}{1.12}

  \begin{tabular}{@{}lccc@{}}
    \toprule
      & Government
      & Enterprise
      & Medical institution \\
    \cmidrule(lr){2-2}
    \cmidrule(lr){3-3}
    \cmidrule(l){4-4}
    Method
      & Social welfare
      & Profit
      & Utility \\
    \midrule
    Random
      & 117.96
      & 79.48
      & 0.629 \\
    ProcureGym RL
      & 228.82
      & 188.99
      & 0.382 \\
    \addlinespace[2pt]
    \textbf{Ours}
      & \textbf{291.02}
      & \textbf{252.97}
      & \textbf{0.859} \\
    \midrule
    Ours vs. ProcureGym RL
      & $+27.2\%$
      & $+33.9\%$
      & $+0.477$ \\
    \bottomrule
  \end{tabular}
\end{table}

\emph{Medical-institution utility is the outcome on which the three-actor design departs most sharply from enterprise-only modeling.}
In \Cref{fig:procurement-overall}(c), the three-actor simulation concentrates institutional utility tightly around its median of 0.859, whereas the enterprise-only baselines disperse widely around far lower medians (0.382 for ProcureGym RL), and even the random baseline reaches only 0.629.
In the enterprise-only baselines, the medical institution takes no decisions, so its utility is a by-product of enterprise bidding under fixed demand reports. Assigning the institution a learned demand-reporting and execution policy allows this objective to be optimized directly and consistently across drugs.

\emph{Winner-set fidelity is preserved while stakeholder outcomes improve.}
The three-actor simulation attains the highest mean winner-set F1 score against the historically selected enterprise sets, 0.790 in \Cref{fig:procurement-overall}(d), exceeding ProcureGym RL by 0.006, with the LLM, rule-based, and random controllers lower still.
The ordering among the enterprise-only controllers is consistent with the selection-accuracy evaluation reported for ProcureGym, in which RL bidding controllers reached 75\% selection accuracy against 66\% for LLM and 64\% for rule-based controllers~\citep{wang2026procuregymmultiagentmarkovgame}.
That the F1 gain is small while the stakeholder gains are large indicates that government and institutional policies reshape prices, quantities, and utilities within largely the same set of selected suppliers.

\emph{Mechanism: the simultaneous improvement across the three objectives follows from the checkpoint-selection rule.}
As detailed in \Cref{app:cases}, candidate checkpoints are first restricted to the Pareto-nondominated set defined by mean social welfare, aggregate enterprise profit, and institutional utility, and the final checkpoint is selected by a balanced lexicographic criterion that prioritizes the lowest normalized stakeholder objective. The reported medians therefore describe policies that no other checkpoint dominates on all three objectives.

\paragraph{Takeaway.}
The framework's behavior function accommodates learned strategic policies as readily as rules or language-model reasoning, and a regulated market is tracked as a longitudinal process in which government rule setting, enterprise bidding, and institutional demand reporting co-evolve round by round, with every stakeholder's outcome improved and the selected supplier sets kept aligned with the historical record.
The same environment admits shocks to a single policy or scenario variable, which the framework expresses as scheduled events on $E$, with the maximum valid bidding price and the procurement volume identified as the levers that dominate strategic outcomes~\citep{wang2026procuregymmultiagentmarkovgame}, which extends the policy-sandbox paradigm from urban interventions to market mechanism design.

\subsection{Case Family III: Macro-Index Forecasting}
\label{sec:cases:macro}

Macro-index forecasting tests whether aggregate economic quantities can be reconstructed bottom-up from agent-level decisions.
The three cases in this family evaluate simulated aggregates against official statistics and market data over multi-year windows, and they emphasize the ground-truth element of the template: point-in-time evaluation windows, held-out periods after the response model's knowledge cutoff, and ablations of the state that agents carry from month to month.
The household case nowcasts the Consumer Confidence Index from a simulated household population (\S\ref{sec:cases:consumersim}), and the firm case nowcasts the Purchasing Managers' Index from a simulated firm panel (\S\ref{sec:cases:pmi}). Both rely on the point-in-time guarantees of the \eventtool (\S\ref{sec:infra:signals}), so that each simulated month sees only the information available at the corresponding real-world date.
The market case extends the same bottom-up approach from official indices to brand-level market shares in the German passenger-car market, validated against real vehicle registration data (\S\ref{sec:cases:marketsim}).
{The held-out windows after the response model's knowledge cutoff are where the foundation basis is tested most directly, since the base model cannot have memorized the target and what remains is the grounded population and environment supplied by the infrastructure. The ablations of the carried state are versions that remove one component of the study state at a time.}

\subsubsection{Case Study 5: Nowcasting the Consumer Confidence Index from Simulated Households}
\label{sec:cases:consumersim}

\paragraph{Research Question.}
 Consumer confidence measures summarize households’ perceptions and expectations regarding their financial circumstances, economic conditions, and major purchases. The target series are the University of Michigan’s Index of Consumer Sentiment for the United States, the European Commission’s Consumer Confidence Indicator for the European Union (EU27), and the Consumer Confidence Index published by the Economic and Social Research Institute (ESRI) of Japan’s Cabinet Office. For notational convenience, we refer to these measures collectively as CCI series. The published index aggregates
categorical judgments made by households with different resources, exposures,
and attention, and it moves abruptly when salient events enter public
view~\citep{huang2026consumersim}, yet it is usually modeled as one persistent
time series.  This case asks whether the official CCI can instead be
reconstructed from the bottom up: heterogeneous households encounter the
information available in each month, answer survey-like questions, and
collectively generate the observed index trajectory, and it asks what the
household panel adds to the headline series, in the months of abrupt change
and in the identity of the households that move.
The hypothesis is directional: the simulated index is expected to track the
official series more closely than aggregate time-series baselines, with the
largest advantage in high-salience months, and the panel is expected to show
distinct signal sensitivities across income, housing, and political groups.

\paragraph{Framework Mapping.}
The case instantiates the framework with a survey-calibrated household
population, a month-by-month information environment, and a behavior function
that produces survey responses.
\begin{itemize}
  \item \textbf{Population \Pop.}  A microdata-calibrated synthetic population:
  a fixed core of 5,000 households sampled from the Survey of Income and Program
  Participation with its survey weights (age, sex, race and ethnicity,
  education, homeownership, residence, income, employment), with behavioral
  attributes such as financial satisfaction, future optimism, and social trust
  imputed from General Social Survey conditional distributions. Core responses
  are expanded to a representative normal population by demographic cell.
  \item \textbf{Environment \Env.}  A point-in-time Situational Signal Field
  assembling the macroeconomic, financial, labor, housing, policy, and news
  information available before each month's survey cutoff, matched to the
  households it exposes (mortgage rates to homeowners, equity movements to
  asset holders) and decayed over time so that recent and visible signals carry
  more weight.
  \item \textbf{Behavior function $f$.}  A GPT-4o survey-response kernel maps
  each household and month to a probability distribution over the response
  categories of the region's official CCI questionnaire. Post-stratified
  Bayesian belief expansion propagates core responses to the full population,
  and Behavioral Inertia Alignment blends the behavioral forecast with the
  previous month's official value through a region-specific weight.
\end{itemize}

\paragraph{Experimental Design.}
\begin{itemize}
  \item \textbf{Ground truth and window.} The official monthly CCI of each region, compared within region because the three indices use different scales (\Cref{tab:consumersim-regional-config}). The reported backtest spans January~2020--March~2026 (75 months), with target-month CCI values excluded from the response environment. The underlying study additionally reports a three-month U.S.\ diagnostic on April--June~2026, after the response model's March~2026 knowledge cutoff.
  \item \textbf{Reference conditions.} Twelve aggregate baselines fitted on the same information set: autoregressive ridge, expectations regressions, a news-semantic factor proxy, rolling means, SARIMAX, exponential smoothing, Theta, Auto ARIMA, Prophet variants, and macro, market, and news ridge regressions~\citep{huang2026consumersim}.
  \item \textbf{Longitudinal protocol.} A fixed core of 5,000 households answering the region's official question battery every month, expanded to a normal population of 250,000 households (the $50\times$ setting at which the study's scale sweep reaches its accuracy plateau) and post-stratified by demographic cell. Behavioral Inertia Alignment blends each month's behavioral forecast with the previous month's official value through a region-specific weight selected by validation error on January--June~2024 and then frozen (0.4, 0.6, and 0.7 for the U.S., EU27, and Japan). One calibrated run per region.
  \item \textbf{Mechanism probe.} A U.S.\ ablation removing, in turn, post-stratified belief expansion, inertia alignment, the Situational Signal Field, and the persona substrate, together with a population-scale sweep from $5\times$ to $100\times$ expansion, both reported by the study on an evaluation window separate from the 75-month backtest.
\end{itemize}
{The ablation and the scale sweep are versions that remove or resize one component of the study state, and the monthly signal field is a pre-materialized timeline of broadcasts resolved at build time through the \eventtool.}

\paragraph{Results.}
\emph{Level fidelity: the bottom-up reconstruction tracks the official index more closely than
every aggregate baseline in all three regions.}
Against the twelve baselines, the simulation ranks first on all four reported metrics in each
region: MAE/RMSE values are $3.45/4.56$ for the United States, $1.607/2.737$ for the EU27, and $1.363/2.040$ for Japan. Pearson correlations are $0.9590$, $0.882$, and $0.824$, and Spearman correlations are $0.9519$, $0.849$, and $0.840$, respectively~\citep{huang2026consumersim}.  The EU27 and Japan runs
reuse the U.S. response architecture with region-matched survey batteries and
signal fields, so the ranking holds without a full local-microdata rebuild.

\begin{figure}[t]
  \centering
  \includegraphics[width=\linewidth]{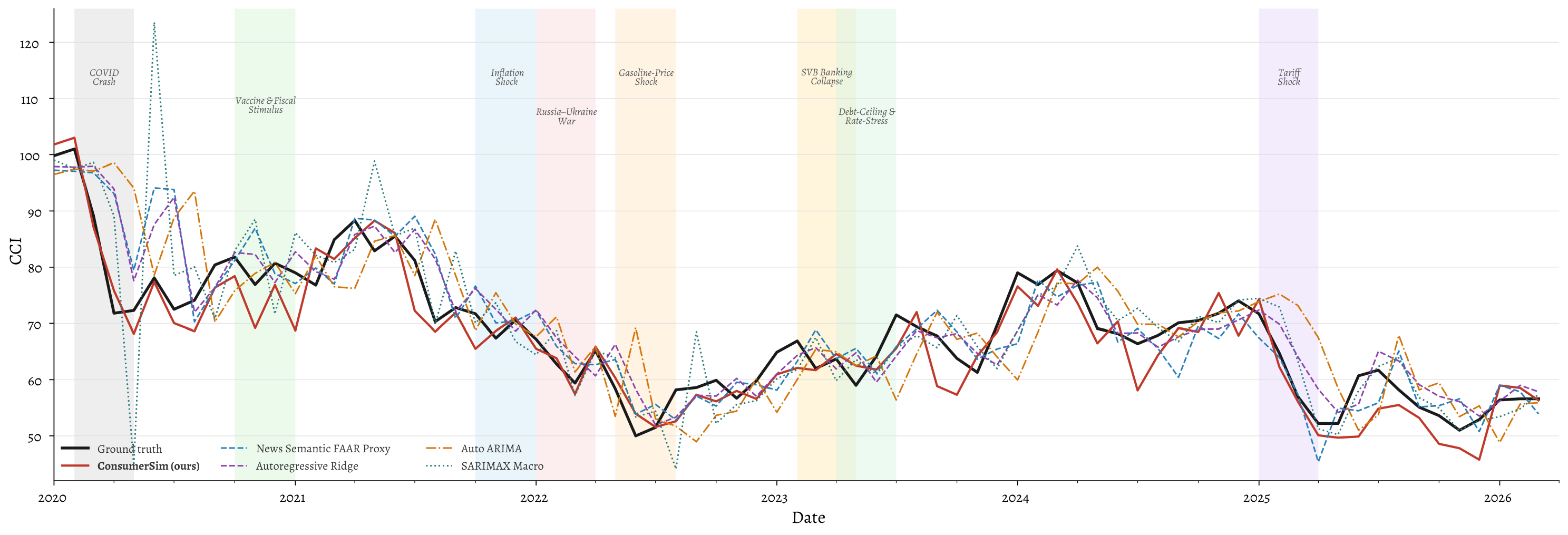}
  \caption{Official U.S. CCI, the simulated index, and representative baselines over January~2020--March~2026, with high-salience regimes annotated. Reproduced from~\citet{huang2026consumersim}.}
  \label{fig:consumersim-cci}
\end{figure}

\emph{Trajectory fidelity: the advantage concentrates in the months when salient shocks pull confidence
away from smooth persistence.}  \Cref{fig:consumersim-cci} shows the U.S.
series with high-salience regimes annotated. The simulated index follows
several abrupt changes that aggregate baselines smooth or lag.  Across the
19 annotated windows between December~2014 and February~2025, each evaluated over the preceding, event, and following
month, the simulation ranks first among the 14 methods of the study's shock-window comparison on both MAE and RMSE, and it holds
the first position individually at the COVID crash, the 2021 inflation shock,
the outbreak of the Russia--Ukraine war, the SVB banking collapse, and the 2025
tariff shock.

\emph{Mechanism: the accuracy depends on every component of the mapping, and it saturates once
the normal population is large enough.}
In the study's U.S.\ ablation, reported on its own evaluation window so that its absolute errors are not those of the 75-month backtest, the full model records an MAE of 1.36. Removing
post-stratified belief expansion raises it to 2.03, removing
Behavioral Inertia Alignment to 2.53, and removing the Situational
Signal Field or the persona substrate to 3.35 and 3.48, so the
environment and population components matter most for aggregate fit.  With the
5,000-household core fixed, expanding coverage from $5\times$ to $50\times$
reduces MAE from 6.10 to 1.72 and raises Pearson $r$ from 0.682 to 0.979, and
the $100\times$ run changes these values only to 1.70 and 0.980
(\Cref{app:cases}), placing the $50\times$ configuration at the
accuracy--scale plateau.

\emph{The panel carries information beyond the headline series, and it identifies who responds to what.}
Replacing official CCI with the simulated index in an otherwise matched
housing model raises one-month-ahead $R^2$ by 0.055 and lowers out-of-sample
RMSE by 7.8\%, with gains of 0.023 and 0.017 in $R^2$ at two- and three-month
horizons across new home sales, housing starts, and building permits. Gains
for vehicle sales and durable goods are weaker.  At the group level, the panel
identifies who responds to what: the bottom income quartile is most sensitive
to broad macro, price, and fiscal shocks, homeowners to housing and monetary
signals, renters to gasoline prices, the top income quartile to
financial-sector stress, and Republican-aligned households to geopolitical and trade-policy news,
with cross-group directional alignment from 0.53 (financial-sector shocks) to
0.97 (broad macro shocks).

\emph{Boundary: the results hold beyond the response model's knowledge cutoff.}
On April--June~2026, the three
months after the March~2026 cutoff, the simulation records MAE/RMSE of $2.74/3.10$ and
$r=0.971$, against $3.53/3.98$ and $0.914$ for the strongest baseline. This
diagnostic contains three observations, each region reports a single
calibrated run (\Cref{app:cases:protocol}), and all figures are reconstruction
and prediction results without causal attribution to the annotated events.

\paragraph{Takeaway.}
A population aligned to survey microdata and a time-stamped information environment turn an aggregate expectation index into an interpretable longitudinal response process that tracks the official series in three regions and leads the aggregate baselines in the months of abrupt change.
Its distinctive output is a diagnostic account of who updates, in response to what, and with what persistence, which the headline index cannot supply.

\subsubsection{Case Study 6: Nowcasting the Purchasing Managers' Index from Simulated Firms}
\label{sec:cases:pmi}

\paragraph{Research Question.}
The U.S. ISM Manufacturing Purchasing Managers' Index (PMI) is a survey-based
diffusion index: purchasing managers report each month whether new orders,
production, employment, supplier deliveries, and inventories improved,
remained unchanged, or deteriorated, and the equal-weight average of the five
component diffusion indices forms the headline, with 50 marking the expansion
threshold.  Because the index is assembled from individual firm responses,
it offers a natural test of bottom-up macro-index forecasting.  This case asks
whether the headline PMI can be predicted by reconstructing that survey
process: simulated firms evolve under the information available at the time,
their procurement managers answer the survey items, and the responses are
aggregated into a monthly headline index, and it asks whether the firms'
carried state, their memory and the inertia of their reported level, is what
makes the reconstruction informative.
The hypothesis is directional: the simulated panel is expected to call the
sign of the monthly change more often than persistence and market consensus,
including after the response model's knowledge cutoff, and removing the
mechanisms that carry state across months is expected to degrade the level
path most.

\paragraph{Framework Mapping.}
The case instantiates the framework with a persistent firm panel, a
point-in-time environment, and a hybrid structural--LLM behavior engine that
reproduces the survey response step.
\begin{itemize}
  \item \textbf{Population \Pop.} A persistent panel of 300 synthetic
  manufacturing firms aligned to industry, size, and regional statistics.
  Each firm carries heterogeneous exposure to exports, input costs, interest
  rates, and supply chains, as well as a manager-specific reporting style.
  \item \textbf{Environment \Env.} A monthly sequence of vintage-safe
  industrial, orders, financial, regional-survey, and news signals, restricted
  to what was available by the 21st day of each month.
  \item \textbf{Behavior function $f$.} A hybrid engine that first updates
  five latent operating states (new orders, production, employment, supplier
  deliveries, and inventories) and then asks an LLM procurement manager to
  answer the corresponding five survey items.  Hard responses are
  post-stratified and converted to diffusion indices, whose equal-weight
  average yields the simulated headline PMI.
\end{itemize}

\paragraph{Experimental Design.}
\begin{itemize}
  \item \textbf{Ground truth and window.} The first-print ISM Manufacturing PMI. Parameters are fitted on 2015--2018 and model development is evaluated on 2019--2021. Two held-out windows cover January~2022--October~2023 (clean test) and November~2023--June~2026 (32 months after the configured October~2023 knowledge cutoff of the response model). All historical inputs pass through an as-of-date signal store, and PMI or ISM index values are excluded from the prompt.
  \item \textbf{Reference conditions.} Persistence, a market-consensus proxy, a pre-fitted Bates--Granger ensemble of the simulation with consensus, the predecessor response kernel, and a mechanical-threshold twin (\Cref{tab:pmi-full-heldout}).
  \item \textbf{Longitudinal protocol.} A fixed panel of 300 firms warmed up from 2014 and followed monthly from 2015, one survey response per firm per month, with a three-month memory warm-up preceding each evaluation window. A single frozen configuration with the fitted change scale, update step, and ensemble weight held fixed. The case follows the build-from-scratch path (Path~B in \S\ref{sec:infra:pipeline}).
  \item \textbf{Mechanism probe.} Ablations on the 2019--2021 development window with a fixed 165-firm subpanel: without longitudinal memory, without inertia alignment, without post-stratification, without the news field, without the manager memorandum, with a homogeneous firm persona, and with a deterministic response kernel (\Cref{tab:pmi-ablation-full}).
\end{itemize}
{Each ablation is a version of the study state that removes one component while the panel, the seed, and the signal vintages are held fixed.}

\paragraph{Results.}
\emph{Level fidelity: the simulated firm panel retains predictive value after the knowledge cutoff
of its response model.}  On the post-cutoff window
(\Cref{tab:pmi-post-cutoff}), the main simulation signal, the agent-based
change nowcast, obtains a mean absolute error (MAE) of 0.892 PMI points,
compared with 0.894 for persistence and 0.922 for the market-consensus
baseline, and its month-over-month direction accuracy of 58.6\% exceeds the
51.7\% of consensus.  Since the LLM cannot have memorized this window, the
margin over persistence and consensus is attributable to the simulated
survey process itself.

\emph{Trajectory fidelity: the simulation alone makes the best directional calls after the cutoff, and combined with consensus it yields the strongest configuration on the clean test window.}
In \Cref{tab:pmi-full-heldout},  on January~2022--October~2023 the pre-fitted
Bates--Granger ensemble records an MAE of 0.928, an RMSE of 1.151, a
direction accuracy of 80.0\%, and a correct expansion/contraction call in
100.0\% of months, improving on both consensus (MAE 0.936, direction 75.0\%)
and the simulation alone (MAE 1.025, direction 65.0\%).  After the cutoff, the
ensemble attains the lowest RMSE (1.188) and, together with consensus, the
highest side-of-50 accuracy (90.6\%), while the simulation alone keeps the
lowest MAE and the best directional calls.

\emph{Mechanism: the level path depends on the state the firms carry across months.}
In the ablations of \Cref{tab:pmi-ablation-full}, removing inertia alignment
produces the largest level-path degradation ($+1.037$ MAE relative to the
full model) and removing the firms' longitudinal memory the second largest ($+0.253$),
whereas removing post-stratification, the news field, the manager memorandum, or
firm-persona heterogeneity shifts the change-path MAE by at most a few hundredths
of a point. The two mechanisms that carry state from one month to the next are
thus the ones the level path rests on.

\begin{figure}[t]
  \centering
  \includegraphics[width=0.8\linewidth]{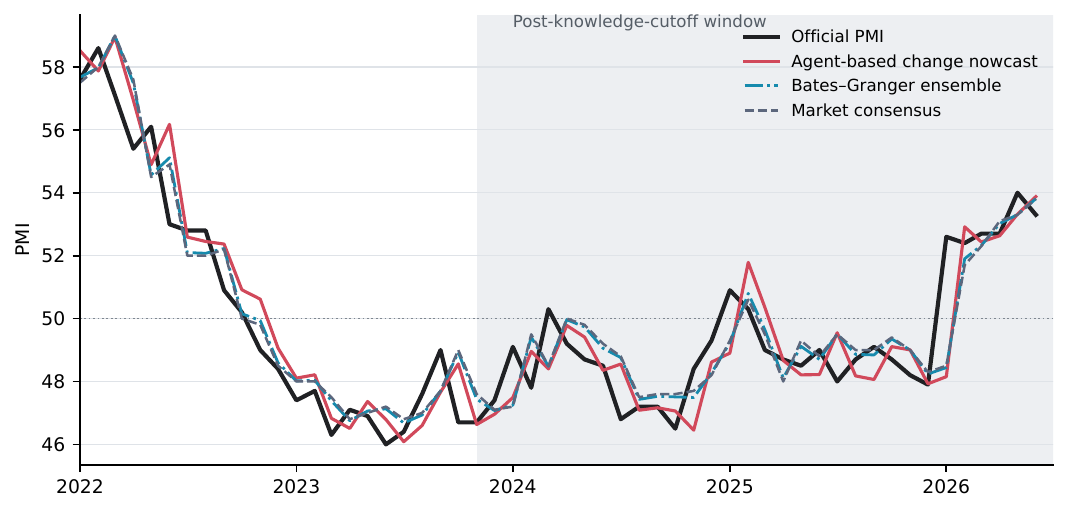}
  \caption{Official first-print ISM Manufacturing PMI and the principal
  nowcasting series from January~2022 to June~2026.  The shaded interval is the
  post-knowledge-cutoff evaluation window. The horizontal line marks the
  expansion threshold of 50.}
  \label{fig:pmi-trends}
\end{figure}

\emph{Boundary: the series in \Cref{fig:pmi-trends} locate where the signal is weakest.}
The blind window contains 32 months, the
consensus series is a proxy assembled from a public data source, and the
simulator tracks the repeated 2025 tariff shocks less closely than the rest of the window.

\begin{table}[t]
  \centering
  \caption{PMI nowcasting performance after the configured LLM knowledge
  cutoff (November~2023--June~2026, 32 months).  Direction measures whether a
  method correctly calls the sign of the month-over-month change. Persistence
  makes no directional call.}
  \label{tab:pmi-post-cutoff}
  \small
  \begin{tabular}{lrrrr}
    \toprule
    Method & MAE & RMSE & Direction & Side of 50 \\
    \midrule
    Agent-based change nowcast & \textbf{0.892} & 1.256 & \textbf{58.6\%} & 87.5\% \\
    Bates--Granger ensemble & 0.898 & \textbf{1.188} & 51.7\% & \textbf{90.6\%} \\
    Market consensus & 0.922 & 1.202 & 51.7\% & \textbf{90.6\%} \\
    Persistence & 0.894 & 1.279 & --- & 84.4\% \\
    \bottomrule
  \end{tabular}
\end{table}

\paragraph{Takeaway.}
A persistent firm panel, a point-in-time-safe environment, and a hybrid structural--LLM behavior engine produce a signal that carries information beyond aggregate time-series extrapolation, including after the response model's knowledge cutoff, and the ablations attribute the level path to the state the firms carry from month to month.
The panel supports firm-, industry-, and region-level diagnosis of an aggregate forecast, so the simulation serves as an incremental nowcasting component alongside official survey measurement and market consensus.

\subsubsection{Case Study 7: Brand Competition in the German Passenger-Car Market}
\label{sec:cases:marketsim}

\paragraph{Research Question.}
Monthly brand shares of German private passenger-car registrations, published
by the German Federal Motor Transport Authority (KBA), summarize the outcome of
household purchase decisions taken under replacement cycles, budget
constraints, product preferences, external events, and peer influence.
This case asks how the shares of established German manufacturers, Tesla,
BYD, and other import brands evolve as German households enter the
passenger-car market, treating the observed brand trajectory as the aggregate
of those individual decisions.  The goal is to recover both the monthly
market-share curve and the consumer decisions behind changes in that curve.
The hypothesis is directional: a consumer panel exposed to the monthly market
environment is expected to reproduce the level of every brand group's share and
the contrasting courses of the two entrant brands, the rise of one and the
decline and rebound of the other.

\paragraph{Framework Mapping.}
The case maps a consumer panel, a monthly market environment, and an
LLM-driven purchase decision onto the framework, so that brand shares emerge
from individual choices.
\begin{itemize}
  \item \textbf{Population \Pop.} A representative consumer panel whose
  demographic and query-specific attributes, market-entry times, purchase
  times, geographic locations, and social ties are generated from a
  structured run specification.
  \item \textbf{Environment \Env.} A monthly sequence of macro indicators,
  retrieved events, entity-level sentiment, and product-specific milestones.
  \item \textbf{Behavior function $f$.} An LLM that role-plays consumers in
  real mode and a deterministic utility model for offline runs. Both routes
  expose population priors, external signals, price fit, and peer influence
  as interpretable decision inputs.  The resulting behavior records whether,
  when, and from which entity each consumer purchases, together with belief
  checkpoints and social influence traces.
\end{itemize}

\paragraph{Experimental Design.}
\begin{itemize}
  \item \textbf{Ground truth and window.} Monthly shares of German private passenger-car registrations published by the KBA. The archived simulation panel spans January~2023--July~2026 and the registration panel ends in May~2026, leaving 41 overlapping months and 533 brand--month cells (\Cref{tab:marketsim-protocol}). KBA observations enter a post-hoc calibration of the simulated shares, so the comparison is a calibrated reconstruction.
  \item \textbf{Reference conditions.} Brand-level agreement is assessed against the registration data itself, cell by cell, across the 13 brand groups.
  \item \textbf{Longitudinal protocol.} A typed specification defines 13 brand groups, including a long-tail ``Other'' category, and four powertrain types (battery electric, plug-in hybrid, hybrid, and internal combustion). In total 600 representative agents enter the market on their own schedules, the monthly environment is collected, individual purchase choices are simulated, and brand and powertrain shares are aggregated and aligned with the registration shares. The case follows the build-from-scratch path (Path~B in \S\ref{sec:infra:pipeline}).
\end{itemize}
{Purchase decisions are typed actions of the behavior function, and the belief checkpoints and influence traces are panel rows, so a movement in a brand's share can be traced to the agents whose decisions produced it.}

\begin{figure}[!t]
  \centering
  \includegraphics[width=0.8\linewidth]{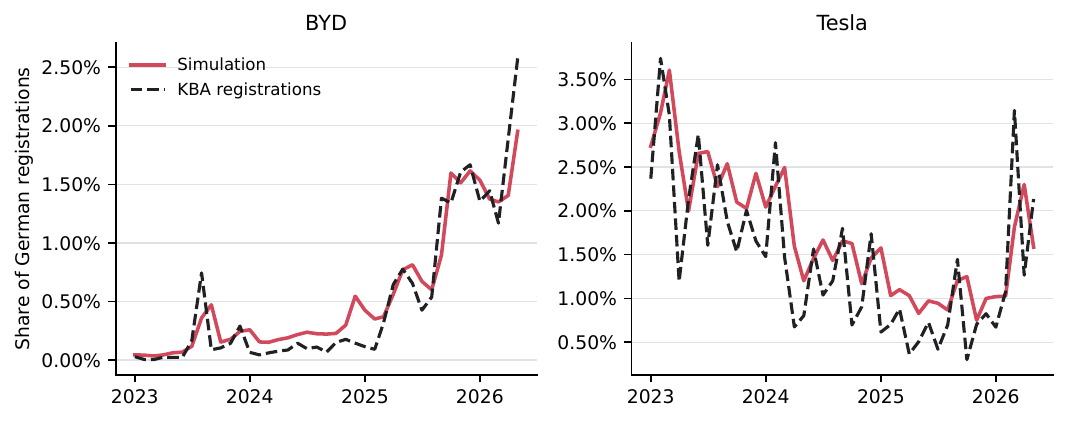}
  \caption{Monthly simulated and observed registration shares for BYD and
  Tesla in Germany over January~2023--May~2026 (41 overlapping months).  Solid
  lines show the calibrated simulation panel. Dashed lines show KBA registrations.}
  \label{fig:marketsim-germany}
\end{figure}

\paragraph{Results.}
\emph{Level fidelity: the simulated panel reproduces the brand-share structure of the German market
to within half a percentage point per cell.}  Across all 533 brand--month cells
in the overlap (\Cref{tab:marketsim-brand-comparison}), the mean absolute
discrepancy between simulated and KBA shares is 0.00442 in share, or 0.442
percentage points, with an RMSE of 0.618 percentage points. Mean simulated and
observed shares agree closely for every brand group, from VW (19.174\% versus
18.970\%) to BYD (0.549\% versus 0.508\%).

\emph{Brand-level accuracy varies with market position.}  In
\Cref{tab:marketsim-brand-comparison}, the two largest groups carry the
largest absolute discrepancies, the ``Other'' category (MAE 0.834 percentage
points) and VW (0.801 percentage points), while the entrant BYD has the
smallest MAE (0.157 percentage points) and the highest month-to-month
correlation ($r=0.951$).

\emph{Trajectory fidelity: the simulation captures the contrasting trajectories of the two entrant
brands.}
In \Cref{fig:marketsim-germany},  the calibrated panel closely tracks
BYD's rise and reproduces Tesla's broad decline and rebound with more
month-to-month smoothing ($r=0.761$, MAE $=0.508$ percentage points).

\emph{Boundary: these scores quantify reconstruction fidelity.}  Because KBA observations
enter the post-hoc calibration (pseudo-count 30, exponential smoothing with
$\alpha=0.45$, and a 0.03 per-cell residual cap), the comparison is a
calibrated backtest with reference blending. Out-of-sample forecasting
performance is a separate question that this evaluation does not address.

\paragraph{Takeaway.}
Household attributes and market events connect, through the population and environment components, to observable brand competition at the level of individual brand groups and their monthly courses.
Because the run preserves consumer trajectories and a social graph, a market-share movement can be traced back to particular constraints, signals, and peer exposures, which gives the platform a reusable micro-to-macro template for longitudinal consumer-market studies.

\section{Discussion}
\label{sec:discussion}

\subsection{Limitations}
\label{sec:discussion:limitations}

\paragraph{LLM fidelity and calibration.}
The behavioral realism of any LLM-driven simulation is bounded by the capabilities and biases of the underlying language model.
Our ABM benchmark (\S\ref{sec:cases:abm}) shows that LLM-$f$ reproduces emergent dynamics in the majority of tasks, but tasks requiring precise numerical computation (e.g., NaSch gap arithmetic) or fine-grained continuous control (e.g., Boids heading angles) exhibit measurably lower consistency.
More fundamentally, LLM agents inherit the training-data distribution of their backbone model: they may over-represent English-language, Western perspectives and under-represent populations absent from pre-training corpora.
We mitigate this through persona grounding, anchoring each agent to a demographically calibrated profile from the \userpool, but the gap between a prompted persona and a real person's decision process remains an open empirical question.

\paragraph{Computational cost and scale.}
Longitudinal simulation is inherently more expensive than cross-sectional prediction: a 15-step run over 19,235 agents issues $\sim$275K LLM calls even with archetype-level batching (\S\ref{sec:cases:chicago}).
The hybrid-$f$ paradigm (\S\ref{sec:cases:hisim}) offers one path to tractability, and falling inference costs will further ease the constraint, but today's frontier models impose practical limits on the population size and temporal depth that can be simulated within a research budget.

\paragraph{Validation methodology.}
Ground truth varies across case families: rule dynamics and observed opinion trajectories (\S\ref{sec:cases:mechanisms}), census trends and historical procurement outcomes (\S\ref{sec:cases:policy}), and official statistics and market data (\S\ref{sec:cases:macro}).
Each provides a different quality of reference signal, and none fully captures the counterfactual scenarios that are the primary application of longitudinal simulation.
Validating ``what if'' experiments, the most valuable use case, remains inherently difficult because the counterfactual did not happen.
Accordingly, paired counterfactual branches (\S\ref{sec:framework:loop}) are simulation-internal contrasts under the model's assumptions. Claims about real-world causal effects require external validation.

\paragraph{The human side of the loop.}
{The controllable research loop records each edit and its note, but not the reasoning that produced it, and the case studies report the version that was finally accepted rather than the tree of versions that led to it, so this report does not quantify how much the researcher's edits improved each result over an autonomous pass. Measuring that contribution, for instance by running the same study with and without researcher edits, is a natural next evaluation.}

\paragraph{Synthetic personas and population claims.}
The multi-source pool design deliberately bounds its own claims.
The ${\sim}$1B figure counts \emph{addressable} upstream persona records: PersonaHub personas are LLM-synthesized text, and MatrAIx is calibrated only on one-dimensional marginals and explicitly represents no real population, leaving joint distributions and source-selection bias uncorrected.
Representativeness in \sysname therefore derives from the IPF alignment step against declared target marginals. Persona texts generated from attributes are explicitly labeled as synthetic and are never treated as real behavioral data.
Synthetic personas approximate real populations, and results built on restricted or synthetic pools are labeled as such in generated reports.

\subsection{Ethical Considerations}
\label{sec:discussion:ethics}

\paragraph{Privacy and data governance.}
The \userpool aggregates real social-media profiles and open persona corpora across five sub-pools under tiered usage terms (\S\ref{sec:infra:governance}).
All data undergoes de-identification and is accessed only through typed MCP tools that enforce field-level access control. No raw personally identifiable information is exposed to simulation agents.
The population alignment step (\S\ref{sec:framework:overview}) projects aggregate demographic distributions onto synthetic archetypes via IPF, further insulating individual-level data.
Nevertheless, the risk of re-identification through rich persona combinations warrants ongoing review, particularly as the pool expands to new sources.

\paragraph{Dual-use and policy implications.}
A platform that simulates social dynamics with high fidelity could be misused, for example, to optimize misinformation campaigns or to justify discriminatory policies by presenting simulation outputs as ``evidence.''
We address this through three design choices.
(i)~Full auditability: every LLM call, its prompt, and its output are recorded in the trajectory store and can be inspected post-hoc.
(ii)~The agentic pipeline requires explicit human checkpoints at study design, parameter confirmation, and result interpretation, and every edit that changes a study is recorded with the version it produced.
(iii)~The platform is positioned as a research tool for hypothesis generation and exploration.
Simulation outputs are accompanied by provenance metadata (model version, temperature, prompt template) to discourage out-of-context citation.

\paragraph{Fairness and representation.}
Simulation populations should reflect the diversity of the real populations they model.
The \userpool's multi-source design and the IPF alignment procedure ensure demographic representativeness along known dimensions, but unmeasured dimensions (e.g. disability status, sexual orientation, religious affiliation) may remain under-represented depending on source-survey coverage.
Researchers using \sysname should report which population dimensions were aligned and which were not.

\subsection{Future Work}
\label{sec:discussion:future}

\paragraph{Adaptive and multi-model behavior functions.}
Current LLM-$f$ implementations use a single backbone model throughout a run.
Future work will explore adaptive scheduling, routing simple decisions to smaller models and complex ones to frontier models, and ensemble behavior functions that aggregate multiple models' outputs to improve robustness and reduce sensitivity to any single model's biases.
Complementing the population and environment services, we also plan a \emph{Behavior MCP}: a third service that provides interchangeable behavior backbones, exposing different underlying language models and agent frameworks (such as ReAct-style reasoning agents) behind the same behavior-function interface, so that a study can vary $f$ as systematically as it varies $P$ and $E$.

\paragraph{Tighter real-data grounding.}
The \eventtool currently provides point-in-time access to 21 data sources.
We plan to expand both temporal resolution (real-time feeds for financial and social-media data) and spatial granularity (sub-city geographies, building-level data), enabling studies that operate at finer scales than the current census-tract level.

\paragraph{Cross-study transfer and meta-simulation.}
The standardized \bfpe abstraction and panel storage format make it possible to transfer calibrated population segments and validated environment configurations across studies.
Longer-term, we envision meta-simulation capabilities: running multiple studies in parallel, comparing their trajectories, and identifying intervention strategies that remain robust across modeling assumptions.

\section{Conclusion}
\label{sec:conclusion}

This report has presented \sysname, {which carries the alignment-centered design of \svone~\citep{zhang2025socioverse} into a human-AI co-evolutionary paradigm built from two loops and one infrastructure,} formalized through a unified abstraction.
{The longitudinal simulation loop fixes a demographically aligned population $P$, grounds the environment $E$ in real-world data, and iterates environments and behaviors over time, producing per-agent panel trajectories in which a declared intervention forks a counterfactual branch. The controllable research loop treats the study itself as an editable state, so that an edit raised by the researcher or by an agent, on any component and at any stage, becomes a new version with a recorded lineage. The social science agentic infrastructure carries both loops through composable skills with researcher checkpoints, two data services, and an audit trail.}

We have validated the framework across three case families and seven case studies, whose ground truths range from rule dynamics and opinion trajectories to census trends, procurement outcomes, official statistics, and market data.
The breadth of these settings demonstrates that \sysname generalizes across domains.
{In each case, the reported result is a version that a human researcher and the agentic pipeline reached together, through declared interventions inside the run and edits to the study state between runs.}

Looking ahead, the combination of falling LLM inference costs, expanding real-data coverage, and the growing demand for processual social-science methods positions {the human-AI co-evolutionary paradigm, rather than either the handcrafted or the fully autonomous extreme,} as a practical research methodology.
\sysname's open architecture, including modular abstract interfaces, MCP-based data access, and an agentic skill pipeline, is designed to lower the barrier to entry for domain researchers and to grow with the ecosystem.
We release the framework, data connectors, benchmark suite, and case-study configurations as open-source resources to support reproducibility and community-driven extension.

\appendix
\section{Case Study Details and Additional Results}
\label{app:cases}

This appendix provides supplementary details for each case study that were omitted from the main text for space.

\subsection{Reporting Protocol}
\label{app:cases:protocol}

\Cref{tab:reporting-protocol} consolidates, for every case study, how stochasticity is handled, what is repeated, and how uncertainty is reported.
The cases fall into two regimes.
Synthetic-dynamics tasks (the ABM benchmark suite, the Chicago extension, and the procurement game) admit inexpensive replication, so they report variation across independent runs: seed-level standard deviations, $t$-based confidence intervals, or large fixed-policy evaluation batches.
Real-data longitudinal backtests (the three macro-index forecasting cases) evaluate a single calibrated configuration against official statistics over multi-year windows, because one run spans the full panel and replication at this scale exceeds a typical research budget; for these cases the tables report point estimates, uncertainty is conveyed through comparison against baseline methods on the same evaluation window, and the single-run status is disclosed here.
The hybrid opinion-dynamics case reports the mean of three runs per scenario from the underlying study.

\begin{table}[H]
  \centering
  \caption{Reporting protocol per case study.}
  \label{tab:reporting-protocol}
  \footnotesize
  \begin{tabular}{p{0.20\textwidth}p{0.34\textwidth}p{0.36\textwidth}}
    \toprule
    Case & Independent repetition & Uncertainty reporting \\
    \midrule
    Classic ABM suite & rule-$f$ control: 10 stochastic runs per task; each LLM: 3 runs & mean $\pm$ standard deviation; the controlled group (rule-$f$ scored against itself, 0.911) bounds attainable consistency \\
    \addlinespace
    Hybrid opinion dynamics & 3 runs per scenario, averaged (3 scenarios $\times$ 14 rounds; LLM temperature 0) & micro-level accuracy and F1 per scenario; macro trajectory correlation and bias against the observed series \\
    \addlinespace
    Chicago segregation & 5 perturbation seeds (42--46); rule baselines: 10 seeds; ablations on a 218-tract subset: 3 seeds (5 for race-blind) against matched controls & mean $\pm$ $t$-based 95\% confidence interval on endpoint indices \\
    \addlinespace
    Drug procurement & training seed 42 per drug--round task; final checkpoint evaluated over 1{,}000 episodes; random baseline averaged over 5 runs & distributions and medians across 325 drug--round tasks; fixed-seed replay yields the canonical trajectory \\
    \addlinespace
    CCI nowcast & single calibrated simulation per region (75-month backtest); inertia weights frozen after calibration & point estimates (MAE, RMSE, Pearson $r$) against official indices within a 13-method comparison (12 aggregate baselines; the shock-window ranking uses 14 methods); population-scale sensitivity reported below \\
    \addlinespace
    PMI nowcast & single frozen configuration; parameters fitted on 2015--2018, selected on 2019--2021 & point estimates on two held-out windows against persistence, consensus, and ensemble baselines (\Cref{tab:pmi-full-heldout}); mechanism ablations on the development window \\
    \addlinespace
    German car market & single calibrated backtest (600 agents); the offline route uses a deterministic utility model & reconstruction errors over 533 brand--month cells (\Cref{tab:marketsim-brand-comparison}); labeled as reconstruction, not held-out forecasting \\
    \bottomrule
  \end{tabular}
\end{table}

\subsection{ABM Benchmark Suite}

\paragraph{Task roster and hyperparameters.}
\Cref{tab:abm-suite} catalogues the 10 benchmark tasks by family, with their source models, population sizes, simulation lengths, and key configuration parameters.
All tasks use GPT-4o (temperature 0.7, max tokens 256) as the default LLM; ablations on DeepSeek-V3 and Qwen3-235B are reported in the main text.

\begin{table}[h]
  \centering\footnotesize
  \caption{The 10-task ABM benchmark suite: source models, population size $N$, simulation steps, and key configuration parameters. Each task is run under both rule-$f$ and LLM-$f$ on the same observation and action types.}
  \label{tab:abm-suite}
  \begin{tabular}{lllrrl}
    \toprule
    Family & Task & Source ABM & $N$ & Steps & Key parameter \\
    \midrule
    \multirow{3}{*}{Flow}
      & NaSch        & \citet{nagel1992cellular}     & 200 & 50  & $v_{\max}=5$, $p_{\text{brake}}=0.3$ \\
      & Boids        & \citet{reynolds1987flocks}    & 40  & 100 & radius=5, heading tol.\ $30^\circ$ \\
      & Social Force & \citet{helbing2000simulating} & 80  & 200 & desired speed 1.2\,m/s \\
    \addlinespace
    \multirow{3}{*}{Market}
      & Sugarscape     & \citet{epstein1996growing}   & 200 & 100 & metabolism $\in [1,4]$, vision $\in [1,6]$ \\
      & Minority Game  & \citet{challet1997emergence} & 301 & 100 & memory $m=3$, strategies $s=2$ \\
      & Axelrod IPD    & \citet{axelrod1984evolution} & 64  & 200 & 5 strategies, round-robin \\
    \addlinespace
    \multirow{2}{*}{Organization}
      & Schelling       & \citet{schelling1971dynamic} & 810 & 30 & threshold 0.3, grid $30 \times 30$ \\
      & Civil Violence  & \citet{epstein2002modeling}  & 230 & 50 & legitimacy 0.82, vision 7 \\
    \addlinespace
    \multirow{2}{*}{Diffusion}
      & SIR rumor          & \citet{kermack1927contribution} & 500 & 30 & $\beta=0.3$, $\gamma=0.1$ \\
      & Hegselmann--Krause & \citet{hegselmann2002opinion}   & 100 & 50 & confidence bound 0.2 \\
    \bottomrule
  \end{tabular}
\end{table}

\paragraph{Consistency-score metric vectors.}
Per-task consistency scores for all four conditions (controlled group, GPT-4o, DeepSeek-V3, Qwen3-235B) are reported in \Cref{fig:abm-consistency}.
Each task's score is computed over a small vector of task-level outcome metrics: the rule-based ABM's ten-run mean serves as the reference value for every metric, each evaluated run receives one score against that reference, and scores are averaged per condition.
\Cref{tab:abm-metric-vectors} catalogues the metric vector of every task~\citep{zhang2025abm}.

\begin{table}[H]
  \centering
  \caption{Per-task metric vectors entering the consistency score.}
  \label{tab:abm-metric-vectors}
  \footnotesize
  \begin{tabular}{lp{3.2cm}p{7.6cm}}
    \toprule
    Task & Metric vector & Definitions \\
    \midrule
    NaSch & mean speed; flow rate; stopped fraction & mean vehicle speed (cells/step); throughput $q=\bar{v}\rho$ at density $\rho=0.25$; share of vehicles at speed 0 \\
    Social Force & mean speed; density; flow rate & mean speed of active pedestrians; active pedestrians per unit area; their product \\
    Boids & polarization; NND; spread; mean speed & alignment order $\Phi=\lvert\langle\hat{v}\rangle\rvert$; mean nearest-neighbour distance; mean distance to flock centroid; mean speed \\
    Sugarscape & mean wealth; survivors; Gini & mean sugar of survivors; number of survivors; Gini coefficient of survivor wealth \\
    Axelrod IPD & 15 per-strategy scores & mean score per turn of each Axelrod-1980 tournament strategy, in fixed order \\
    Minority Game & mean attendance; volatility; variance & mean count choosing side A (${\approx}N/2$); its standard deviation $\sigma$; $\sigma^2$ \\
    Schelling & initial happy; final happy & satisfied residents in the initial random configuration and at convergence \\
    Civil Violence & final active; final jailed; peak active & fractions rebelling and jailed at the last step; largest simultaneous rebellion \\
    SIR rumor & final reach; $N$ & agents ever informed; effective population size (largest connected component) \\
    Hegselmann--Krause & clusters at $\varepsilon\in\{0.01,0.15,0.25\}$ & final opinion-cluster count at three confidence bounds \\
    \bottomrule
  \end{tabular}
\end{table}

\begin{figure}[H]
  \centering
  \begin{subfigure}[b]{0.42\linewidth}
    \includegraphics[width=\linewidth]{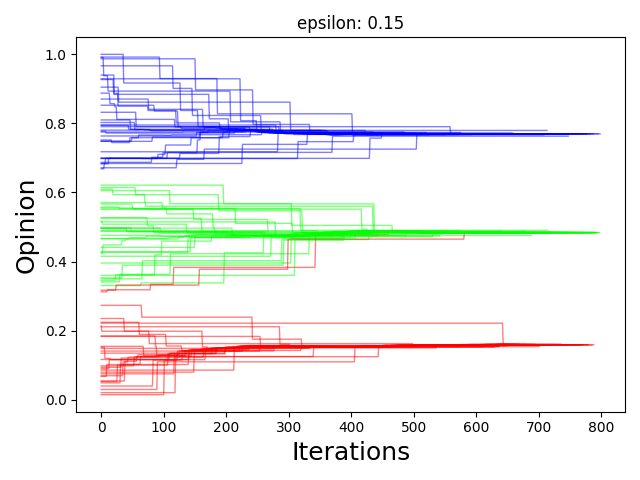}
    \caption{HK, rule-$f$}
  \end{subfigure}\hfill
  \begin{subfigure}[b]{0.42\linewidth}
    \includegraphics[width=\linewidth]{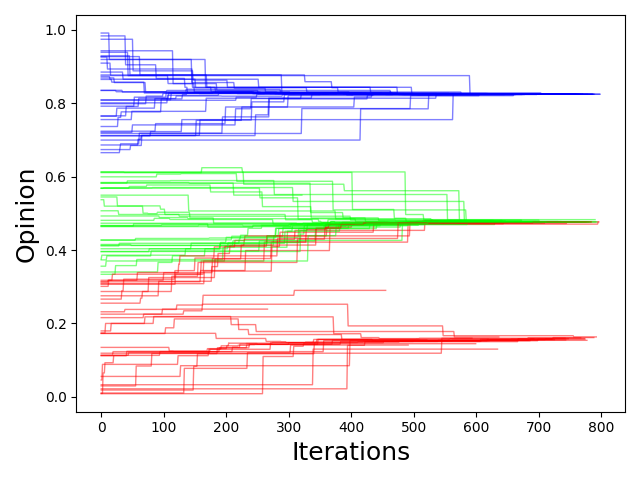}
    \caption{HK, LLM-$f$}
  \end{subfigure}\\[4pt]
  \begin{subfigure}[b]{0.42\linewidth}
    \includegraphics[width=\linewidth]{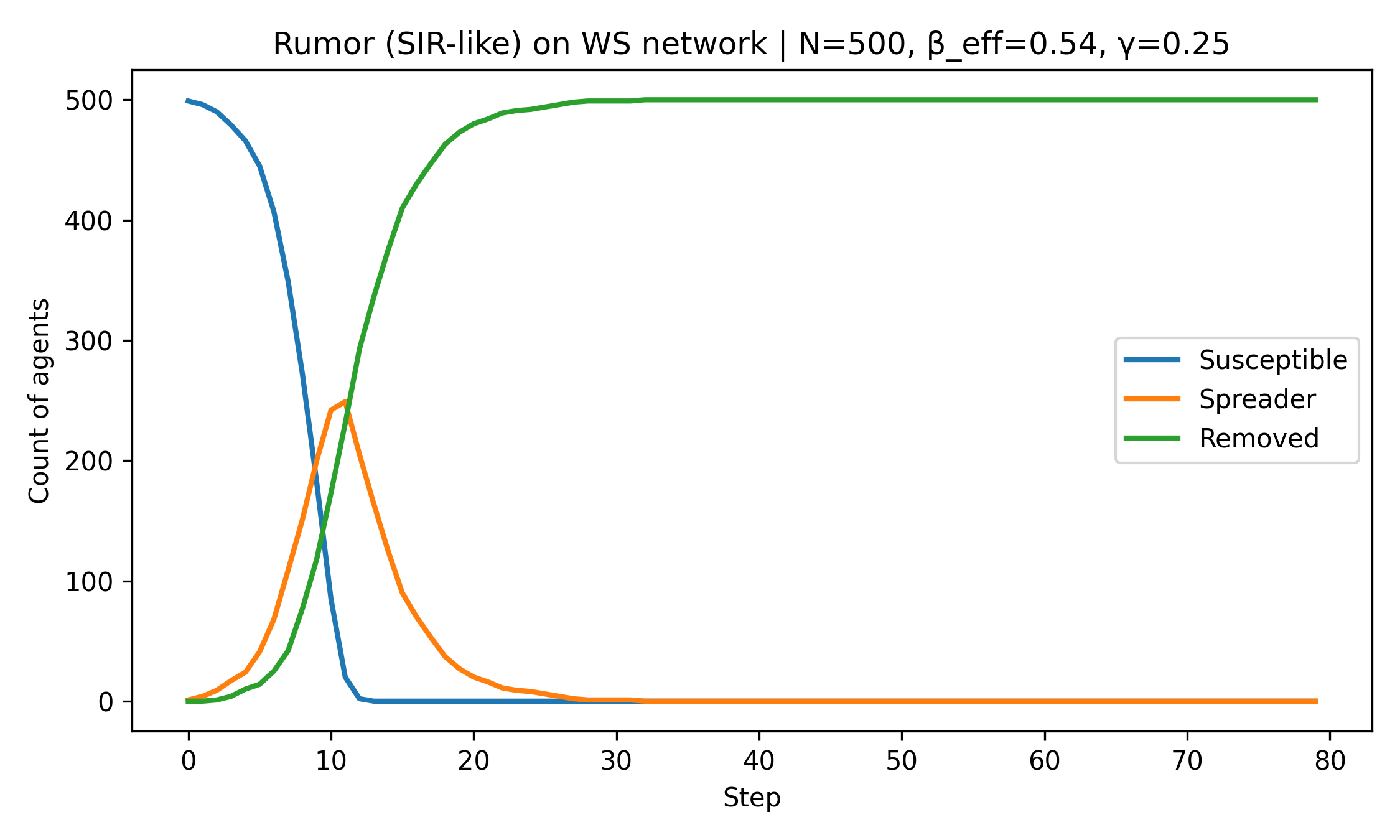}
    \caption{SIR rumor, rule-$f$}
  \end{subfigure}\hfill
  \begin{subfigure}[b]{0.42\linewidth}
    \includegraphics[width=\linewidth]{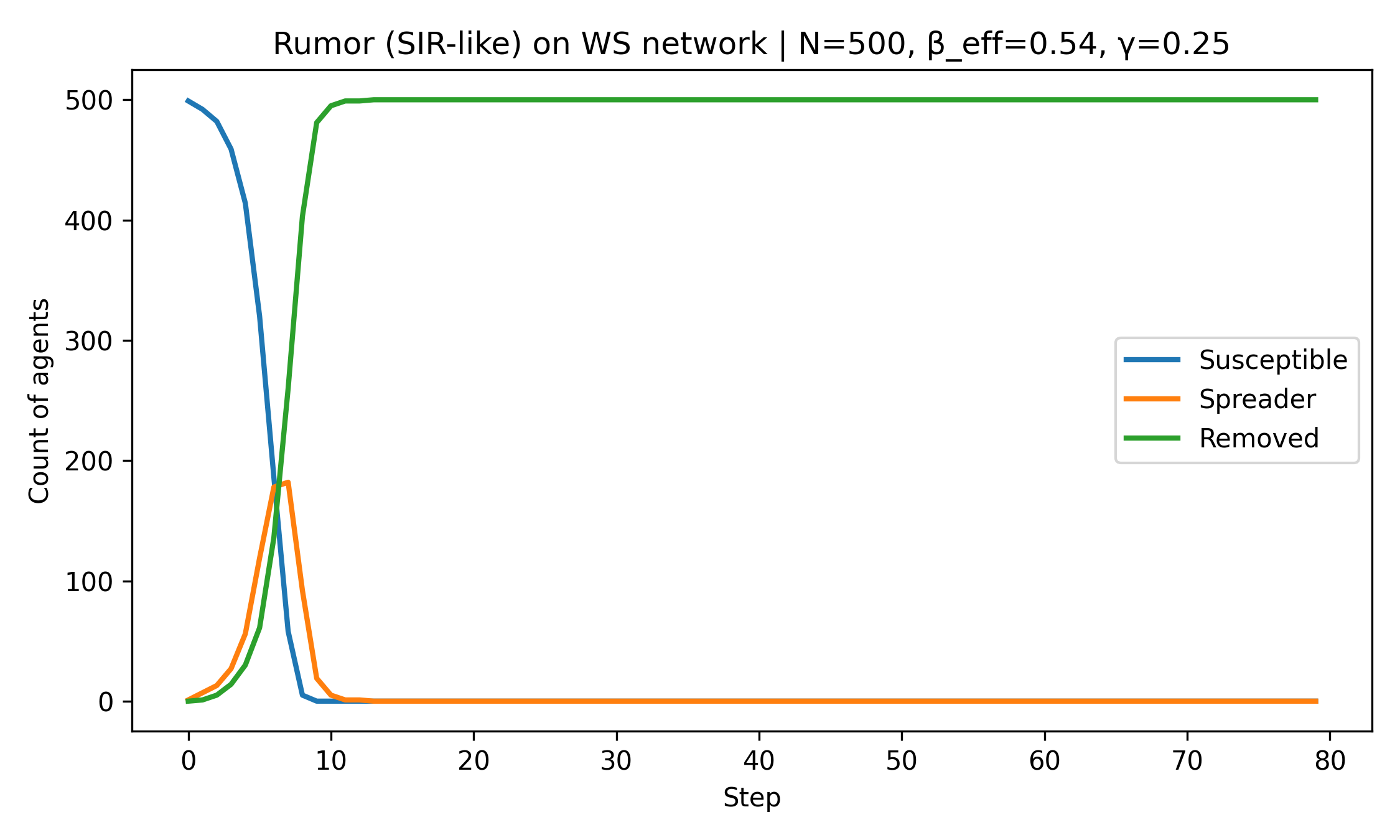}
    \caption{SIR rumor, LLM-$f$}
  \end{subfigure}
  \caption{Qualitative dynamics under rule-$f$ (left) vs.\ LLM-$f$ (right). The top row shows Hegselmann--Krause opinion trajectories ($\varepsilon=0.15$) and the bottom row SIR rumor spreading on a Watts--Strogatz network. Figures adapted from~\citet{zhang2025abm}.}
  \label{fig:abm-dynamics-compare}
\end{figure}

\paragraph{Qualitative dynamics comparison.}
\Cref{fig:abm-dynamics-compare} shows two representative side-by-side comparisons of rule-$f$ and LLM-$f$ group dynamics: Hegselmann--Krause opinion trajectories at confidence bound $\varepsilon=0.15$ (both conditions converge to three opinion clusters) and the SIR rumor curve on a Watts--Strogatz contact network (both reproduce the characteristic spreader peak and saturation of the informed population).

\subsection{Hybrid Opinion Dynamics}

\paragraph{Scenario configuration.}
Three event-driven scenarios are evaluated:
\begin{itemize}
  \item \textbf{Roe v.\ Wade} (abortion rights): triggered by the June 2022 Supreme Court decision; core users include activists, journalists, and political commentators with documented stance histories.
  \item \textbf{MeToo} (sexual harassment): calibrated on the October 2017 viral-hashtag window and validated on a later campaign window; core users include public figures who engaged prominently in the discourse.
  \item \textbf{Black Lives Matter} (racial justice): triggered by the May 2020 George Floyd incident; core users include community organizers and media personalities.
\end{itemize}

\paragraph{ABM model selection.}
Five interchangeable opinion dynamics models govern ordinary-user attitude updates:
bounded confidence (Deffuant--Weisbuch), Hegselmann--Krause, Lorenz, relative agreement, and social judgement.
The best-performing model is selected per scenario based on macro-level trajectory fit.

\paragraph{Simulation configuration.}
Core users are driven by LLM agents (temperature 0, max tokens 256) on a customized AgentVerse stack; ordinary users run on Mesa implementations of the five opinion-dynamics ABMs.
Each event simulates 1{,}000 users (300 core users selected by influence and activity ranking, plus 700 annotated ordinary users) for 14 rounds.
Per-scenario micro-level scores (stance accuracy/F1, content similarity, behavior-type accuracy/F1) are reported in \Cref{tab:hisim-micro} of the main text.

\paragraph{Cost and scalability.}
The hybrid design pins LLM API cost to the number of \emph{core} users: simulation runtime is dominated by the 300 core users' LLM calls, and scaling the ordinary-user population adds only negligible ABM computation, whereas a pure-LLM simulation scales API calls linearly with the full population~\citep{mou2024unveiling}.
In the scalability analysis of the original paper, growing the ordinary-user count around a fixed 300-user core leaves all system metrics except attitude bias nearly unchanged while runtime stays essentially flat, supporting sampling-based hybrid simulation as the practical configuration for larger populations.

\subsection{Chicago Segregation}

\paragraph{Archetype construction.}
The 241 household archetypes are constructed from four dimensions of the 2010 Census:
race (5 categories: Non-Hispanic White, Non-Hispanic Black, Hispanic, Non-Hispanic Asian, Other),
income bracket (5 levels: $<$\$25K, \$25--50K, \$50--75K, \$75--100K, $>$\$100K),
family type (4 categories: married with children, married no children, single parent, single adult),
and neighborhood type (urban core, suburban, transitional).
Not all combinations are populated; the final 241 archetypes cover the observed co-occurrence patterns in Chicago's census tracts.

Each archetype's behavioral parameters (mobility rate, ideal own-group percentage, satisfaction factor weights) are calibrated from nine published sociological studies on residential mobility and neighborhood preferences.
The calibration procedure assigns importance weights to seven satisfaction factors (racial composition, housing cost, school quality, safety, transit access, social anchor, job proximity) based on documented group-level preferences.

\paragraph{Tract mapping.}
The 781 census tracts are connected via Queen contiguity (shared edge or vertex).
Each tract carries dynamic attributes (racial composition, safety score, school quality index, transit access score) that update endogenously as households move.
Initial tract attributes are drawn from the American Community Survey 2006--2010 5-year estimates.

\paragraph{Rule-based reference suite.}
The baseline comparison in the main text runs five rule-based references on the same tract graph, under the same per-step move budget, capacity limits, inflow caps, and execution machinery as the LLM condition, each for ten perturbation seeds~\citep{zhang2025abm}:
(i)~\emph{random reassignment}, distributing households uniformly across tracts (the lower bound on $D_{BW}$ under spatial independence);
(ii)~\emph{classical Schelling} with a single similarity threshold $\tau=0.5$ on the Queen-contiguity graph, relocating to a random feasible neighboring tract;
(iii)~the same rule with best-improvement destination choice (feasible neighbor with the highest own-group share);
(iv)~a \emph{per-group threshold} rule calibrated from the same literature-derived archetype parameters the LLM receives, moving when the own-group share falls below the archetype's calibrated minimum;
and (v)~a \emph{logit-style utility} rule scoring candidate tracts by a weighted sum of racial fit, safety, school access, transit, amenities, and housing-cost match using each archetype's satisfaction weights.
Beyond the endpoint indices (\Cref{tab:chicago-endpoints}), the conditions differ in their mobility signature: the LLM-driven population's per-step movers decline toward rest (from roughly 1{,}050 to 250) as the target pattern is approached, whereas every rule-based condition keeps relocating at the budget cap through step~15, and the per-group threshold rule is still rising at the end of the horizon~\citep{zhang2025abm}.

\subsection{Drug Procurement}

For each drug-specific task, the RL pipeline trains a role-structured policy set comprising one government policy, one medical-institution policy, and one enterprise policy for each participating firm; parameters are not shared across enterprise actors. The government and medical-institution agents are optimised with PPO, whereas enterprise agents use a centralised-critic PPO formulation under centralised training and decentralised execution, denoted MAPPO. All actors condition solely on role-specific local observations. The government and enterprise critics additionally condition on the concatenated shared state, while the medical-institution critic remains local. The actor and critic each contain two \texttt{Tanh}-activated hidden layers, with widths of 128 and 256, respectively, and layer normalisation applied to their inputs. In the reference configuration, each task is trained with seed 42 for 10,000 episodes of 50 steps. Learning rates are initialised at $5\times10^{-5}$ and decay linearly to $5\times10^{-6}$ for the government and enterprise policies and to $7.5\times10^{-6}$ for the medical-institution policy. PPO uses $\gamma=0.99$, a generalised advantage estimation parameter of $\lambda=0.95$, a clipping ratio of 0.2, a value-loss coefficient of 0.5, a value-function clipping threshold of 50, a maximum gradient norm of 0.5, and a target Kullback--Leibler divergence of 0.01. Each update aggregates 10 complete episodes, corresponding to 500 environment transitions, and performs three optimisation epochs with five-episode minibatches of 250 transitions. The entropy coefficient is annealed from $3\times10^{-3}$ to $2\times10^{-4}$ over the first 60\% of training for the government and enterprise policies, and from $5\times10^{-4}$ to $5\times10^{-5}$ over the first 10\% for the medical-institution policy.

The three agent classes optimise distinct objectives: government social welfare, firm-specific enterprise profit, and medical-institution utility. Checkpoints are saved and evaluated every 2,000 training episodes, using 100 episodes for each interim evaluation and 1,000 episodes for the final evaluation. Candidate checkpoints are first restricted to the Pareto-nondominated set defined by mean government social welfare, mean aggregate enterprise profit, and mean medical-institution utility. The final checkpoint is then selected using a balanced lexicographic criterion that prioritises the lowest normalised stakeholder objective, followed by the mean and sum across the three objectives, thereby avoiding their reduction to a fixed weighted scalarisation.

\subsection{Nowcasting the Consumer Confidence Index}

\paragraph{Regional configurations.}
The consumer-confidence case uses one response architecture across the U.S., EU27, and Japan,
while adapting the official survey battery, signal field, and persistence
parameter to each region (\Cref{tab:consumersim-regional-config}).  The U.S.
population is constructed from SIPP and GSS evidence; the EU27 and Japan runs
are cross-context adaptations rather than full local-microdata
rebuilds~\citep{huang2026consumersim}.

\begin{table}[H]
  \centering
  \caption{Region-specific configuration of the consumer-confidence case.  The inertia weights are
  selected by validation RMSE on January--June~2024 and then frozen.}
  \label{tab:consumersim-regional-config}
  \small
  \resizebox{\textwidth}{!}{%
  \begin{tabular}{lllll}
    \toprule
    Region & Population substrate & Official target and scoring & Environment adaptation & Inertia $\lambda$ \\
    \midrule
    U.S. & \makecell[l]{SIPP-calibrated households;\\GSS-grounded behavioral attributes}
      & \makecell[l]{Michigan five-item battery;\\published ICC/ICE normalization}
      & \makecell[l]{U.S. macro, labor, housing,\\financial, policy, and news signals}
      & 0.4 \\
    \addlinespace[0.35em]
    EU27 & \makecell[l]{Shared response architecture;\\no full local microdata rebuild}
      & \makecell[l]{EC/Eurostat balances: finances, economy,\\unemployment (reversed), and savings}
      & \makecell[l]{EU27 target scale and\\region-matched macro/market field}
      & 0.6 \\
    \addlinespace[0.35em]
    Japan & \makecell[l]{Shared response architecture;\\no full local microdata rebuild}
      & \makecell[l]{Cabinet Office/ESRI: livelihood, income,\\employment, and durable buying}
      & \makecell[l]{Japanese target scale and\\region-matched macro/market field}
      & 0.7 \\
    \bottomrule
  \end{tabular}}
\end{table}

The three official indices use different scales; levels are therefore compared
within, not across, regions.

\paragraph{Month-level audit.}
\Cref{tab:consumersim-monthly-audit} reports the rolling public-interface
snapshot available on July~13, 2026.\footnote{Public interface data of the consumer-confidence subsystem:
\url{https://sii-research.github.io/ConsumerSim/data/consumersim_site_data.csv}
(snapshot retrieved August~15, 2026).}
April--June~2026 are later operational predictions and are excluded from the
main-text evaluation.

\begin{table}[H]
  \centering
  \caption{Recent monthly simulated predictions and official CCI values for
  all three regions.  ``Pred.'' denotes the value shown on the public
  interface of the consumer-confidence case.}
  \label{tab:consumersim-monthly-audit}
  \footnotesize
  \setlength{\tabcolsep}{5pt}
  \begin{tabular}{lrrrrrr}
    \toprule
    & \multicolumn{2}{c}{U.S.} & \multicolumn{2}{c}{EU27} & \multicolumn{2}{c}{Japan} \\
    \cmidrule(lr){2-3}\cmidrule(lr){4-5}\cmidrule(lr){6-7}
    Month & Pred. & Official & Pred. & Official & Pred. & Official \\
    \midrule
    2025-08 & 62.12 & 61.7 & $-13.42$ & $-13.2$ & 34.76 & 34.0 \\
    2025-09 & 63.01 & 58.2 & $-13.23$ & $-13.4$ & 34.07 & 34.9 \\
    2025-10 & 59.61 & 55.1 & $-13.39$ & $-13.5$ & 34.89 & 35.4 \\
    2025-11 & 56.30 & 53.6 & $-12.96$ & $-12.5$ & 35.31 & 35.9 \\
    2025-12 & 55.32 & 54.0 & $-12.11$ & $-12.3$ & 35.90 & 37.2 \\
    2026-01 & 55.91 & 54.9 & $-12.40$ & $-12.6$ & 37.22 & 36.9 \\
    2026-02 & 57.26 & 56.4 & $-12.54$ & $-12.8$ & 37.00 & 37.4 \\
    2026-03 & 58.65 & 57.6 & $-11.81$ & $-11.9$ & 37.69 & 37.7 \\
    2026-04 & 57.47 & 56.4 & $-11.97$ & $-12.4$ & 39.65 & 39.3 \\
    2026-05 & 53.00 & 53.2 & $-13.60$ & $-13.9$ & 35.10 & 36.0 \\
    2026-06 & 52.70 & 49.5 & $-15.10$ & $-17.0$ & 34.80 & 33.8 \\
    \bottomrule
  \end{tabular}
\end{table}

\paragraph{Population-size sensitivity.}
With a fixed 5,000-agent core, expansion from $5\times$ to $50\times$ reduces
MAE from 6.10 to 1.72 and raises Pearson $r$ from 0.682 to 0.979.  The
$100\times$ run changes these values only to 1.70 and 0.980, supporting the
$50\times$ configuration as the accuracy--scale plateau.  The sweep is reported by the underlying study on a recent U.S.\ evaluation window distinct from the 75-month backtest.

\subsection{Nowcasting the Purchasing Managers' Index}

\paragraph{Temporal protocol.}
Inputs are frozen at the 21st-day cutoff and evaluated against first-print ISM
Manufacturing PMI.  Parameters are calibrated on 2015--2018 and selected on
2019--2021; evaluation uses a clean test (2022-01--2023-10) and a post-cutoff
test (2023-11--2026-06), each preceded by a three-month memory warm-up.

\paragraph{Tuned and frozen quantities.}
The change scale $c=0.2272$, update step $\kappa=0.2$, and Bates--Granger weight
$w=0.1675$ are fitted before held-out evaluation and then frozen.  The protocol
uses 300 firms, one vote per firm, five equally weighted PMI components, and an
October~2023 response-model knowledge cutoff.

\paragraph{Complete held-out comparison.}
\Cref{tab:pmi-full-heldout} compares the frozen model with consensus,
persistence, the predecessor response kernel, and a mechanical-threshold twin.

\begin{table}[H]
  \centering
  \caption{Complete PMI comparison on the two held-out windows.  Direction is
  the accuracy of $\operatorname{sign}(\widehat{Y}_t-Y_{t-1})$. Persistence
  makes no directional call.}
  \label{tab:pmi-full-heldout}
  \footnotesize
  \setlength{\tabcolsep}{4pt}
  \begin{tabular}{llrrrr}
    \toprule
    Window & Method & MAE & RMSE & Direction & Side of 50 \\
    \midrule
    \multirow{7}{*}{\makecell[l]{Clean test\\2022-01--2023-10}}
      & Agent-based change nowcast & 1.025 & 1.228 & 65.0\% & 95.5\% \\
      & Bates--Granger ensemble & \textbf{0.928} & \textbf{1.151} & \textbf{80.0\%} & \textbf{100.0\%} \\
      & Earlier-kernel change nowcast & 1.044 & 1.246 & 70.0\% & 95.5\% \\
      & Earlier-kernel ensemble & 0.933 & 1.157 & \textbf{80.0\%} & \textbf{100.0\%} \\
      & Mechanical-threshold twin & 1.041 & 1.243 & 70.0\% & 95.5\% \\
      & Market consensus & 0.936 & 1.160 & 75.0\% & 95.5\% \\
      & Persistence & 1.073 & 1.293 & --- & 95.5\% \\
    \addlinespace
    \multirow{7}{*}{\makecell[l]{Post-cutoff\\2023-11--2026-06}}
      & Agent-based change nowcast & \textbf{0.892} & 1.256 & 58.6\% & 87.5\% \\
      & Bates--Granger ensemble & 0.898 & 1.188 & 51.7\% & \textbf{90.6\%} \\
      & Earlier-kernel change nowcast & 0.942 & 1.258 & \textbf{65.5\%} & 87.5\% \\
      & Earlier-kernel ensemble & 0.900 & \textbf{1.184} & 55.2\% & 87.5\% \\
      & Mechanical-threshold twin & 0.940 & 1.257 & \textbf{65.5\%} & 87.5\% \\
      & Market consensus & 0.922 & 1.202 & 51.7\% & \textbf{90.6\%} \\
      & Persistence & 0.894 & 1.279 & --- & 84.4\% \\
    \bottomrule
  \end{tabular}
\end{table}

\paragraph{Ablation matrix.}
Mechanism ablations use the 2019--2021 development window and a fixed 165-firm
subpanel.  Downstream coefficients remain frozen; inertia applies only to the
level path.

\begin{table}[H]
  \centering
  \caption{Complete PMI mechanism-ablation matrix.  $\Delta$MAE is relative to
  the full model under the same output path. Positive values indicate higher
  error after removal.}
  \label{tab:pmi-ablation-full}
  \small
  \begin{tabular}{llrrr}
    \toprule
    Variant & Output path & MAE & Direction & $\Delta$MAE \\
    \midrule
    Full model & Level & 4.868 & 0.400 & 0.000 \\
    Full model & Change & 1.948 & 0.441 & 0.000 \\
    Without post-stratification & Level & 4.817 & 0.400 & $-0.051$ \\
    Without post-stratification & Change & 1.917 & 0.471 & $-0.031$ \\
    Without inertia alignment & Level & 5.905 & 0.429 & 1.037 \\
    Deterministic response kernel & Level & 5.044 & 0.600 & 0.176 \\
    Deterministic response kernel & Change & 1.900 & 0.514 & $-0.048$ \\
    Without news field & Level & 4.877 & 0.400 & 0.009 \\
    Without news field & Change & 1.953 & 0.441 & 0.005 \\
    Without manager memorandum & Level & 4.904 & 0.400 & 0.036 \\
    Without manager memorandum & Change & 1.932 & 0.471 & $-0.016$ \\
    Homogeneous firm persona & Level & 4.697 & 0.429 & $-0.171$ \\
    Homogeneous firm persona & Change & 1.957 & 0.441 & 0.009 \\
    Without longitudinal memory & Level & 5.121 & 0.400 & 0.253 \\
    Without longitudinal memory & Change & 1.942 & 0.471 & $-0.006$ \\
    \bottomrule
  \end{tabular}
\end{table}

Removing inertia causes the largest level-path degradation ($+1.037$ MAE);
change-path differences are small and should be interpreted as diagnostics on
this 36-month reduced panel.

\subsection{German Passenger-Car Market}

\paragraph{Evaluation protocol.}
The German passenger-car case is evaluated as a monthly market-share
reconstruction.  Simulated and KBA observations are aligned at the brand and
powertrain levels according to \Cref{tab:marketsim-protocol}.

\begin{table}[H]
  \centering
  \caption{Evaluation protocol for the German passenger-car case.}
  \label{tab:marketsim-protocol}
  \small
  \begin{tabular}{p{0.25\textwidth}p{0.67\textwidth}}
    \toprule
    Protocol element & Specification \\
    \midrule
    Market and outcome
      & Monthly shares of German private passenger-car registrations; KBA registrations provide the empirical reference. \\
    Units of analysis
      & Brand--month cells for the primary comparison and powertrain--month cells for supplementary reconstruction. \\
    Market taxonomy
      & 13 mutually exclusive brand groups and four powertrains (BEV, PHEV, HEV, and internal combustion). \\
    Time coverage
      & Simulation: January~2023--July~2026; KBA reference: January~2023--May~2026. \\
    Matched sample
      & 41 overlapping months $\times$ 13 brand groups $=533$ brand--month cells. \\
    Comparison rule
      & Simulated and observed shares are aligned by calendar month and market category; errors are reported in percentage points. \\
    \bottomrule
  \end{tabular}
\end{table}

\paragraph{Market configuration.}
The calibrated backtest uses 600 representative agents.  Counts are normalized
to approximately 80,000 monthly units before conversion to market shares; the
two simulated months without KBA observations are excluded from evaluation.

\paragraph{Calibration diagnostics.}
Post-hoc calibration uses a pseudo-count of 30, exponential smoothing
($\alpha=0.45$), and a 0.03 per-cell residual cap.  Across matched brand cells,
MAE is 0.442 percentage points and RMSE is 0.618 percentage points.  Because
KBA observations enter calibration, these are reconstruction rather than
held-out forecasting scores.

\paragraph{Brand-level comparison.}
\Cref{tab:marketsim-brand-comparison} summarizes market level, absolute error,
and month-to-month co-movement.  BYD has the smallest MAE and highest
correlation; the broad \emph{Other} group and VW have the largest discrepancies.

\begin{table}[H]
  \centering
  \caption{Brand-level simulated versus KBA registration shares,
  January~2023--May~2026 ($n=41$ months per brand).}
  \label{tab:marketsim-brand-comparison}
  \small
  \begin{tabular}{lrrrr}
    \toprule
    Brand group & Mean KBA share & Mean simulated share & MAE (pp) & Pearson $r$ \\
    \midrule
    VW          & 18.970\% & 19.174\% & 0.801 & 0.558 \\
    BMW         &  8.442\% &  8.623\% & 0.539 & 0.633 \\
    Mercedes    &  9.299\% &  9.162\% & 0.554 & 0.709 \\
    Audi        &  7.646\% &  7.687\% & 0.414 & 0.796 \\
    Skoda       &  7.274\% &  7.020\% & 0.484 & 0.859 \\
    Seat/Cupra  &  5.232\% &  5.125\% & 0.309 & 0.799 \\
    Opel        &  5.003\% &  4.940\% & 0.391 & 0.686 \\
    Toyota      &  2.935\% &  2.942\% & 0.266 & 0.770 \\
    Hyundai     &  3.460\% &  3.319\% & 0.239 & 0.666 \\
    Renault     &  2.118\% &  2.189\% & 0.253 & 0.546 \\
    Tesla       &  1.444\% &  1.736\% & 0.508 & 0.761 \\
    BYD         &  0.508\% &  0.549\% & \textbf{0.157} & \textbf{0.951} \\
    Other       & 27.400\% & 27.535\% & 0.834 & 0.704 \\
    \midrule
    All 533 cells & --- & --- & 0.442 & --- \\
    \bottomrule
  \end{tabular}
\end{table}


\begin{thebibliography}{54}
\providecommand{\natexlab}[1]{#1}
\providecommand{\url}[1]{\texttt{#1}}
\expandafter\ifx\csname urlstyle\endcsname\relax
  \providecommand{\doi}[1]{doi: #1}\else
  \providecommand{\doi}{doi: \begingroup \urlstyle{rm}\Url}\fi

\bibitem[Schelling(1971)]{schelling1971dynamic}
Thomas~C. Schelling.
\newblock Dynamic models of segregation.
\newblock \emph{Journal of Mathematical Sociology}, 1\penalty0 (2):\penalty0
  143--186, 1971.

\bibitem[Epstein and Axtell(1996)]{epstein1996growing}
Joshua~M. Epstein and Robert Axtell.
\newblock \emph{Growing Artificial Societies: Social Science from the Bottom
  Up}.
\newblock Brookings Institution Press \& MIT Press, 1996.

\bibitem[Park et~al.(2023)Park, O'Brien, Cai, Morris, Liang, and
  Bernstein]{park2023generative}
Joon~Sung Park, Joseph O'Brien, Carrie~Jun Cai, Meredith~Ringel Morris, Percy
  Liang, and Michael~S. Bernstein.
\newblock Generative agents: Interactive simulacra of human behavior.
\newblock In \emph{Proceedings of the 36th Annual ACM Symposium on User
  Interface Software and Technology (UIST)}, 2023.
\newblock \doi{10.1145/3586183.3606763}.

\bibitem[Argyle et~al.(2023)Argyle, Busby, Fulda, Gubler, Rytting, and
  Wingate]{argyle2023out}
Lisa~P. Argyle, Ethan~C. Busby, Nancy Fulda, Joshua~R. Gubler, Christopher
  Rytting, and David Wingate.
\newblock Out of one, many: Using language models to simulate human samples.
\newblock \emph{Political Analysis}, 31\penalty0 (3):\penalty0 337--351, 2023.

\bibitem[Gao et~al.(2023)Gao, Lan, Lu, Mao, Piao, Wang, Jin, and Li]{gao2023s3}
Chen Gao, Xiaochong Lan, Zhihong Lu, Jinzhu Mao, Jinghua Piao, Huandong Wang,
  Depeng Jin, and Yong Li.
\newblock {S$^3$}: Social-network simulation system with large language
  model-empowered agents.
\newblock \emph{arXiv preprint arXiv:2307.14984}, 2023.

\bibitem[Vezhnevets et~al.(2023)Vezhnevets, Agapiou, Aharon, Ziv, Matyas,
  Du{\'e}{\~n}ez-Guzm{\'a}n, Cunningham, Osindero, Karmon, and
  Leibo]{vezhnevets2023concordia}
Alexander~Sasha Vezhnevets, John~P. Agapiou, Avia Aharon, Ron Ziv, Jayd Matyas,
  Edgar~A. Du{\'e}{\~n}ez-Guzm{\'a}n, William~A. Cunningham, Simon Osindero,
  Danny Karmon, and Joel~Z. Leibo.
\newblock Generative agent-based modeling with actions grounded in physical,
  social, or digital space using {Concordia}.
\newblock \emph{arXiv preprint arXiv:2312.03664}, 2023.

\bibitem[Wang et~al.(2025)Wang, Gao, Bo, Chen, and Wen]{wang2025yulan}
Lei Wang, Heyang Gao, Xiaohe Bo, Xu~Chen, and Ji-Rong Wen.
\newblock Yulan-onesim: Towards the next generation of social simulator with
  large language models.
\newblock \emph{arXiv preprint arXiv:2505.07581}, 2025.

\bibitem[Piao et~al.(2026)Piao, Zhang, Huang, Zhang, Wang, Zhao, Li, Sun,
  Chang, Xu, Wang, Zhang, Rong, Su, Meng, Liu, Meng, Wang, and
  Li]{piao2026agentsociety2}
Jinghua Piao, Jun Zhang, Haoyu Huang, Keming Zhang, Jing~Yi Wang, Xinran Zhao,
  Songwei Li, Boyuan Sun, Jiayi Chang, Fengli Xu, Chunyan Wang, Fang Zhang,
  Ke~Rong, Jun Su, Tianguang Meng, Yi~Liu, Qingguo Meng, Yu~Wang, and Yong Li.
\newblock Agentsociety 2: An integrated research environment for executable
  social science.
\newblock \emph{arXiv preprint arXiv:2607.11895}, 2026.
\newblock Version 2.

\bibitem[Li et~al.(2026{\natexlab{a}})Li, Monteiro, Shirado, and
  Das]{li2026whatif}
Yuxuan Li, Kyzyl Monteiro, Hirokazu Shirado, and Sauvik Das.
\newblock Whatif: Interactive exploration of llm-powered social simulations for
  policy reasoning.
\newblock \emph{arXiv preprint arXiv:2604.17615}, 2026{\natexlab{a}}.

\bibitem[Lee et~al.(2025)Lee, Di~Paola, Hong, Nguyen, and
  Seering]{lee2025injectforkcompare}
HwiJoon Lee, Martina Di~Paola, Yoo~Jin Hong, Quang-Huy Nguyen, and Joseph
  Seering.
\newblock Inject, fork, compare: Defining an interaction vocabulary for
  multi-agent simulation platforms.
\newblock \emph{arXiv preprint arXiv:2509.13712}, 2025.

\bibitem[Mou et~al.(2024)Mou, Wei, and Huang]{mou2024unveiling}
Xinyi Mou, Zhongyu Wei, and Xuanjing Huang.
\newblock Unveiling the truth and facilitating change: Towards agent-based
  large-scale social movement simulation.
\newblock In \emph{Findings of the Association for Computational Linguistics:
  ACL 2024}, 2024.
\newblock arXiv:2402.16333.

\bibitem[Park et~al.(2024)Park, Zou, Kamphorst, Egan, Shaw, Hill, Cai, Morris,
  Liang, Willer, and Bernstein]{park2024generative1000}
Joon~Sung Park, Carolyn~Q. Zou, Jonne Kamphorst, Niles Egan, Aaron Shaw,
  Benjamin~Mako Hill, Carrie Cai, Meredith~Ringel Morris, Percy Liang, Robb
  Willer, and Michael~S. Bernstein.
\newblock {LLM} agents grounded in self-reports enable general-purpose
  simulation of individuals.
\newblock \emph{arXiv preprint arXiv:2411.10109}, 2024.
\newblock Circulated in earlier versions as ``Generative Agent Simulations of
  1,000 People''.

\bibitem[Piao et~al.(2025)Piao, Yan, Zhang, Li, Yan, Lan, Lu, Zheng, Wang,
  Zhou, Gao, Xu, Zhang, Rong, Su, and Li]{piao2025agentsociety}
Jinghua Piao, Yuwei Yan, Jun Zhang, Nian Li, Junbo Yan, Xiaochong Lan, Zhihong
  Lu, Zhiheng Zheng, Jing~Yi Wang, Di~Zhou, Chen Gao, Fengli Xu, Fang Zhang,
  Ke~Rong, Jun Su, and Yong Li.
\newblock Agentsociety: Large-scale simulation of llm-driven generative agents
  advances understanding of human behaviors and society.
\newblock \emph{arXiv preprint arXiv:2502.08691}, 2025.

\bibitem[Zhang et~al.(2025{\natexlab{a}})Zhang, Lin, Mou, Yang, Liu, Sun, Lyu,
  Yang, Qi, Chen, Li, Yan, Hu, Chen, Wang, Huang, Luo, Tang, Wu, Zhou, and
  Wei]{zhang2025socioverse}
Xinnong Zhang, Jiayu Lin, Xinyi Mou, Shiyue Yang, Xiawei Liu, Libo Sun, Hanjia
  Lyu, Yihang Yang, Weihong Qi, Yue Chen, Guanying Li, Ling Yan, Yao Hu, Siming
  Chen, Yu~Wang, Xuanjing Huang, Jiebo Luo, Shiping Tang, Libo Wu, Baohua Zhou,
  and Zhongyu Wei.
\newblock Socioverse: A world model for social simulation powered by llm agents
  and a pool of 10 million real-world users.
\newblock \emph{arXiv preprint arXiv:2504.10157}, 2025{\natexlab{a}}.

\bibitem[Yang et~al.(2024)Yang, Zhang, Zheng, Jiang, Gan, Wang, Ling, Chen, Ma,
  Dong, Gupta, Hu, Yin, Li, Jia, Wang, Ghanem, Lu, Lu, Ouyang, Qiao, Torr, and
  Shao]{yang2024oasis}
Ziyi Yang, Zaibin Zhang, Zirui Zheng, Yuxian Jiang, Ziyue Gan, Zhiyu Wang,
  Zijian Ling, Jinsong Chen, Martz Ma, Bowen Dong, Prateek Gupta, Shuyue Hu,
  Zhenfei Yin, Guohao Li, Xu~Jia, Lijun Wang, Bernard Ghanem, Huchuan Lu,
  Chaochao Lu, Wanli Ouyang, Yu~Qiao, Philip Torr, and Jing Shao.
\newblock Oasis: Open agent social interaction simulations with one million
  agents.
\newblock \emph{arXiv preprint arXiv:2411.11581}, 2024.

\bibitem[Zhang et~al.(2026)Zhang, Wang, Galhotra, and
  Cardie]{zhang2026autoresearch}
Zhengxin Zhang, Ning Wang, Sainyam Galhotra, and Claire Cardie.
\newblock How far are we from true auto-research?
\newblock \emph{arXiv preprint arXiv:2605.19156}, 2026.

\bibitem[Miyai et~al.(2025)Miyai, Toyooka, Otonari, Zhao, and
  Aizawa]{miyai2025jrscientist}
Atsuyuki Miyai, Mashiro Toyooka, Takashi Otonari, Zaiying Zhao, and Kiyoharu
  Aizawa.
\newblock Jr. {AI} scientist and its risk report: Autonomous scientific
  exploration from a baseline paper.
\newblock \emph{arXiv preprint arXiv:2511.04583}, 2025.

\bibitem[Zhu et~al.(2025)Zhu, Xie, Weng, Wu, Lin, Yang, and
  Zhang]{zhu2025implementation}
Minjun Zhu, Qiujie Xie, Yixuan Weng, Jian Wu, Zhen Lin, Linyi Yang, and Yue
  Zhang.
\newblock {AI} scientists fail without strong implementation capability.
\newblock \emph{arXiv preprint arXiv:2506.01372}, 2025.

\bibitem[Schmidgall et~al.(2025)Schmidgall, Su, Wang, Sun, Wu, Yu, Liu, Moor,
  Liu, and Barsoum]{schmidgall2025agentlab}
Samuel Schmidgall, Yusheng Su, Ze~Wang, Ximeng Sun, Jialian Wu, Xiaodong Yu,
  Jiang Liu, Michael Moor, Zicheng Liu, and Emad Barsoum.
\newblock Agent laboratory: Using {LLM} agents as research assistants.
\newblock \emph{arXiv preprint arXiv:2501.04227}, 2025.

\bibitem[Gottweis et~al.(2025)Gottweis, Weng, Daryin, Tu, Sirkovic, Myaskovsky,
  et~al.]{gottweis2025coscientist}
Juraj Gottweis, Wei-Hung Weng, Alexander Daryin, Tao Tu, Petar Sirkovic, Artiom
  Myaskovsky, et~al.
\newblock Accelerating scientific discovery with {Co-Scientist}.
\newblock \emph{arXiv preprint arXiv:2502.18864}, 2025.

\bibitem[Zhu and Wang(2026)]{zhu2026hler}
Chen Zhu and Xiaolu Wang.
\newblock {HLER}: Human-in-the-loop economic research via multi-agent pipelines
  for empirical discovery.
\newblock \emph{arXiv preprint arXiv:2603.07444}, 2026.

\bibitem[Abbott(2001)]{abbott2001time}
Andrew Abbott.
\newblock \emph{Time Matters: On Theory and Method}.
\newblock University of Chicago Press, 2001.

\bibitem[Halaby(2004)]{halaby2004panel}
Charles~N. Halaby.
\newblock Panel models in sociological research: Theory into practice.
\newblock \emph{Annual Review of Sociology}, 30:\penalty0 507--544, 2004.
\newblock \doi{10.1146/annurev.soc.30.012703.110629}.

\bibitem[Holland(1986)]{holland1986statistics}
Paul~W. Holland.
\newblock Statistics and causal inference.
\newblock \emph{Journal of the American Statistical Association}, 81\penalty0
  (396):\penalty0 945--960, 1986.

\bibitem[Rubin(1974)]{rubin1974estimating}
Donald~B. Rubin.
\newblock Estimating causal effects of treatments in randomized and
  nonrandomized studies.
\newblock \emph{Journal of Educational Psychology}, 66\penalty0 (5):\penalty0
  688--701, 1974.
\newblock \doi{10.1037/h0037350}.

\bibitem[Campbell(1969)]{campbell1969reforms}
Donald~T. Campbell.
\newblock Reforms as experiments.
\newblock \emph{American Psychologist}, 24\penalty0 (4):\penalty0 409--429,
  1969.

\bibitem[Morgan and Winship(2015)]{morgan2015counterfactuals}
Stephen~L. Morgan and Christopher Winship.
\newblock \emph{Counterfactuals and Causal Inference: Methods and Principles
  for Social Research}.
\newblock Cambridge University Press, 2 edition, 2015.
\newblock \doi{10.1017/CBO9781107587991}.

\bibitem[Lu et~al.(2024)Lu, Lu, Lange, Foerster, Clune, and
  Ha]{lu2024aiscientist}
Chris Lu, Cong Lu, Robert~Tjarko Lange, Jakob Foerster, Jeff Clune, and David
  Ha.
\newblock The {AI} scientist: Towards fully automated open-ended scientific
  discovery.
\newblock \emph{arXiv preprint arXiv:2408.06292}, 2024.

\bibitem[Horvitz(1999)]{horvitz1999mixed}
Eric Horvitz.
\newblock Principles of mixed-initiative user interfaces.
\newblock In \emph{Proceedings of the SIGCHI Conference on Human Factors in
  Computing Systems (CHI)}, pages 159--166, 1999.
\newblock \doi{10.1145/302979.303030}.

\bibitem[Amershi et~al.(2019)Amershi, Weld, Vorvoreanu, Fourney, Nushi,
  Collisson, Suh, Iqbal, Bennett, Inkpen, Teevan, Kikin-Gil, and
  Horvitz]{amershi2019guidelines}
Saleema Amershi, Dan Weld, Mihaela Vorvoreanu, Adam Fourney, Besmira Nushi,
  Penny Collisson, Jina Suh, Shamsi Iqbal, Paul~N. Bennett, Kori Inkpen, Jaime
  Teevan, Ruth Kikin-Gil, and Eric Horvitz.
\newblock Guidelines for human-ai interaction.
\newblock In \emph{Proceedings of the 2019 CHI Conference on Human Factors in
  Computing Systems (CHI)}, pages 1--13, 2019.
\newblock \doi{10.1145/3290605.3300233}.

\bibitem[Shneiderman(2020)]{shneiderman2020hcai}
Ben Shneiderman.
\newblock Human-centered artificial intelligence: Reliable, safe \&
  trustworthy.
\newblock \emph{International Journal of Human--Computer Interaction},
  36\penalty0 (6):\penalty0 495--504, 2020.
\newblock \doi{10.1080/10447318.2020.1741118}.

\bibitem[Messeri and Crockett(2024)]{messeri2024illusions}
Lisa Messeri and M.~J. Crockett.
\newblock Artificial intelligence and illusions of understanding in scientific
  research.
\newblock \emph{Nature}, 627\penalty0 (8002):\penalty0 49--58, 2024.
\newblock \doi{10.1038/s41586-024-07146-0}.

\bibitem[Wang et~al.(2023)Wang, Fu, Du, Gao, Huang, Liu, Chandak, Liu,
  Van~Katwyk, Deac, et~al.]{wang2023scientific}
Hanchen Wang, Tianfan Fu, Yuanqi Du, Wenhao Gao, Kexin Huang, Ziming Liu, Payal
  Chandak, Shengchao Liu, Peter Van~Katwyk, Andreea Deac, et~al.
\newblock Scientific discovery in the age of artificial intelligence.
\newblock \emph{Nature}, 620:\penalty0 47--60, 2023.
\newblock \doi{10.1038/s41586-023-06221-2}.

\bibitem[Bail(2024)]{bail2024generative}
Christopher~A. Bail.
\newblock Can generative {AI} improve social science?
\newblock \emph{Proceedings of the National Academy of Sciences}, 121\penalty0
  (21):\penalty0 e2314021121, 2024.
\newblock \doi{10.1073/pnas.2314021121}.

\bibitem[Granovetter(1985)]{granovetter1985economic}
Mark Granovetter.
\newblock Economic action and social structure: The problem of embeddedness.
\newblock \emph{American Journal of Sociology}, 91\penalty0 (3):\penalty0
  481--510, 1985.

\bibitem[Santurkar et~al.(2023)Santurkar, Durmus, Ladhak, Lee, Liang, and
  Hashimoto]{santurkar2023whose}
Shibani Santurkar, Esin Durmus, Faisal Ladhak, Cinoo Lee, Percy Liang, and
  Tatsunori Hashimoto.
\newblock Whose opinions do language models reflect?
\newblock In \emph{Proceedings of the 40th International Conference on Machine
  Learning (ICML)}, 2023.
\newblock arXiv:2303.17548.

\bibitem[Dillion et~al.(2023)Dillion, Tandon, Gu, and Gray]{dillion2023replace}
Danica Dillion, Niket Tandon, Yuling Gu, and Kurt Gray.
\newblock Can {AI} language models replace human participants?
\newblock \emph{Trends in Cognitive Sciences}, 27\penalty0 (7):\penalty0
  597--600, 2023.
\newblock \doi{10.1016/j.tics.2023.04.008}.

\bibitem[Li et~al.(2024)Li, Gao, Li, Li, and Liao]{li2023econagent}
Nian Li, Chen Gao, Mingyu Li, Yong Li, and Qingmin Liao.
\newblock {EconAgent}: Large language model-empowered agents for simulating
  macroeconomic activities.
\newblock In \emph{Proceedings of the 62nd Annual Meeting of the Association
  for Computational Linguistics (ACL)}, pages 15523--15536, 2024.
\newblock arXiv:2310.10436.

\bibitem[Lewin(1936)]{lewin1936principles}
Kurt Lewin.
\newblock \emph{Principles of Topological Psychology}.
\newblock McGraw-Hill, 1936.

\bibitem[{NVIDIA}(2025)]{nvidia2025nemotron}
{NVIDIA}.
\newblock Nemotron-personas-usa: Synthetic personas aligned to real-world
  demographic distributions.
\newblock \url{https://huggingface.co/datasets/nvidia/Nemotron-Personas-USA},
  2025.
\newblock Synthetic persona dataset, CC-BY-4.0.

\bibitem[Li et~al.(2026{\natexlab{b}})Li, Hao, Hou, Huang, Wen, Huang, Liu,
  Liu, Fan, Wang, et~al.]{matraix2026persona}
Xiaomin Li, Yuexing Hao, Jianheng Hou, Jintao Huang, Qianfeng Wen, Shirley
  Huang, Yifan Liu, Xiaoyi Liu, Yilan Fan, Yijun Wang, et~al.
\newblock {MatrAIx}: Simulating the world with 8.3 billion persona agents.
\newblock \emph{arXiv preprint arXiv:2608.04205}, 2026{\natexlab{b}}.
\newblock Persona 1M public coreset:
  \url{https://huggingface.co/datasets/MatrAIx2026/MatrAIx_Persona_1M_Public_Release}.

\bibitem[Ge et~al.(2024)Ge, Chan, Wang, Yu, Mi, and Yu]{ge2024personahub}
Tao Ge, Xin Chan, Xiaoyang Wang, Dian Yu, Haitao Mi, and Dong Yu.
\newblock Scaling synthetic data creation with 1,000,000,000 personas.
\newblock \emph{arXiv preprint arXiv:2406.20094}, 2024.

\bibitem[Zhang et~al.(2025{\natexlab{b}})]{zhang2025abm}
Xinnong Zhang et~al.
\newblock Employing large language models in agent-based social science
  simulations.
\newblock \url{https://github.com/Lishi905/SocioVerse-ABM}, 2025{\natexlab{b}}.
\newblock Code, configurations, and results of the companion ABM benchmark
  study.

\bibitem[Deffuant et~al.(2000)Deffuant, Neau, Amblard, and
  Weisbuch]{deffuant2000mixing}
Guillaume Deffuant, David Neau, Frederic Amblard, and Gerard Weisbuch.
\newblock Mixing beliefs among interacting agents.
\newblock \emph{Advances in Complex Systems}, 3:\penalty0 87--98, 2000.

\bibitem[Wang et~al.(2026)Wang, Xu, Ding, Li, He, Liu, and
  Wei]{wang2026procuregymmultiagentmarkovgame}
Jia Wang, Qian Xu, Xuanwen Ding, Zhuangqi Li, Chao He, Bao Liu, and Zhongyu
  Wei.
\newblock Procuregym: A multi-agent markov game framework for modeling national
  volume-based drug procurement, 2026.
\newblock URL \url{https://arxiv.org/abs/2603.23880}.

\bibitem[Huang et~al.(2026)Huang, Yin, Lin, Zhang, Wang, Wang, Huang, Jin, and
  Wei]{huang2026consumersim}
Yixu Huang, Yunlu Yin, Jiayu Lin, Xinnong Zhang, Jia Wang, Siyuan Wang,
  Xuanjing Huang, Liyin Jin, and Zhongyu Wei.
\newblock Uncovering salience-driven dynamics in consumer confidence with
  generative social simulation.
\newblock \emph{arXiv preprint arXiv:2606.30395}, 2026.
\newblock Version 1.

\bibitem[Nagel and Schreckenberg(1992)]{nagel1992cellular}
Kai Nagel and Michael Schreckenberg.
\newblock A cellular automaton model for freeway traffic.
\newblock \emph{Journal de Physique I}, 2\penalty0 (12):\penalty0 2221--2229,
  1992.

\bibitem[Reynolds(1987)]{reynolds1987flocks}
Craig~W. Reynolds.
\newblock Flocks, herds and schools: A distributed behavioral model.
\newblock In \emph{Proceedings of the 14th Annual Conference on Computer
  Graphics and Interactive Techniques (SIGGRAPH)}, pages 25--34, 1987.

\bibitem[Helbing et~al.(2000)Helbing, Farkas, and
  Vicsek]{helbing2000simulating}
Dirk Helbing, Ill{\'e}s Farkas, and Tam{\'a}s Vicsek.
\newblock Simulating dynamical features of escape panic.
\newblock \emph{Nature}, 407:\penalty0 487--490, 2000.

\bibitem[Challet and Zhang(1997)]{challet1997emergence}
Damien Challet and Yi-Cheng Zhang.
\newblock Emergence of cooperation and organization in an evolutionary game.
\newblock \emph{Physica A: Statistical Mechanics and its Applications},
  246\penalty0 (3--4):\penalty0 407--418, 1997.

\bibitem[Axelrod(1984)]{axelrod1984evolution}
Robert Axelrod.
\newblock \emph{The Evolution of Cooperation}.
\newblock Basic Books, 1984.

\bibitem[Epstein(2002)]{epstein2002modeling}
Joshua~M. Epstein.
\newblock Modeling civil violence: An agent-based computational approach.
\newblock \emph{Proceedings of the National Academy of Sciences}, 99\penalty0
  (suppl 3):\penalty0 7243--7250, 2002.

\bibitem[Kermack and McKendrick(1927)]{kermack1927contribution}
William~Ogilvy Kermack and Anderson~G. McKendrick.
\newblock A contribution to the mathematical theory of epidemics.
\newblock \emph{Proceedings of the Royal Society of London. Series A},
  115\penalty0 (772):\penalty0 700--721, 1927.

\bibitem[Hegselmann and Krause(2002)]{hegselmann2002opinion}
Rainer Hegselmann and Ulrich Krause.
\newblock Opinion dynamics and bounded confidence: Models, analysis and
  simulation.
\newblock \emph{Journal of Artificial Societies and Social Simulation},
  5\penalty0 (3), 2002.

\end{thebibliography}
\end{document}